\documentclass[manuscript,nonacm]{acmart}
\setcopyright{none}
\renewcommand\footnotetextcopyrightpermission[1]{}
\authorsaddresses{}

\AtBeginDocument{%
  }

\setcopyright{acmlicensed}
\copyrightyear{2027}
\acmYear{2027}
\acmDOI{XXXXXXX.XXXXXXX}
\acmConference[IUI '27]{ACM Conference on Intelligent User Interfaces}{February 08--11,
  2027}{Helsinki, Finland}
\acmISBN{978-1-4503-XXXX-X/2018/06}

\acmSubmissionID{7653}

\definecolor{brown}{rgb}{0.59, 0.29, 0.0}
\definecolor{darkgray}{rgb}{0.59, 0.59, 0.59}
\definecolor{tablegray}{gray}{.9}

\usepackage[utf8]{inputenc}
\usepackage{diagbox}
\usepackage{colortbl}
\usepackage{subcaption}
\usepackage{tabularx}
\usepackage[percent]{overpic}

\usepackage{nameref}
\usepackage{todonotes}
\usepackage{multirow}

\usepackage{xspace}
\usepackage{enumitem}
\usepackage{mathtools}
\usepackage{commath}

\usepackage{amssymb}% http://ctan.org/pkg/amssymb
\usepackage{pifont}% http://ctan.org/pkg/pifont

\newcommand{\customtilde}{{\raise.17ex\hbox{$\scriptstyle\sim$}}}

\definecolor{Gray}{gray}{0.9}

\newcommand{\anovans}[5]{{\small [$F(#1,#2)$\,$=$\,$#3$, $p$\,$#4$\,$#5$]}}
\newcommand{\anova}[6]{{\small [$F(#1,#2)$\,$=$\,$#3$, $p$\,$#4$\,$#5$, $\eta_{p}^{2}$\,$=$\,$#6$]}}
\newcommand{\pval}[2]{{\small ($p\,#1\,#2$)}} % for p-values
\newcommand{\pvald}[3]{{\small ($p\,#1\,#2$, $d = #3$)}} % for p-values 
\newcommand{\pvall}[2]{{\small $p\,#1\,#2$}} % for p-values with parenthesis
\newcommand{\ttestd}[5]{{\small [$t(#1) = #2$, $p$\,$#3$\,$#4$, $d$ = $#5$]}}

\usepackage[normalem]{ulem}

\newcommand{\crit}[1]{C#1}
\newcommand{\rques}[1]{RQ#1}
\newcommand{\emref}[1]{\emph{emotional reference}}

\newcommand{\ivfilter}[1]{\textsc{image adaptation technique#1}}
\newcommand{\cdefault}[1]{\emph{original#1}}
\newcommand{\cgray}[1]{\emph{grayscale#1}}
\newcommand{\cdiff}[1]{\emph{RGDR#1}}
\newcommand{\cstyle}[1]{\emph{style opt.#1}}
\newcommand{\cparam}[1]{\emph{param. opt.#1}}

\newcommand{\diff}[1]{RGDR#1}
\newcommand{\style}[1]{style optimization#1}
\newcommand{\param}[1]{parametric optimization#1}
\newcommand{\man}[1]{manual#1}
\newcommand{\gray}[1]{grayscale#1}
\newcommand{\cfgedit}[1]{NTO-edit#1}
\newcommand{\cfg}[1]{NTO-instruct#1}
\begin{document}

\title{Regressor-Guided Image Editing Shifts Emotion and Disengagement Timing in Social Media}

\author{Christoph Gebhardt}
% \email{christoph.gebhardt@ost.ch}
\orcid{0000-0001-7162-0133}
\affiliation{%
  \institution{Eastern Switzerland University of Applied Sciences \& ETH Z\"urich}
  \city{St.Gallen}
  % \state{St.Gallen}
  \country{Switzerland}
}

\author{Robin Willardt}
\orcid{0000-0002-2495-3450}
\affiliation{%
  \institution{UNSW Sydney}
  \city{Sydney}
  \country{Australia}}

\author{Seyedmorteza Sadat}
\orcid{0009-0003-4668-5703}
\affiliation{%
  \institution{ETH Z\"urich}
  \city{Z\"urich}
  \country{Switzerland}
}

\author{Chih-Wei "Charlotte" Ning}
\orcid{0009-0003-5838-5062}
\affiliation{%
  \institution{ETH Z\"urich}
  \city{Z\"urich}
  \country{Switzerland}
 }

\author{Andreas Brombach}
\orcid{0009-0006-0215-3432}
\affiliation{%
  \institution{ETH Z\"urich}
  \city{Z\"urich}
  \country{Switzerland}
  }

\author{Jie Song}
\orcid{0009-0003-7484-1937}
\affiliation{%
  \institution{The Hong Kong University of Science and Technology (Guangzhou)}
  \city{Guangzhou}
  % \state{Guangdong}
  \country{China}
  }

\author{Otmar Hilliges}
\orcid{0000-0002-5068-3474}
\affiliation{%
  \institution{ETH Z\"urich}
  \city{Z\"urich}
  \country{Switzerland}
  }

\author{Christian Holz}
\orcid{0000-0001-9655-9519}
\affiliation{%
  \institution{ETH Z\"urich}
  \city{Z\"urich}
  \country{Switzerland}
  }

\renewcommand{\shortauthors}{Gebhardt et al.}

\begin{abstract}

Internet overuse is a widespread phenomenon in today's digital society. Existing interventions, such as time limits or grayscaling, often rely on restrictive controls that provoke psychological reactance and are frequently circumvented. Building on prior work showing that emotional responses mediate the relationship between content consumption and online engagement, we investigate whether regulating the emotional impact of images can reduce online use in a non-coercive manner.

We introduce and systematically analyze three regressor-guided image-editing approaches, spanning low-level attribute optimization, latent style-space optimization, and diffusion-based editing. While the first two modify low-level visual features (e.g., contrast, color), the diffusion-based method enables higher-level changes (e.g., adjusting clothing, facial features).

A controlled image-rating study shows that only the diffusion-based approach shifts perceived emotion toward a neutral reference without reducing perceived quality. In a follow-up social media experiment, edited images were associated with earlier disengagement among users who left the feed.
% , mirroring grayscaling's effect.

\end{abstract}




\begin{teaserfigure}
  \includegraphics[width=\textwidth]{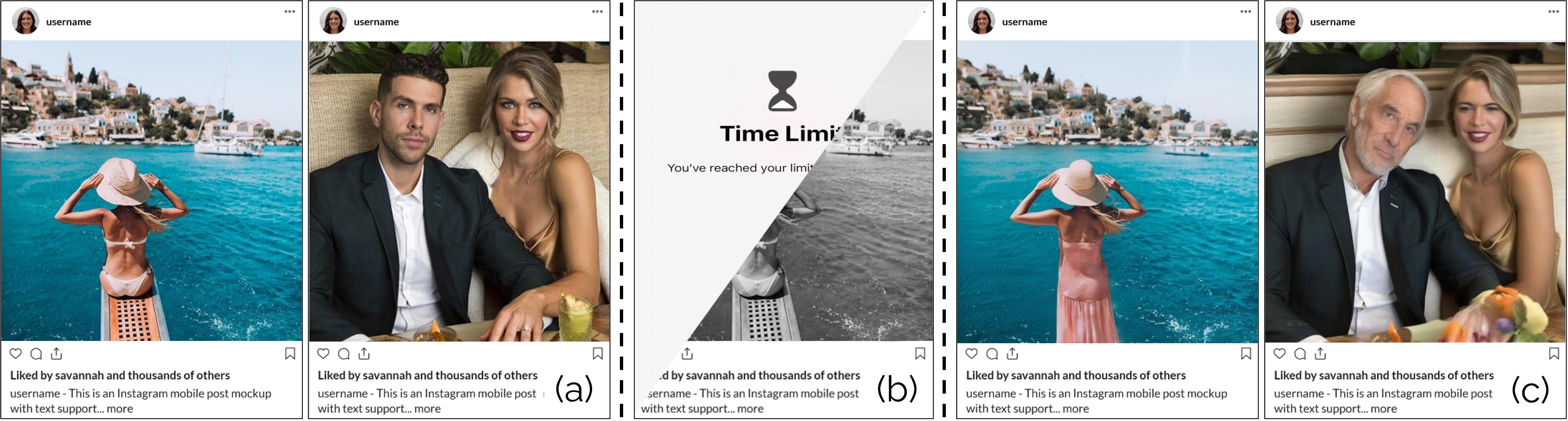}
  \caption{
    Traditional approaches to reducing online engagement, such as time limits or image grayscaling (b), can be restrictive and are often bypassed. 
    We introduce three regressor-guided image-editing methods designed to lower images’ emotional intensity without degrading visual quality. 
    Our diffusion-based variant takes the original image (a) and, through classifier and classifier-free guidance, produces edited images that reduce emotionally salient cues, here increasing clothing coverage and apparent age (c).
    % Such demographic shifts reflect how emotional salience is encoded in the affective databases the guidance regressor is trained on; we examine the resulting fairness implications in Sec.~\ref{sec:ethics}.
    In a controlled rating study, this approach lowered perceived emotional intensity while maintaining high perceived quality; in a follow-up social media study, presenting these edited images in a feed was associated with earlier disengagement among users who left the feed.
  % Captions
  % "Woman in a bikini and hat sitting on a boat deck, facing the sea and coastal town." 
  % "A man in a dark suit and a woman in a dress posing together at a dining table."
  }
  \label{fig:teaser}
  \Description{Comparison of Instagram posts showing original photos, a time limit screen, and adapted photos. Panel (a) displays two original posts with people posing outdoors and indoors. Panel (b) shows the time limit overlay partially obscuring a post. Panel (c) presents altered versions of the original posts with changed physical and facial appearances.}
\end{teaserfigure}


\maketitle

\section{Introduction}
The Internet stands as one of the most significant inventions of the 20th century. 
Among other benefits, it democratizes access to information and facilitates unprecedented ease of communication.
However, alongside these benefits, its use can also yield adverse effects for users.
Research increasingly links it to forms of addiction, including social media \cite{YoungUsersSocialMediaAddiction2021, OnlineSocialNetworkAddiction2015}, online gaming \cite{young2009understanding}, and pornography \cite{love2015neuroscience}.
This pattern of behavior, termed problematic Internet use (PIU), involves difficulty controlling Internet activity.
While prevalence estimates vary, recent data highlight the growing scale of the issue. For instance, PIU rates among adults are estimated at 10.1\% in the United Kingdom and 10.4\% in the United States \cite{lopez2023problematic}, and are even higher among adolescents (e.g., 31\% in Canada \cite{lavoie2023relationship}).
Users also report that the Internet distracts from work \cite{Gupta20161165} and wellbeing \cite{BurkeRelationshipFacebookWellBeing2016}, leading many to seek ways to reduce their usage \cite{lyngs2022goldilocks}.

To support users in managing their Internet consumption, a variety of tools have been introduced through research (e.g., \cite{kim2019goalkeeper,lu2024interactout}) and commercial products (e.g., \cite{ColdTurkey,Apple2023}).
However, these tools often impose strict restrictions (e.g., time limits) that provoke psychological reactance and are frequently circumvented~\cite{LyngsHackMyself2020}. Consequently, researchers argue that effective interventions should change behavior without being perceived as overly coercive~\cite{lyngs2022goldilocks}.

Meanwhile, marketing research highlights the pivotal role of emotions in shaping online behavior.
Emotional experiences characterized by positive valence \cite{garcia2016dynamics} and, to an even greater extent, heightened arousal levels \cite{berger2012makes,NelsonField2013}, boost online engagement.
This suggests that emotional processes mediate the relationship between digital content consumption and online engagement \cite{yu2014we,schreiner2021impact}.
Psychological research further shows that this emotional response is shaped by image properties at multiple levels of abstraction: from low-level properties such as saturation, brightness, and contrast \cite{bekhtereva2017bringing,redies2020global}, to high-level, semantic content such as facial expression or nudity \cite{nummenmaa2012gender,aharon2001beautiful}.
Separately, field studies of session-ending behavior show that declining emotional interest in content is consistently among the most common reasons for voluntarily ending a social media session \cite{rixen2023loop}.
% reported by users as "boredom"
%
Together, this raises the question of whether directly editing the emotional properties of viewed images, rather than restricting access outright, could offer a subtler, non-coercive route to reducing online use.  
% It also raises a second question: whether the level of visual abstraction at which an edit operates shapes its effectiveness.

We address this question by introducing three regressor-guided image-editing
approaches that span from low-level parameter adjustment to high-level semantic
editing: \param{}, which adjusts global image properties
such as brightness and contrast; \style{}, which alters latent image style while leaving depicted content intact; and regressor-guided diffusion resampling (\diff{}), which enables semantic changes such as clothing or facial expression. All three are guided by a regressor grounded in a psychological model of emotion~\cite{russell_circumplex_1980}, steering edits toward a common \emref{} we define in Section~\ref{sec:design-criteria} (neutral valence, low arousal). We treat grayscaling, a content-agnostic filter already deployed as a digital wellbeing feature in mobile operating systems, as a fourth, external anchor point at the low-level end of this spectrum throughout our evaluation.
We evaluate all four approaches against our design criteria: emotional shift, perceived quality, and fidelity to the original image.
We then test whether the strongest-performing approach also reduces time
spent browsing. 
Our evaluation proceeds in three stages: technical, perceptual, and behavioral.

Our technical analysis shows that all our approaches move images closer to the \emref{} than baselines while maintaining high visual quality. Fidelity to the original image, however, separates them: style and parametric optimization preserve scene content while deviating photometrically, whereas \diff{} stays closer photometrically but alters semantic content.
To establish at what level of visual abstraction emotion-targeted edits reliably shift perceived emotion without reducing perceived image quality, we conducted a controlled image-rating study ($N=56$). Only \diff{} shifted perceived emotion toward the \emref{} without a significant reduction in perceived quality.

Building on this result, we tested whether an intervention at this abstraction level reduces time spent browsing a social media feed in a preregistered social media study ($N=188$) comparing \diff{} against original and grayscale feeds.
The registered hypotheses were not supported: total time in the feed did not differ across conditions.
Exploratory analyses indicated that conditions differed in \emph{when} participants disengaged, with both \diff{} and grayscale associated with earlier disengagement among participants who left the feed.

Qualitative analysis showed that \diff{} achieved emotional modulation through systematic semantic changes, such as altered expressions and increased clothing coverage (Fig.~\ref{fig:teaser}, left post). It also shifted demographic attributes, making subjects appear older (Fig.~\ref{fig:teaser}, right post) and, in several cases, altering apparent skin tone. Because these edits alter photographs of real people, they raise ethical concerns, which we develop in Section~\ref{sec:ethics}.

In summary, our contributions are:

\begin{itemize}
%[noitemsep,leftmargin=*]

\item Three regressor-guided image-editing approaches spanning low- to high-level visual abstraction, grounded in a common psychological model of emotion.

\item Empirical evidence ($N=56$) that the ability to shift perceived emotion toward a neutral reference without degrading perceived quality is concentrated at the high-abstraction end of this spectrum (\diff{}).

\item A behavioral test of this approach in a social media feed ($N=182$), revealing an exploratory disengagement-timing pattern that mirrors grayscaling, an already-deployed digital wellbeing baseline.

\item A characterization of a consistent trade structure across approaches: emotional efficacy, perceived image quality, and semantic fidelity to the original cannot currently be maximized simultaneously. 
We further find that technical quality metrics (KID/FID) can dissociate from perceived quality.

\end{itemize}

Taken together, these contributions offer an empirical characterization of the design space for emotion-targeted image editing as a subtle, non-coercive route to healthier online behavior. 
To support reproducibility, we plan to release the source code of our algorithms, the data from our experiments, the analysis scripts, and the experimental platform used in our studies upon acceptance.

\section{Related Work}
\label{sec:related-work}
Our work relates to digital self-control tools, research on emotion's influence on online behavior, computational methods for emotion-adapting image generation, and the literature on how image properties shape emotional response. 
% Together, the latter two research lines span a spectrum from low-level image properties to high-level visual content.
% Our three regressor-guided approaches are designed to instantiate this spectrum.

\subsection{Digital self-control tools}
\label{sec:related-work:dscts}
Digital self-control tools (DSCTs) are applications that help users to follow their own long-term goals by avoiding digital distractions and unwanted habitual use. %(e.g., productivity, free time)
These tools mostly focus on increasing productivity or reclaiming free time \cite{freedom,ColdTurkey,SelfControl}.
Specialized DSCTS for parental control \cite{qustodio} or to avoid porn \cite{Covenanteyes} are also available.
Tools employ a wide range of strategies to increase self-control, e.g., hiding content on distracting websites \cite{JDev2019}, limiting functionality \cite{LessPhone2019}, gamification \cite{Forest2018}, or visualize time spent online \cite{Apple2023}.

% \cite{kovacs2018rotating,lottridge2012browser,lyngs2019self,pinder2019push,monge2019race,tseng2019overcoming}) 
Human-computer interaction (HCI) research (e.g., \cite{lyngs2019self,monge2019race,tseng2019overcoming}) has also investigated design patterns that help users regulate internet use, e.g., blocking distractions \cite{kim2019goalkeeper}, reminding users about their long-term goals \cite{ko2015nugu}. 
% Most similar to our research is work that leverages implicit input manipulation techniques to inhibit the natural execution of common user gestures on mobile devices to reduce the usage time \cite{lu2024interactout}.

However, prior work and applications mostly rely on simple decision heuristics to constrain digital media use. 
This often results in overly severe restrictions (e.g., time limits) that trigger psychological reactance \cite{brehm2013psychological,lukoff2022designing} and, hence, are regularly circumvented~\cite{LyngsHackMyself2020}.
This is why researchers advocate for the development of intervention mechanisms that are enforcing enough to change users' behavior without feeling too coercive~\cite{lyngs2022goldilocks}. 
In response, recent work leverages implicit input manipulation techniques to disrupt the execution of user gestures on mobile devices to reduce usage \cite{lu2024interactout}. However, these techniques still depend on inducing frustration, and thus negative affect.

In contrast, our approach aims to encourage natural disengagement by regulating emotional states toward neutral valence and low arousal, without enforcing breaks or inducing negative affect. 
Our image adaptations are further designed to preserve visual quality and remain close to the original content, with the aim of intervening noticeably enough to affect behavior while remaining subtle enough to preserve user autonomy.

\subsection{The role of emotions in online behavior}
\label{sec:emotions_internet_use}
% Emotions are commonly described by valence and arousal, where valence reflects the pleasantness or unpleasantness of an emotion and arousal its intensity \cite{russell_circumplex_1980}.
Research in marketing and psychology has explored how content influences emotions and, in turn, how these emotions affect online behavior.
A common conceptualization of emotions is the Circumplex Model of Affect (CMA) \cite{russell_circumplex_1980}, which characterizes them along two dimensions: valence, the degree of pleasantness, and arousal, the level of intensity.
%Drawing on this framework, we review key findings from prior work.

Studies have shown that engaging with online discussions increases emotional arousal, with higher participation rates observed when users experience positive valence \cite{garcia2016dynamics}.
In the context of brand posts, it was found that high arousal or positive valence tend to boost user engagement \cite{yu2014we}.

On the other hand, research has found that emotions of negative valence, such as anger and disgust, are the driving factors behind the rapid spreading of false news \cite{Vosoughi2018TrueFalseNews}. 
Other work shows that high-arousal emotions like anger and disgust promote the spread of moral content, while low-arousal negative emotions such as sadness reduce spread \cite{Brady2017EmotionMoralizedContent}.

% Furthermore, 
It was also shown that high arousal content, regardless of its valence, drives virality \cite{berger2012makes}.
Similarly, studies indicate that videos eliciting high arousal are more likely to be shared, while differences in valence have minimal impact \cite{NelsonField2013}.
A meta-review of these and related findings concludes that there is evidence suggesting that high emotional arousal and positive valence tend to enhance engagement behavior \cite{schreiner2021impact}.

In summary, these findings suggest that content which polarizes valence and elevates arousal tends to increase online engagement levels. 
Leveraging these insights, our approach adjusts visual content toward neutral valence and lower arousal, with the aim of reducing the time users spend online.

\subsection{Computational approaches to emotional image adaptation}
\label{sec:image_adaptations}
Various approaches have been proposed to adapt the emotional response that images elicit in viewers; we focus on techniques that adjust arbitrary visual content, as
opposed to work that manipulates depicted facial expressions.
% , which constitute a distinct research direction.
% We organize prior work into three categories: approaches based on standard image transformations, methods that leverage generative adversarial networks (GANs), and more recent techniques employing diffusion models (DMs). 

\subsubsection{Image transformations}
\label{sec:rw-transform}
Non-photorealistic rendering has been shown to shift intense emotions toward a more neutral state, typically through the introduction of confusion or loss of detail \cite{mould2012emotional}, and to reduce the repulsiveness of disgusting images \cite{besanccon2018reducing}. Other work uses neural networks to estimate an image's emotional distribution and applies standard transformations to shift its affect toward a target state \cite{ali2017automatic,peng2015mixed}, or optimizes intermediate-layer latent vectors to align an input image with a reference image selected by emotional vocabulary \cite{an2021global}.

\subsubsection{GAN-based methods}
\label{sec:rw-gan}
GANs have been used to align source images' emotional distributions with a target domain for cross-dataset emotion classification \cite{zhao2019cycleemotiongan, zhao2018emotiongan}, and, building on content–style disentanglement \cite{huang2018multimodal,lee2018diverse}, to transfer the style of an emotional reference image onto an input image, either globally \cite{zhu2022emotional} or at the object level via semantic segmentation and multiple reference images \cite{chen2020image}. 
While these approaches rely on reference images to represent a target emotional state, other work instead uses emotional classifiers or regressors: generating emotional landscape images from noise guided by valence-arousal targets \cite{park2020emotional}, or modifying the latent noise vector of a pretrained GAN to shift valence, using a regressor to verify the desired change \cite{goetschalckx2019ganalyze}.

\subsubsection{Diffusion-based methods}
\label{sec:rw-diffusion}
Recently, DMs have surpassed GANs in image synthesis quality \cite{dhariwal2021diffusion,rombach2022high}, motivating diffusion-based emotional adaptation. Several approaches fine-tune Instruct-Pix2Pix \cite{brooks2023instructpix2pix} on emotion-annotated data to adapt an image toward a discrete target emotion (e.g., happiness, fear) while preserving scene structure \cite{yang2025emoedit,lin2025make}, including a multi-agent extension that generates multiple visually distinct edits from the same input and target emotion \cite{mao2025emoagent}. More recent work moves beyond discrete categories, conditioning diffusion models jointly on textual prompts and valence-arousal coordinates: by fusing valence-arousal values into textual features via an emotion-embedding network \cite{dang2025emoticrafter}, by reinforcement-tuning the DM with vision-language-model-derived emotion rewards \cite{jia2025emofeedback2}, or by optimizing an emotion token via losses from off-the-shelf emotion classifiers applied at each denoising timestep \cite{xia2025muse}. 
% supporting both generation and adaptation 

\paragraph{Positioning our approaches.}
To enable a systematic evaluation across the three categories of prior work, we introduce representative approaches for each. First, we implement an \emph{image-transformation–based} method and a \emph{GAN-based} method that rely solely on an emotional regressor to adapt an input image.
This sets them apart from prior image-to-image work in these categories (Sec.~\ref{sec:rw-transform} \& \ref{sec:rw-gan}), which relies on such reference images and is consequently limited to emotional states for which example images are available. Regressor-only guidance instead enables theoretically grounded adaptation toward arbitrary points in the valence-arousal space of the CMA.

Second, existing \emph{diffusion-based} methods typically rely on the attention mechanism of the U-Net to induce emotional changes, using either classifier-based losses or emotionally neutral prompt conditioning. 
However, prior research has shown that, under the high guidance weights these approaches require, such mechanisms can yield images with excessively salient colors, oversaturation, or overly glossy rendering~\cite{karras2024guiding,sadat2024eliminating}.
These artifacts are also visible in the qualitative examples reported for the diffusion-based approaches mentioned above.
Such artifacts are inconsistent with our goal of producing emotionally neutral, low-arousal images. For this reason, we design a \emph{diffusion-based} approach that introduces emotional changes via gradients from an emotional regressor, thereby avoiding these artifacts. %stemming from the attention mechanism.

\subsection{The impact of image properties on emotional response}
\label{sec:image-properties}
Psychology research shows that specific image properties shape viewers’ emotional responses.

% To structure this review, we distinguish between low-level color and light properties, mid-level structural and spatial properties, and high-level content, each addressed in the subsections below.

% These responses can be described using the Circumplex Model of Affect, which characterizes emotions along two dimensions: valence, the degree of pleasantness, and arousal, the level of intensity \cite{russell_circumplex_1980}.
% Drawing on this framework, we review key findings on how image properties influence emotions.

\subsubsection{Low-level: Color and light properties}
Research shows that image attributes related to color and brightness strongly influence emotional responses. Increased \emph{brightness} elicits more positive emotions, is consistently preferred across stimuli, and is recognized as a universal emotional factor across cultures \cite{specker2018universal, gao2007analysis}.
\emph{Color} enhances emotional valence compared to grayscale as seen in brain responses \cite{cano2009affective} and make unpleasant scenes appear more negative and arousing \cite{bekhtereva2017bringing}. Higher \emph{saturation} is also linked to more positive ratings across affective image databases \cite{simola2015affective, redies2020global}. Finally, increasing \emph{contrast} raises valence, whereas reducing it lowers valence with only minor effects on arousal \cite{yang2020emotion}.

\subsubsection{Mid-level: Structural and spatial properties}
Structural and spatial features also shape emotional perception. Greater \emph{variability in edge orientations} correlates with higher arousal \cite{redies2020global}, while \emph{smooth contours} are rated more positively than angular ones \cite{damiano2021cues}. Low-frequency color or grayscale changes evoke higher arousal regardless of valence \cite{delplanque2007spatial}, and reduced resolution or \emph{blur} dampens both valence and arousal \cite{de2010effects}. Finally, \emph{symmetry}, widely considered a core aesthetic principle, is robustly associated with positive valence and arousal \cite{brachmann2017computational, bertamini2013implicit, bertamini2018neural, bertamini2019symmetry, makin2012symmetry}.

\subsubsection{High-level: Content}
Image content has a strong impact on emotional responses.
For example, \emph{nudity} has been shown to increase arousal \cite{nummenmaa2012gender}, while viewing \emph{attractive faces} consistently elicits positive emotions \cite{aharon2001beautiful,gerger2011faces,vartanian2013middle,hahn2014neural}. Additionally, \emph{complexity} is strongly linked to heightened arousal, as more complex images offer a variety of informational cues that capture attention and require greater mental energy \cite{berlyne1971aesthetics}.
Moreover, the \emph{personal semantic meaning} of an image has been found to have the most significant impact on emotional responses \cite{pilarczyk2014emotional,humphrey2012salience}. 
%This aligns with appraisal theory, which suggests that individual goals and desires influence emotional reactions to external objects and events \cite{scherer2001appraisal}.

\paragraph{Implications for our approaches.}
Building on these prior studies, we select emotionally relevant image properties as optimization variables for our parametric method. We further employ these properties to evaluate whether the adapted images change in the theoretically consistent direction. 

% Together with the generative methods introduced in Section~\ref{sec:image_adaptations}, this low- to high-level structure forms the basis of the abstraction spectrum our three approaches instantiate.

\section{Design Criteria for Emotion-Regulating Image Editing}
\label{sec:design-criteria}
% Based on the insights from prior work, we first define the emotional reference toward which we optimize images, and then the design criteria our approaches should fulfill.

\subsection{Emotional reference}
\label{sec:emotional-reference}
Building on the Circumplex Model of Affect and the engagement findings reviewed in Section~\ref{sec:emotions_internet_use}, we target a state of low arousal and neutral valence, the CMA region associated with the weakest engagement pull.

We define this target as the \emref{} and operationalize it using a regression model predicting valence and arousal from images, in accordance with the CMA. This regressor provides the guiding signal for our editing approaches, enabling optimization toward the specified emotional reference rather than an example-driven target.

\subsection{Design criteria}
\label{sec:crit-def}
Our goal is to support users in reducing their time spent online by regulating the emotional impact of online images, without introducing visually unpleasant or overly pronounced changes. We operationalize this goal as four design criteria. An approach should:

\begin{itemize}
\item [\textbf{\crit{1}:}] shift perceived valence toward the neutral midpoint,
\item [\textbf{\crit{2}:}] reduce perceived arousal,
\item [\textbf{\crit{3}:}] preserve overall image quality, and
\item [\textbf{\crit{4}:}] preserve fidelity to the original content.
\end{itemize}

\crit{1} and \crit{2} follow directly from the \emref{} defined above. \crit{3} and \crit{4} reflect a further constraint: an approach that achieves \crit{1} and \crit{2} only by degrading the image, or by departing substantially from its original content, would not constitute a subtle intervention, but a coercive or deceptive one. 
Quality refers to the general visual integrity of an image (e.g., absence of artifacts, coherence), fidelity to an edited image's similarity to its original.
% These four criteria jointly define the design space our three approaches operate within and guide our evaluation throughout the paper.

\section{Method}
\label{sec:method}
Building on the design criteria defined above, we introduce three image-editing approaches that instantiate an abstraction spectrum. All three share the same regressor-guided objective, steering images toward the \emref{}, while pursuing fidelity to the original image through mechanisms suited to their level of abstraction.
We first describe how the shared emotion regressor is trained, before detailing each approach in turn.

\subsection{Emotion regressor}
\label{sec:regressor}
We train the emotion regressor by fine-tuning a pre-trained ResNet50 on a combined dataset of 20,460 images from ten open-science emotional image databases providing human valence and arousal ratings consistent with the CMA (details in App.~\ref{app:regressor}). The regressor achieves a mean absolute error of 0.11 for valence and 0.12 for arousal on a validation split. Given this normalized [0,1] range and our criteria, we set the \emref{} to $v' = 0.5$, $a' = 0.0$.
%
% Val-MAE: 0.1093 Ar-MAE: 0.1235
% valence_mean Mean: 0.5741; Standard Deviation: 0.1565; Minimum: 0.1023; Maximum: 0.9664
% arousal_mean Mean: 0.4556; Standard Deviation: 0.1033; Minimum: 0.1360; Maximum: 0.9281

\begin{figure}[t]
	\centering
    \includegraphics[width=0.75\linewidth]{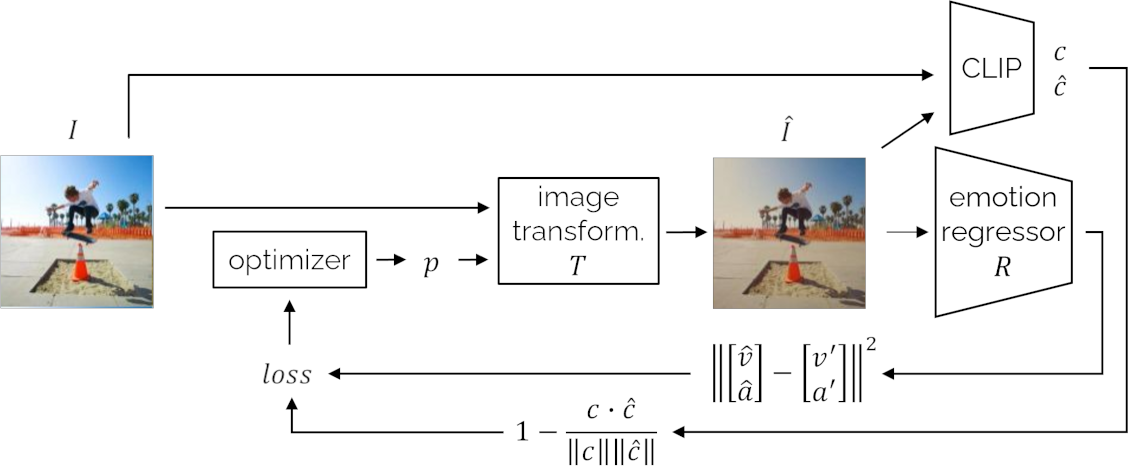} 
    \caption{
    \textbf{Parametric optimization.} 
     To adapt an input image \( I \), the parameters \( p \) of the differentiable image transformations \( T \) are optimized to produce a similar image that elicits a reduced emotional response, represented by \( \hat{I} = T(I, p) \).
    The optimization minimizes the distance between the predicted valence and arousal of the adapted image, \([\hat{v}, \hat{a}] = R(\hat{I})\), and the \emref{}, \([v', a']\). 
    It further ensures a high cosine similarity between the CLIP-space embeddings of the input image \( c = CLIP(I) \) and the transformed image \( \hat{c} = CLIP(\hat{I}) \).
    }
    \Description{Diagram of an image transformation framework with CLIP and emotion regression. An input image is optimized through parameterized transformations and decoded into a modified image. The output is evaluated by CLIP for content similarity and by an emotion regressor for affect alignment. Loss combines CLIP similarity and emotion distance to guide optimization.}

    \label{fig:optimization}
\end{figure}

\subsection{Parametric optimization}
\label{sec:parametric-optimization}

This approach optimizes parameters of differentiable image transformations to modulate the emotional response an image elicits. A key advantage is that the inferred parameters can be applied to images of arbitrary resolution and transferred to sequential frames of a video, as demonstrated in prior work \cite{mejjati2020look}. 
By relying on global transformations, i.e., functions applied uniformly across all pixels such as contrast and color adjustments, the method has no mechanism for altering the semantic content of the image.

To ground the method theoretically, we adapt image properties that have been shown to influence emotional responses (Sec.~\ref{sec:image-properties}).
% Specifically, we apply global transformations that adjust \emph{brightness} (exposure), \emph{saturation} (tone curves \cite{mejjati2020look}), \emph{color} (color curves \cite{hu2018exposure}), \emph{contrast}, \emph{sharpness} (via sharpening and blurring), \emph{overall blur}, and \emph{symmetry} (via translation and scaling).
The objective function leverages the emotion regressor to guide the adaptation towards the \emref{} (\crit{1} \& \crit{2}). In addition, we impose a loss on the CLIP embeddings \cite{radford2021learning} of the input and output images to preserve semantic similarity, which also indirectly constrains visual quality (\crit{3} \& \crit{4}). 
Figure~\ref{fig:optimization} illustrates the approach.

Specifically, the objective function of the parametric optimization is defined as:

\begin{equation}
\arg \min_{p \in \mathcal{P}} \, w_1 \mathcal{L}_{\text{CLIP}}(I, T(I, p)) + w_2 \left\| [v', a'] - R(T(I, p))  \right\|^2
\label{eq:param-opt}
\end{equation}

where $p$ represents the optimized parameters within the feasible set $\mathcal{P}$, $I$ is the input image, $T$ is the differentiable image transformation function, $R$ is the emotion regressor, and $[v', a']$ defines the \emref{}.
The term $\mathcal{L}_{\text{CLIP}}(I, T(I, p)) = 1 - \frac{c \cdot \hat{c}}{\|c\| \|\hat{c}\|}$ defines the cosine similarity between the CLIP embeddings of the original image \( c = \text{CLIP}(I) \) and the transformed image \( \hat{c} = \text{CLIP}(T(I, p)) \).
We detail the parameterization of $T$ in Appendix~\ref{apx:transform}.

\subsection{Style optimization}
\label{sec:style-optimization}
\begin{figure}[tbh]
	\centering
    \includegraphics[width=0.75\linewidth]{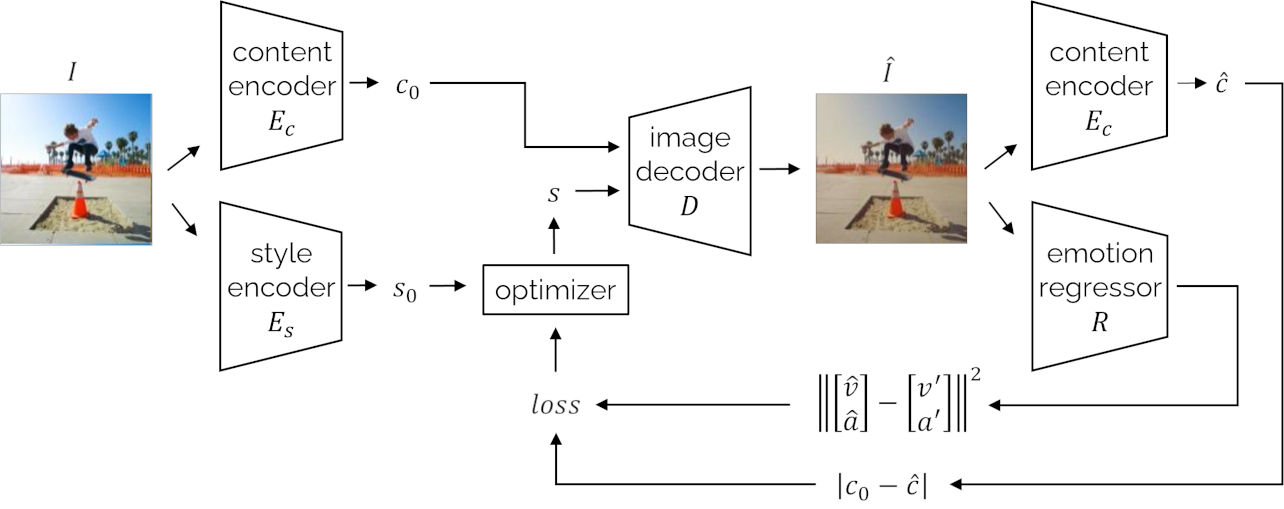} 
    \caption{
    \textbf{Style optimization.} 
    To adapt an input image \( I \), its latent style vector \( s \) is optimized to generate a similar image that elicits a reduced emotional response when decoded with \( I \)'s latent content \( c_0 = E_c(I) \), resulting in \( \hat{I} = D(c_0, s) \). 
    The optimization process minimizes the distance between the predicted valence and arousal of the adapted image, \([\hat{v}, \hat{a}] = R(\hat{I})\), and the target values specified by \emref{}, \([v', a']\). 
    Additionally, the optimization incorporates a term to ensure content consistency by minimizing the L1 loss between the encoded latent content of the generated image, \(\hat{c} = E_c(\hat{I})\), and \( c_0 \). 
    The style vector \( s \) is initialized as the encoded style vector of the input image, \( s_0 = E_s(I) \).
    }
    \Description{Diagram of a content and style optimization framework. An input image is encoded into content and style features, which are combined by an optimizer to guide image decoding. The reconstructed image is re-encoded and evaluated by an emotion regressor. Loss functions enforce similarity in content and desired emotion between the input and output representations.}
    \label{fig:munit}
\end{figure}

Our style-optimization approach modifies a latent style code to alter only low-level image features while preserving the depicted content.
To achieve this, we employ the MUNIT generator as a backbone to disentangle content and style \cite{huang2018multimodal} and optimize the latent style vector in alignment with our criteria. The overall approach is illustrated in Figure~\ref{fig:munit}.

To optimize the latent style, we apply the emotion regressor to the adapted image and compute a loss based on the difference between the predicted emotional values and the \emref{} (\crit{1} \& \crit{2}). In addition, we retain fidelity to the original image by incorporating the content-consistency loss proposed in MUNIT (\crit{4}). Using the pre-trained MUNIT decoder helps preserve high image quality (\crit{3}).

We formulate the optimization as

\begin{equation}
\arg \min_{s \in \mathcal{S}} \, w_1 |E_c(I) - E_c(D(E_c(I), s))| + w_2 \left\| [v', a'] - R(D(E_c(I), s))  \right\|^2 .
\end{equation}

Here, \( R \) denotes the emotion regressor, \([v', a']\) specifies the \emref{}, \( I \) is the input image, and \( E_c \) and \( D \) are the content encoder and decoder of MUNIT, respectively. The optimized style \( s \) is initialized as \( s_0 = E_s(I) \), where \( E_s \) is MUNIT's style encoder.
We train our own MUNIT model using NVIDIA's high-resolution implementation (App.~\ref{apx:munit}).

\subsection{Regressor-guided diffusion resampling (\diff{})}
\label{sec:rgdr}
\begin{figure}[t]
	\centering
    \includegraphics[width=0.95\linewidth]{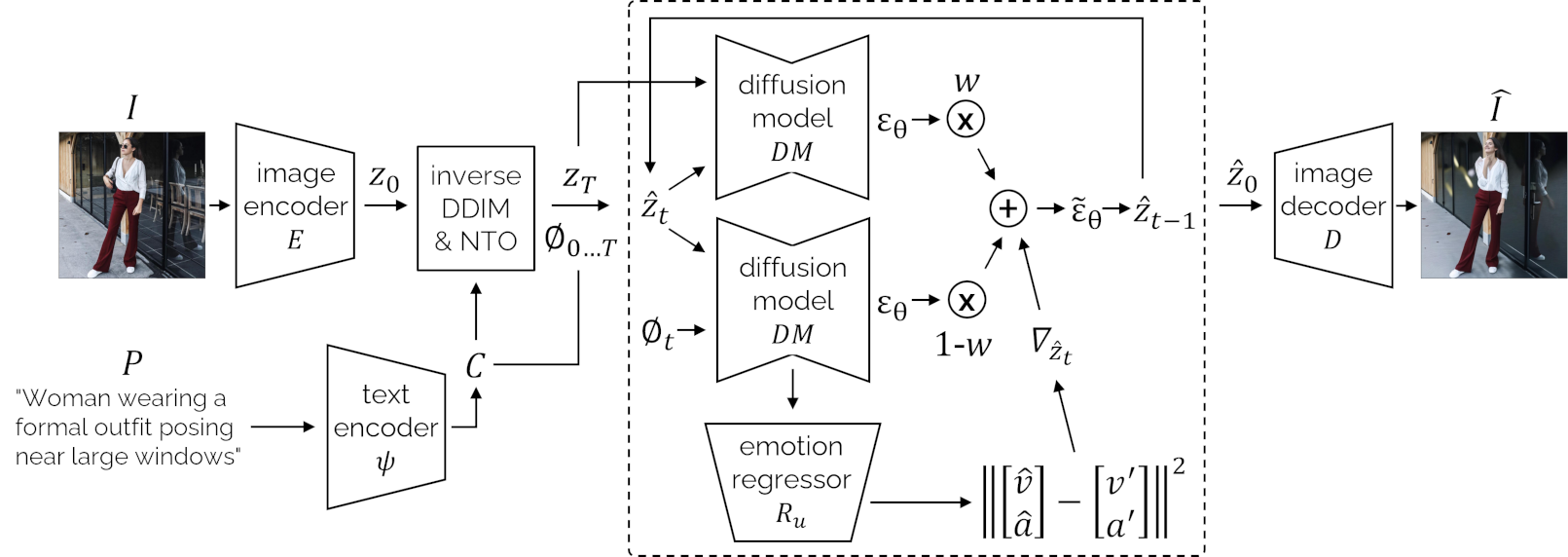} 
    \caption{
    \textbf{Regressor-guided diffusion resampling.}
    To adapt an input image \( I \), it is first encoded into a latent vector \( z_0 = E(I) \) and then inverted to its corresponding noise vector \( z_T = DDIM_{\text{inv}}(z_0, t, \mathcal{C}, \beta) \), while generating unconditional text embeddings for each timestep \(\{\emptyset_t\}_{t=1}^T = NTO([z_T, \ldots, z_0], \mathcal{C})\), ensuring editability of \(\hat{z}_t\) and fidelity to \( I \). 
    At each step of the denoising process, \(\hat{z}_t\) is updated by blending the predicted unconditional and conditional noise \(\epsilon_{\theta}\), further refined by a score \(\nabla_{z_t}\) that quantifies alignment with the \emref{}. 
    With the resulting noise vector $\Tilde{\epsilon}_{\theta}$, \( \hat{z}_{t-1} \) can be obtained. 
    The final latent vector \(\hat{z}_0\) is decoded into the image \(\hat{I} = D(\hat{z}_0)\), which closely resembles \( I \) while modulating the emotional response.
    }
    \Description{Diagram of a diffusion-based framework combining image and text encoding with emotion regression. An input image and caption are encoded into latent representations, which are processed through inverse diffusion steps. Two diffusion model branches generate weighted noise estimates, guided by an emotion regressor that compares predicted and target emotion vectors. The result is decoded back into a reconstructed image.}
    \label{fig:diffusion}
\end{figure}

The third approach leverages a diffusion model to enable higher-level image adaptations while preserving overall image fidelity. 
% We use SDXL \cite{podell2023sdxl} as the backbone and operate at 1024×1024.
Given an input image, \diff{} first inverts it into its corresponding noise vector, as detailed in the following subsections.
%
% using the diffusion model’s noise prediction and an inverse deterministic scheduler \cite{dhariwal2021diffusion,song2020denoising}. It then applies null-text optimization \cite{mokady2023null} to enhance editability of the latent representation. 
%
Subsequently, denoising is guided through two conditioning mechanisms: (1) classifier guidance (CG) \cite{dhariwal2021diffusion}, which steers the process toward neutral valence and low arousal (\crit{1} \& \crit{2}); and (2) classifier-free guidance (CFG) \cite{ho2022classifier}, which incorporates the caption of the original image to preserve fidelity (\crit{4}). Using the pre-trained SDXL backbone helps maintain high perceived image quality (\crit{3}). Figure~\ref{fig:diffusion} provides an overview.

In the following, we formally detail each step of the proposed approach. 
For background on diffusion models and the guidance mechanisms underlying our approach, see
the Supplementary Material.

\subsubsection{Guidance regressor}
To enable CG, we train a regressor $R_u$ that predicts valence and arousal from the output of the diffusion model's middle layer, $u_t$: $[\hat{v}, \hat{a}] = R_u(u_t)$. Compared to regressing directly on the noisy latent $z_t$, this is more memory-efficient, as guidance requires tracing gradients through only half of the diffusion model, and $u_t$ is a denser representation of $z_t$, filtering out irrelevant information and incorporating timestep context. Training details are provided in App.~\ref{app:guidance-regressor}.

\subsubsection{Inverting the input image}
To adapt an image $I$, we first obtain its latent image encoding $z_0 = E(I)$ and the text embedding of its caption $\mathcal{C} = \psi(\mathcal{P})$.
We then perform inverse DDIM \cite{dhariwal2021diffusion,song2020denoising} to attain an optimized noise vector $z_T = DDIM_{\text{inv}}(z_0, t, \mathcal{C}, \beta)$.
Next, we apply null-text optimization (NTO) \cite{mokady2023null} to derive unconditional embeddings for each timestep $\{\emptyset_t\}_{t=1}^T = NTO([z_T, \ldots, z_0], \mathcal{C})$. This ensures that during inference $z_t$ remains editable and that the resulting image stays closely aligned with the input image.

\subsubsection{Dual-conditioned resampling}
In this step, we resample \( \hat{z}_0 \) from \( z_T \) by conditioning the denoising process on the valence-arousal reference values \( v', a' \) and the caption embeddings \( \mathcal{C} \).
During each timestep \( t \), we condition the denoising trajectory by approximating the classifier guidance term using a score, computed as follows:
\begin{equation}
    \nabla_{\hat{z}_t} \log p_\phi(v', a'| \hat{z}_t) = \left\| [v', a'] - R_u(DM_{\text{half}}(\hat{z}_t, t, \emptyset)) \right\|^2,
\end{equation}
where $[v', a']$ defines the \emref{}.
The noise prediction is then carried out with the modified objective:
\begin{equation}
    \hat{\epsilon_\theta}(\hat{z}_t, t, \mathcal{C}, \emptyset_t) = w \cdot \epsilon_\theta(\hat{z}_t, t, \mathcal{C}) + (1 - w) \cdot \epsilon_\theta(\hat{z}_t, t, \emptyset_t) + s \cdot \nabla_{\hat{z}_t} \log p_\phi(v', a' | \hat{z}_t).
    \label{eq:rgdr}
\end{equation}
Finally, we perform the standard DDIM update step to obtain \( \hat{z}_{t-1} = DDIM(\hat{z}_t, t, \mathcal{C}, \beta)\).

From a probabilistic perspective, our approach can be interpreted as a diffusion probabilistic model $p_{\theta}(\hat{z}_0| z_T, \mathcal{P}, v', a')$ that generates an adapted image latent $\hat{z}_0$ conditioned on the noised latent vector of an input image $z_T$, its image caption $\mathcal{P}$, and specified reference values for valence $v'$ and arousal $a'$.

All approaches were implemented in PyTorch; optimizer settings, network details, and our mixed-precision reimplementation of NTO are given in Appendix~\ref{app:implementation}.

\section{Evaluation Overview}
\label{sec:evaluation_overview}
% (Sec.~\ref{sec:image_adaptations})
% (Sec.~\ref{sec:emotions_internet_use})
% Prior work shows that image manipulation affects viewer emotions and that strong emotions boost online engagement, but it remains unclear whether the emotional properties of images can be altered in a targeted fashion to regulate users' experienced emotions and, through that, reduce their time spent online. 
% We investigate two research questions:
Our research raises two questions, which structure our evaluation:

\begin{itemize}
\item [\textbf{\rques{1}:}] Can computational image adaptations effectively reduce emotional arousal and adjust experienced valence toward a neutral level while preserving image quality?
\item [\textbf{\rques{2}:}] Does presenting users with emotionally adapted images shift their experienced emotion and, through that, reduce their engagement duration with a social media feed?
\end{itemize}

Both questions address \crit{1}–\crit{3}, which concern the viewer's experience and are therefore evaluated perceptually. Fidelity to the original image (\crit{4}) is only evaluated technically, as evaluating it perceptually would require a comparative study paradigm, which we leave to future work (Sec.~\ref{sec:limitations}).
For \rques{2}, we choose social media as the evaluation context, as it is among the most widely used Internet services \cite{datareportal__we_are_social__meltwater_most_2023} and is linked to high arousal and polarized valence \cite{mauri_why_2011,de_vries_social_2018}. 

We address both questions in three stages: a technical evaluation against the defined criteria (Sec.~\ref{sec:technical-evaluation}), an image-rating study of perceived arousal, valence, and quality (Sec.~\ref{sec:image-study}), and a social media study of browsing duration (Sec.~\ref{sec:social-study}).
Both studies received IRB approval, were conducted with informed consent and collected only anonymous data.

\section{Technical Evaluation}
\label{sec:technical-evaluation}
We evaluate whether the proposed methods achieve the intended changes in valence and arousal (\crit{1} \& \crit{2}), as predicted by the regressor, while maintaining image quality (\crit{3}) and fidelity to the original image (\crit{4}). 
Because these approaches span a spectrum of visual abstraction, this evaluation also shows how abstraction level relates to each criterion.
We run different parameter configurations of our methods and baselines on the validation set of the COCO database~\cite{lin2014microsoft}.

\subsection{Baselines}
We compare our approaches against three baselines described below:

\begin{itemize}
    \item \textbf{Original}: The original set of images without any modification or adaptation, serving as baseline against which all adaptation approaches are compared.
    
    \item \textbf{Grayscale}: The original images converted to grayscale, removing all color information. Grayscaling is already deployed in mobile operating systems as a digital well-being feature (e.g., Android’s productivity mode, Apple’s color filters). 
    It also represents the low-level end of the visual abstraction spectrum introduced earlier.

    \item \textbf{Manual}: A fixed parameter configuration applied uniformly to all images, without per-image adaptation. It was manually tuned by one of the authors on a subset of the COCO validation images to satisfy the predefined criteria (App.~\ref{apx:manual-baseline}), representing a strong, informed reference against which to assess the benefit of per-image optimization over a manual color filter.
\end{itemize}

For \diff{}, we empirically found that a CFG scale of 2.0 provided the best trade-off between introducing meaningful changes to the image and preserving fidelity to the original content. As textual input, we used the image captions provided by the COCO dataset.
Since the resolution of images in the COCO dataset is approximately 512×512, we utilized the Diffusers implementation of Stable Diffusion from Stability AI as the backbone for \diff{} in this experiment.
% \footnote{\url{https://huggingface.co/stabilityai/stable-diffusion-2-1-base}}
Importantly, using a different backbone than in the other experiments demonstrates that \diff{} is agnostic to the underlying diffusion model.

For \diff{}, \param{}, and \style{}, the \emph{similarity-preserving objective} is controlled by a fixed weight ($w_1 = 1.0$ for parametric/style optimization, and $w = 2.0$ for \diff{}), while the \emph{emotional adaptation objective} is controlled by a variable weight ($w_2 \in \{0.1, 0.3, 0.5, 0.7, 1.0\}$ for parametric/style optimization, and $s \in \{0.05, 0.1, 0.2, 0.3, 0.4\}$ for \diff{}).

\subsection{Metrics}
We used our ResNet50 regressor (Sec. \ref{sec:regressor}) to estimate the valence (\crit{1}) and arousal of images (\crit{2}). To technically evaluate image quality (\crit{3}), we calculated the Kernel Inception Distance (KID) and the Frechet Inception Distance (FID). 
We further quantified resemblance to the original images using the L1 norm and a DINO score (\crit{4}). 
The DINO score is the cosine similarity between DINO CLS-token embeddings~\cite{caron2021emerging} and is an established, human-aligned proxy for content preservation~\cite{basu2023editval}.

\subsection{Results}
\label{sec:technical-evaluation:results}
% We first present a quantitative analysis of whether the proposed approaches introduce changes to images according to our criteria. We then examine the qualitative implications of these changes.

\begin{figure*}[t]
    \centering
    \begin{subfigure}{.33\textwidth}
        \centering
        \includegraphics[width=.95\linewidth]
        {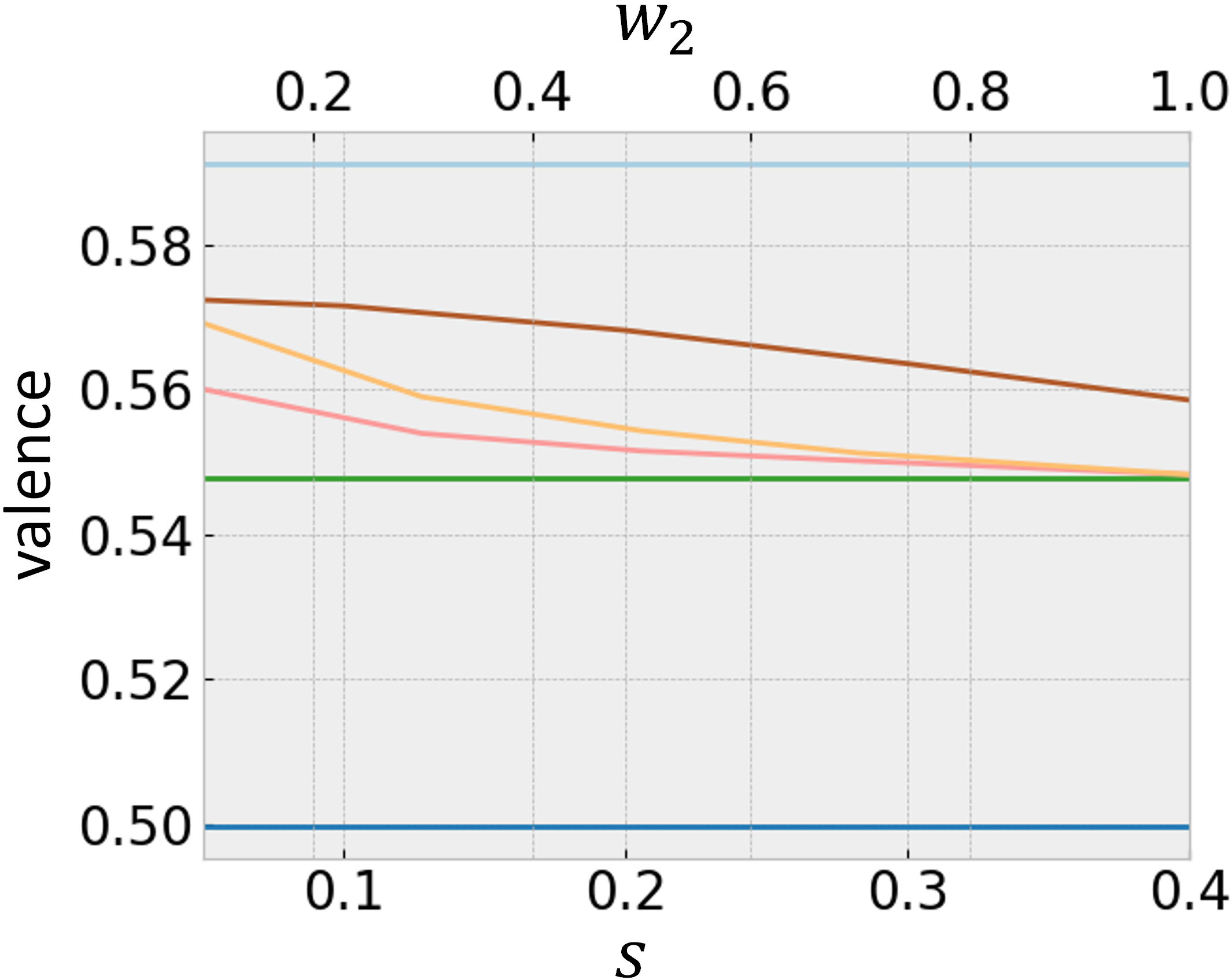}  
        \caption{valence}
        \label{fig:criteria_satisfaction:valence}
    \end{subfigure}
    %\hspace{-4mm}
    \begin{subfigure}{.33\textwidth}
        \centering
        \includegraphics[width=.95\linewidth]
        {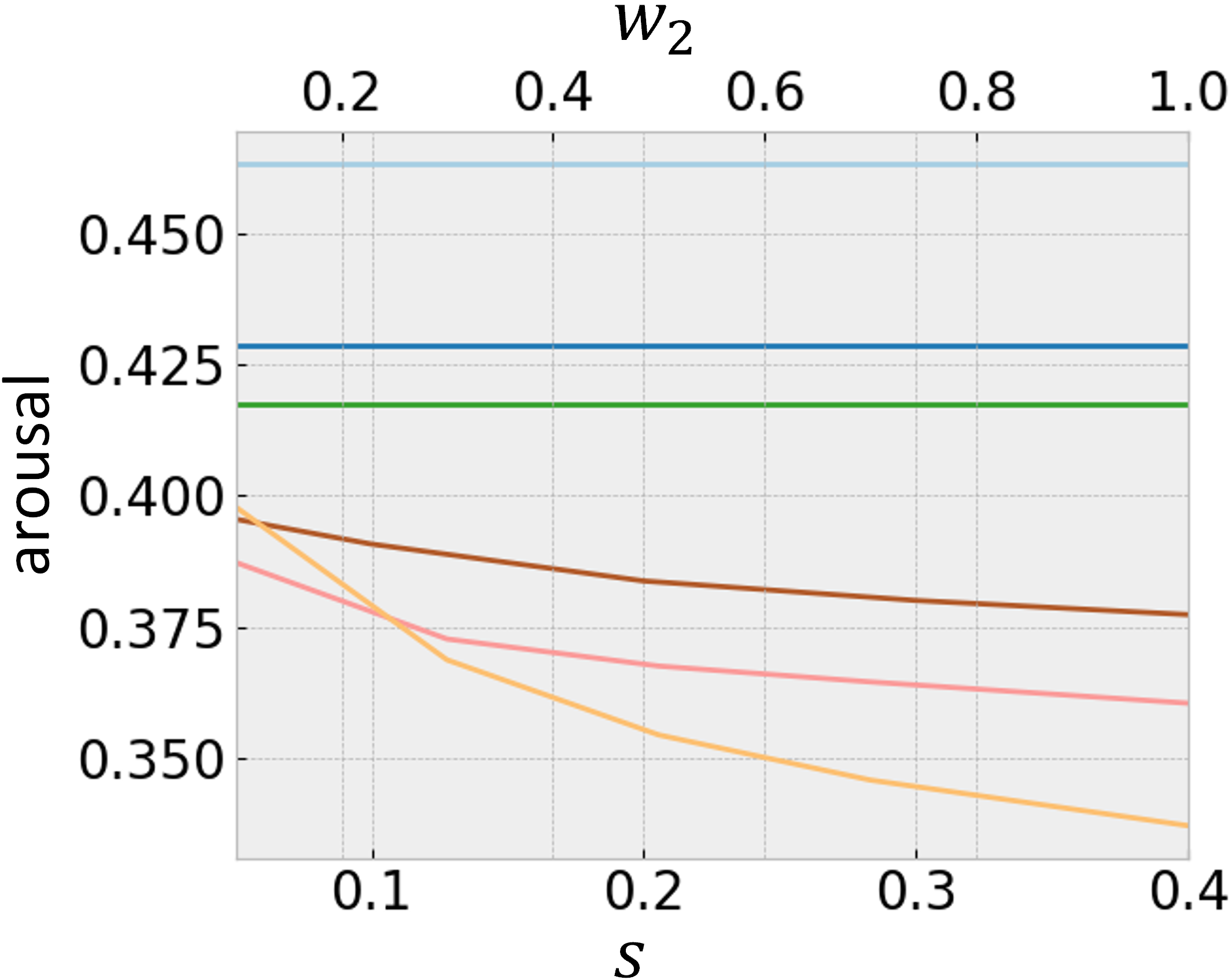}  
        \caption{arousal}
        \label{fig:criteria_satisfaction:arousal}
    \end{subfigure}
    \begin{subfigure}{.33\textwidth}
        \centering
        \includegraphics[width=.95\linewidth]{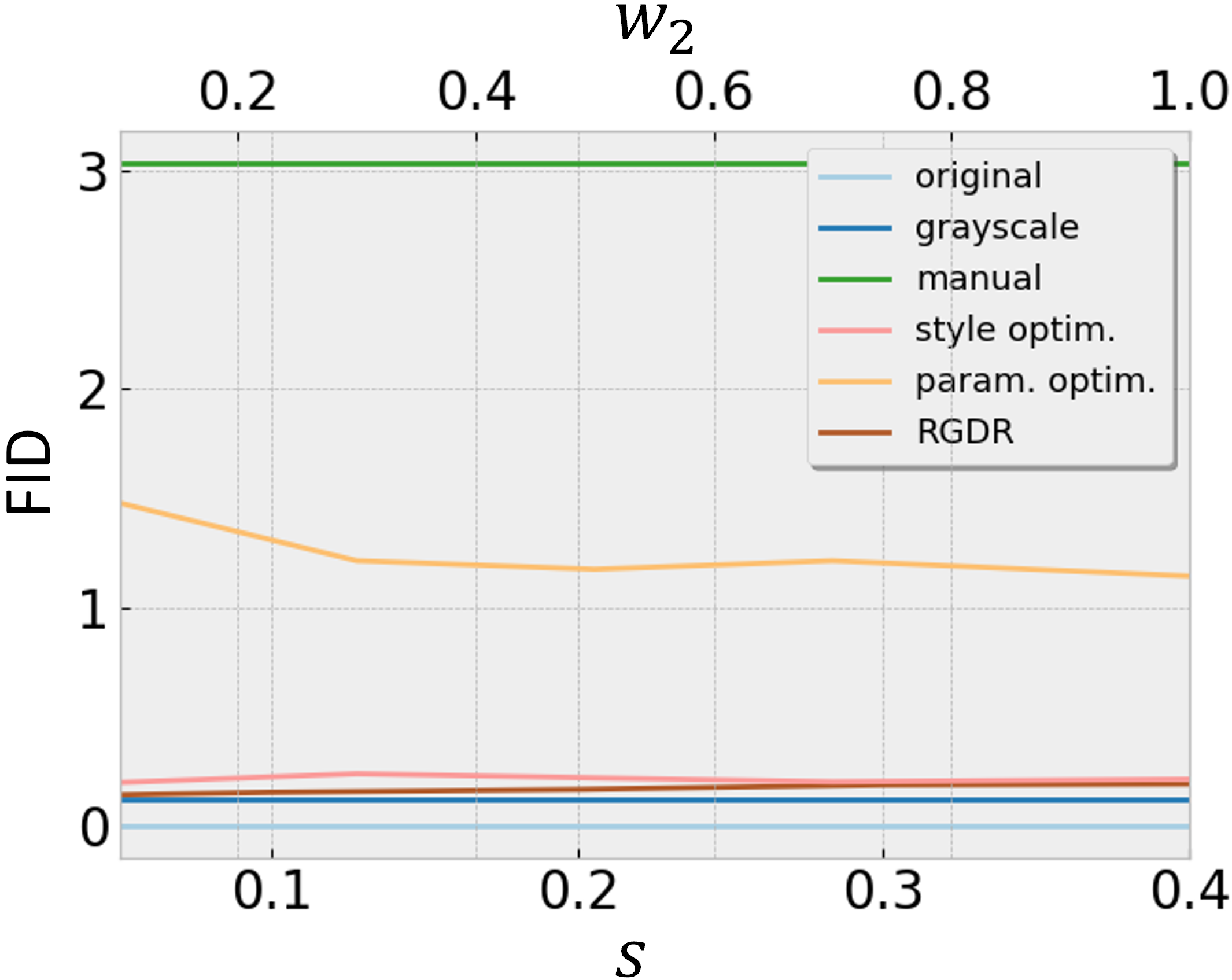}
        \caption{FID}
        \label{fig:criteria_satisfaction:fid}
    \end{subfigure}
    \begin{subfigure}{.33\textwidth}
        \centering
        \includegraphics[width=.95\linewidth]{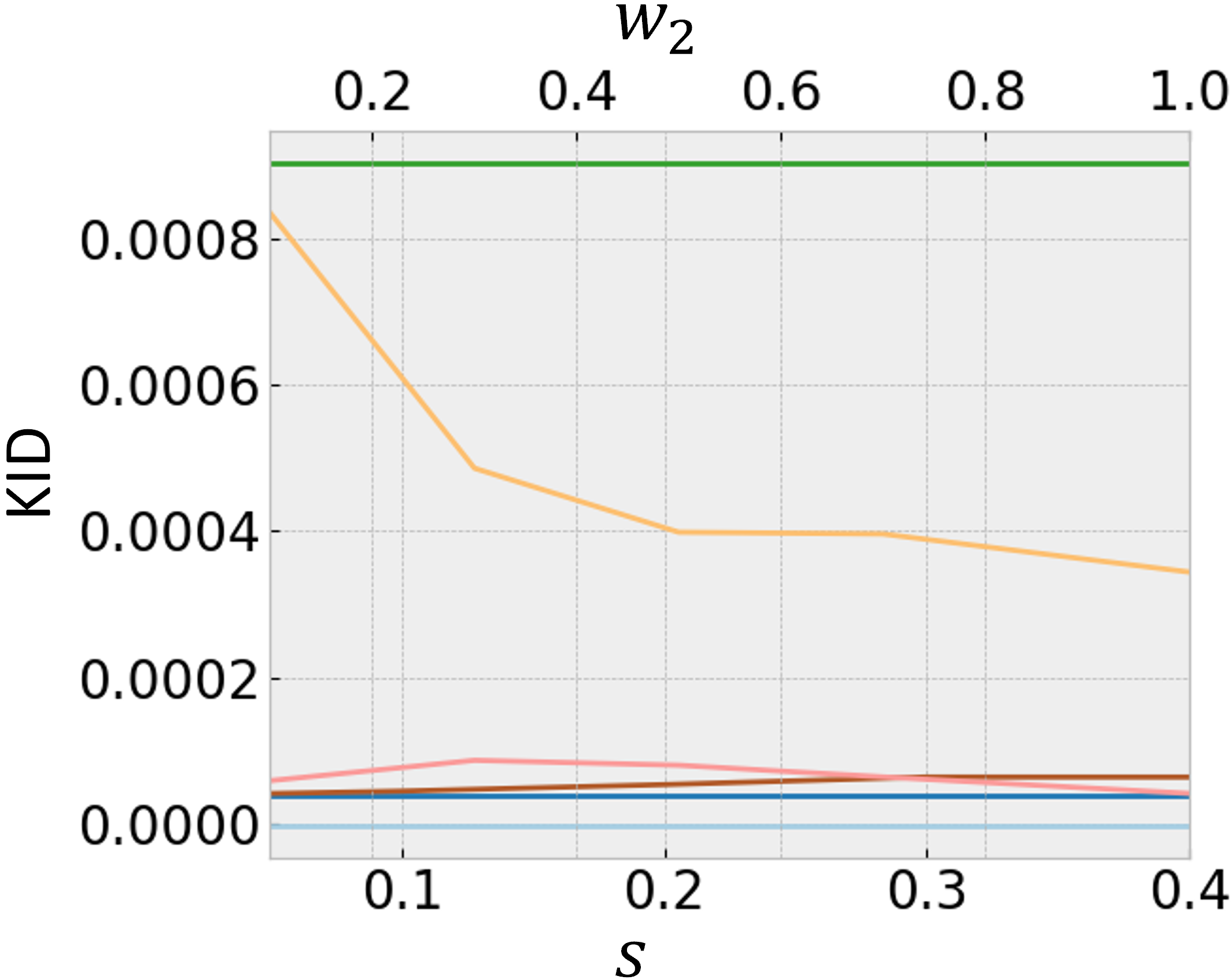}  
        \caption{KID}
        \label{fig:criteria_satisfaction:kid}
    \end{subfigure}
    \begin{subfigure}{.33\textwidth}
        \centering
        \includegraphics[width=.95\linewidth]
        {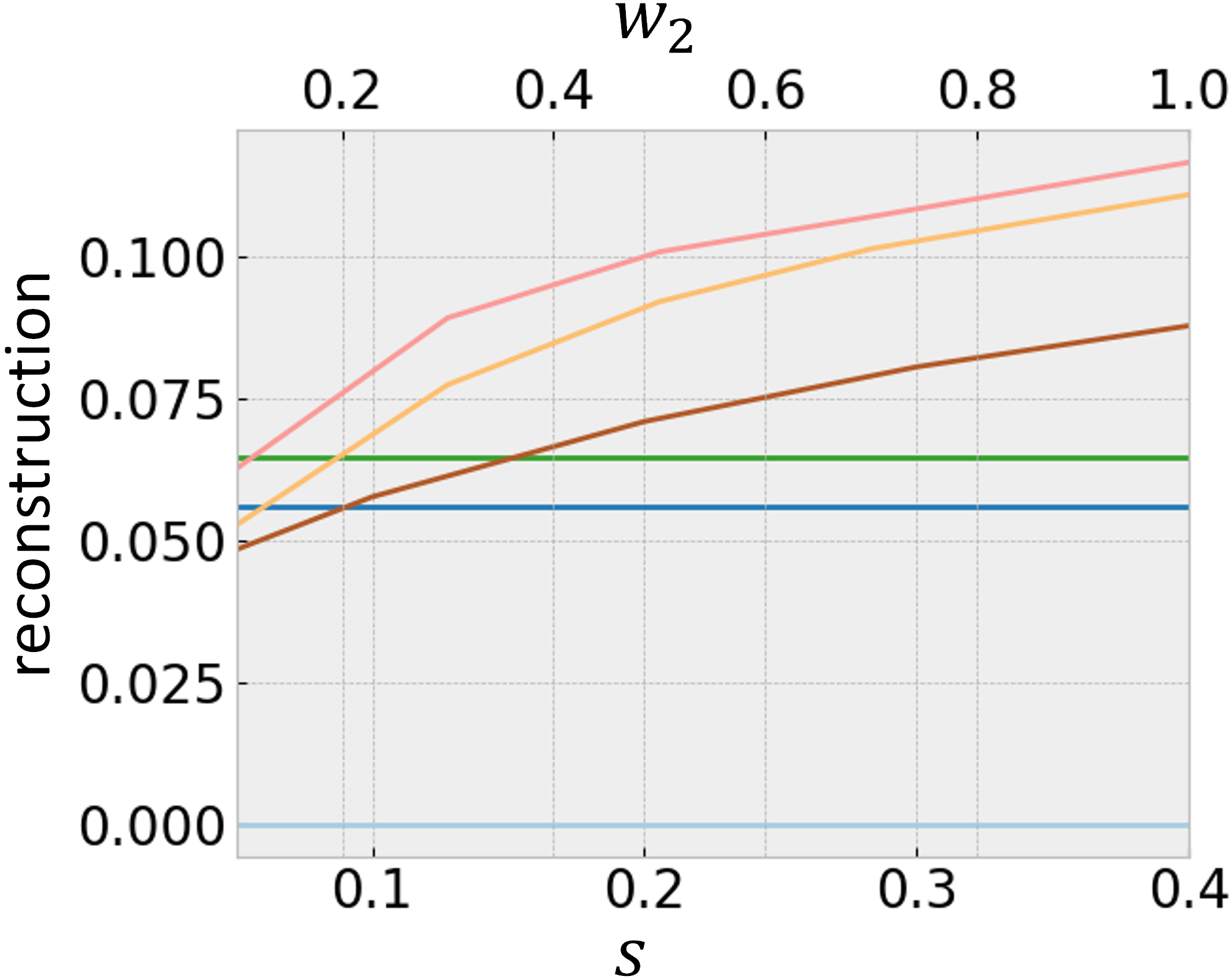}  
        \caption{reconstruction (L1)}
        \label{fig:criteria_satisfaction:l1}
    \end{subfigure}
    \begin{subfigure}{.33\textwidth}
        \centering
        \includegraphics[width=.95\linewidth]{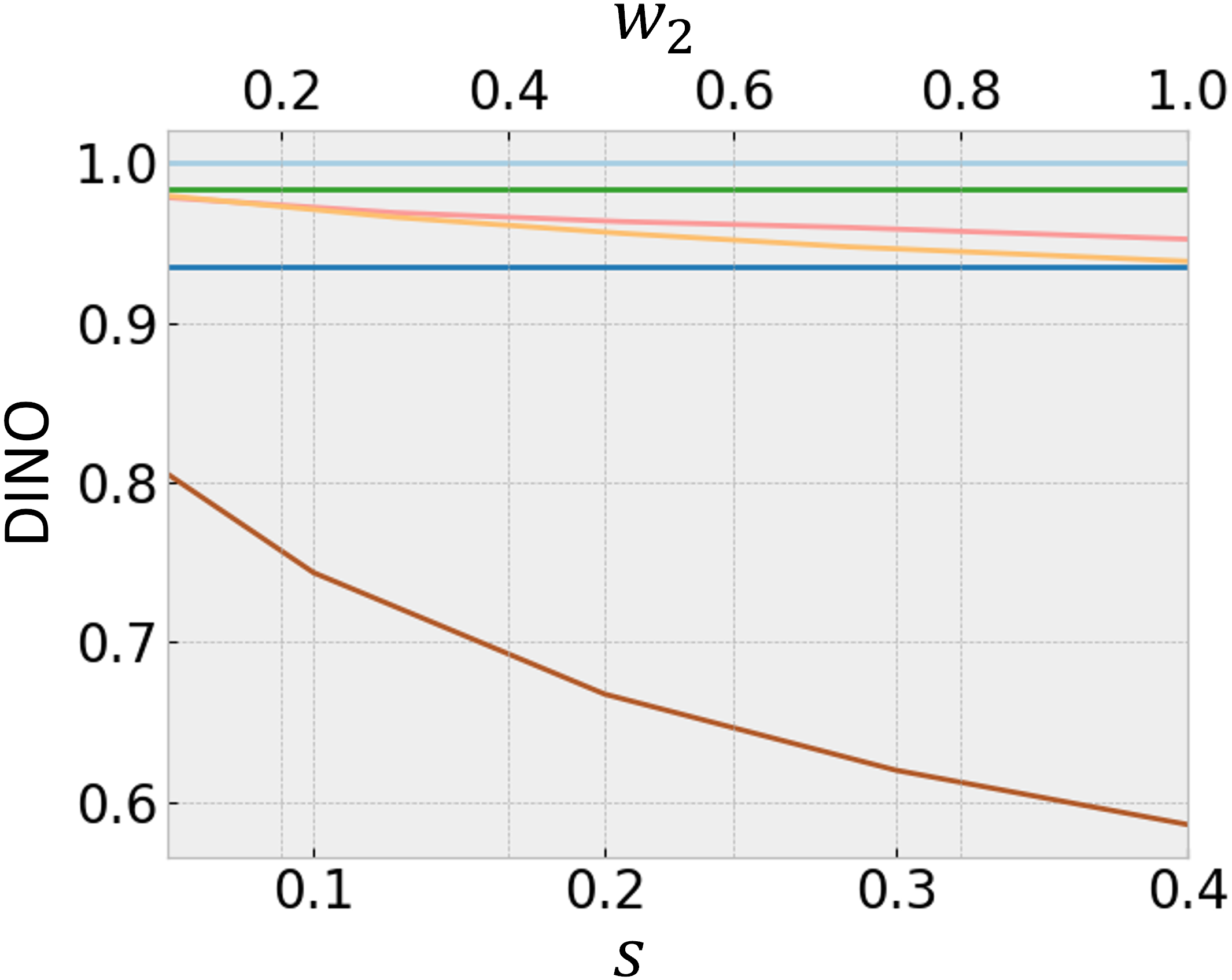}  
        \caption{content preservation (DINO)}
        \label{fig:criteria_satisfaction:dino}
    \end{subfigure}

    \caption{\textbf{Criteria alignment analysis.} The plots show how the metrics evolve as the weighting of the objective terms that introduce emotional changes in the images increases. In each plot, the upper x-axis represents the values of $w_2$ for \param{} and \style{}, while the lower x-axis indicates the values of $s$ for \diff{}. The y-axes show the metric values.}
    
    \Description{Six line plots comparing optimization methods across perceptual and quality metrics. (a) Valence scores remain stable for most methods, with slight decreases for parametric and style optimization. (b) Arousal scores are stable for original and grayscale but decrease for parametric and style optimization. (c) FID is highest for grayscale and parametric optimization, with other methods maintaining lower values. (d) KID decreases for parametric optimization as strength increases, while other methods remain low. (e) Reconstruction error increases with stronger optimization, especially for parametric and style optimization. (f) DINO score decreases for with heigher weights, especially for RGDR.}
    \label{fig:criteria_satisfaction}
\end{figure*}

\paragraph{Criteria alignment}
Figure~\ref{fig:criteria_satisfaction} illustrates how the metrics evolve as the weights of the emotional adaptation objective are adjusted.
In line with our criteria, all approaches move valence toward the neutral midpoint (Fig. \ref{fig:criteria_satisfaction:valence}; \crit{1}) and lower arousal (Fig. \ref{fig:criteria_satisfaction:arousal}; \crit{2}) relative to the original images. 
The \man{} baseline is the exception: it reduces valence but fails to reliably lower arousal, indicating that low-level adaptation benefits from per-image parameter optimization rather than static adjustments.

We confirmed this by fitting ordinary least squares (OLS) regression models to the differences between the valence and arousal values of the original images and their adapted counterparts, for each weight configuration of \param{}, \style{}, and \diff{}.
All models were statistically significant ($p < 0.03$; full terms and $R^2$ in App.~\ref{apx:ols}), indicating that higher weights on the emotional adaptation objective are associated with reductions in both valence and arousal. The negative relationship for valence reflects the original images having a positive average valence, with increasing weight driving them toward the \emref{}; the arousal relationship is directionally consistent with the \emref{}. 

Low KID (Fig. \ref{fig:criteria_satisfaction:kid}) and FID scores (Fig. \ref{fig:criteria_satisfaction:fid}) across baselines and weights confirm that the adapted images maintain high visual quality (\crit{3}).
In terms of \crit{4}, two opposite patterns emerge. Style and \param{} show higher L1 reconstruction error (Fig. \ref{fig:criteria_satisfaction:l1}) than \diff{}, indicating larger photometric change, while their DINO score (Fig. \ref{fig:criteria_satisfaction:dino}) is higher than that of the diffusion-based approach, indicating smaller semantic change. 
Both patterns strengthen as more weight is placed on the emotional adaptation objective.
This dissociation is mechanistically plausible: \style{} and \param{} achieve emotional modulation through global colour-map transformations that leave semantics untouched, whereas \diff{} achieves it through semantic change itself.

\begin{figure*}[h!]%[thb]
    \centering
    \begin{subfigure}{.21\textwidth}
        \centering
        \includegraphics[width=.95\linewidth]
        {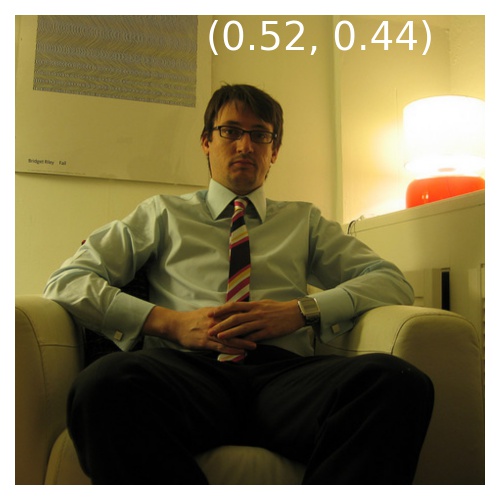}  
        \caption{original}
        \label{fig:coco_results_short:original}
    \end{subfigure}
    \begin{subfigure}{.21\textwidth}
        \centering
        \includegraphics[width=.95\linewidth]
        {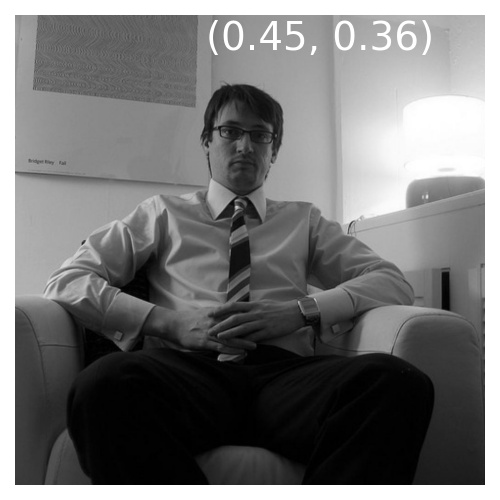}  
        \caption{grayscale}
        \label{fig:coco_results_short:grey}
    \end{subfigure}
    \begin{subfigure}{.21\textwidth}
        \centering
        \includegraphics[width=.95\linewidth]
        {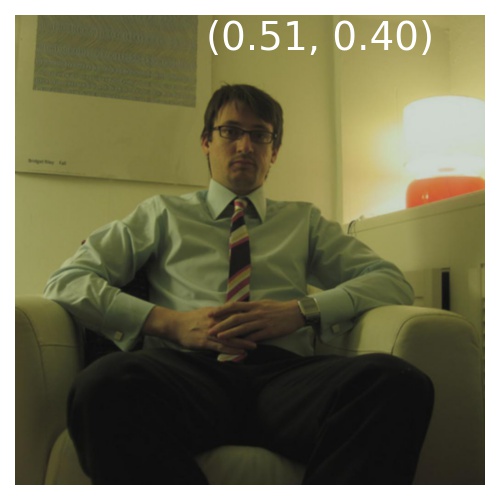}  
        \caption{manual}
        \label{fig:coco_results_short:manual}
    \end{subfigure}
    \begin{subfigure}{1.0\textwidth}
        \centering
        \includegraphics[width=.95\linewidth]
        {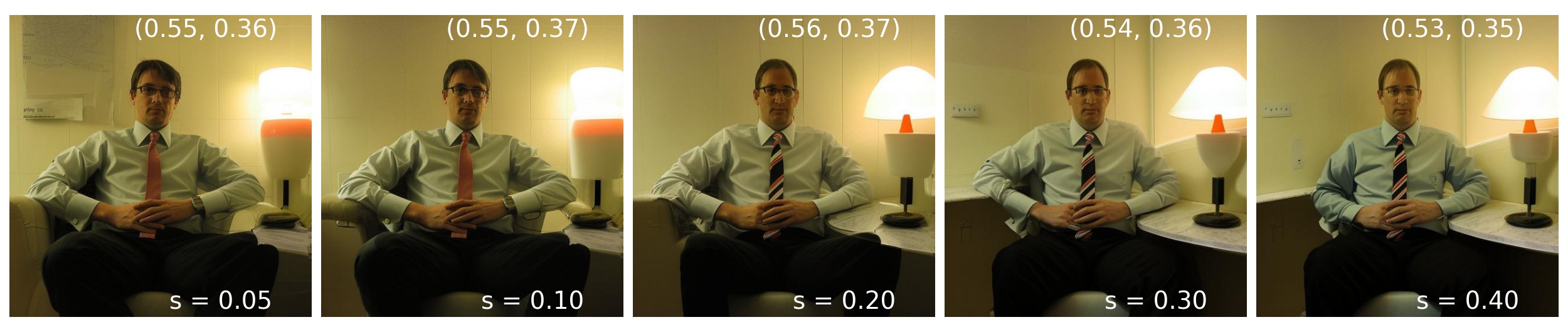}  
        \caption{RGDR, with increasing $s$ from left to right.}
        \label{fig:coco_results_short:diff}
    \end{subfigure}
    \begin{subfigure}{1.0\textwidth}
        \centering
        \includegraphics[width=.95\linewidth]
        {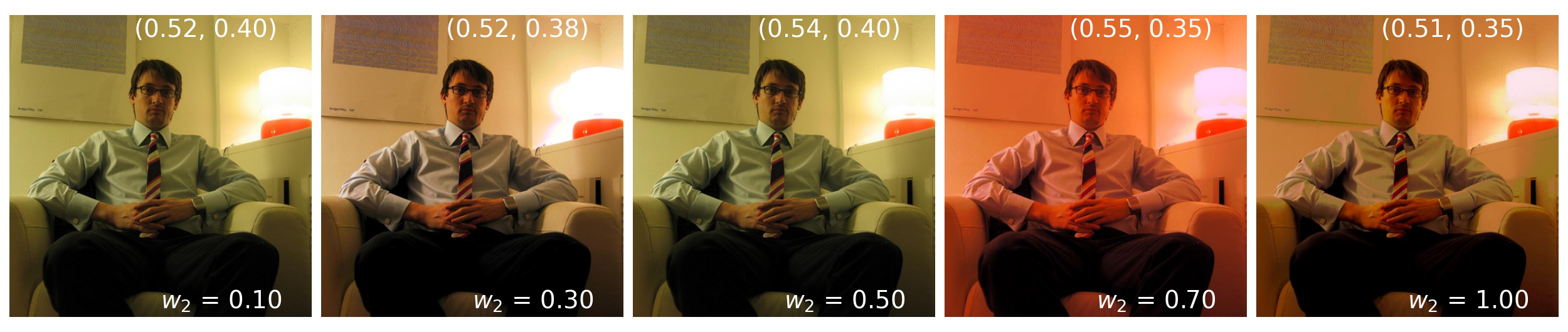}  
        \caption{parametric optimization, with increasing $w_2$ from left to right.}
        \label{fig:coco_results_short:param}
    \end{subfigure}
    \begin{subfigure}{1.0\textwidth}
        \centering
        \includegraphics[width=.95\linewidth]
        {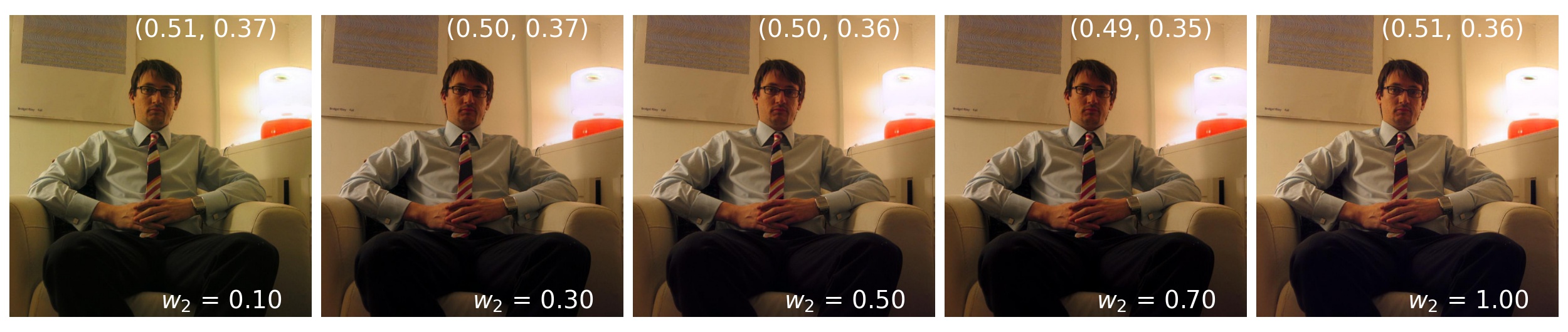}  
        \caption{style optimization, with increasing $w_2$ from left to right.}
        \label{fig:coco_results_short:style}
    \end{subfigure}
    \caption{
    \textbf{Example result}. Predicted valence–arousal values are shown top-right [target: (valence = 0.5, arousal = 0.0)].
    }
    % Image caption: "A man in a shirt and tie sitting on a white chair next to a lamp." 
    \Description{Comparison of image editing and optimization methods applied to the same portrait. Panel (a--c) shows original, grayscale, and manual. Panel (d) shows results of RGDR with increasing parameter $s$, producing gradual changes in the age of the subject. Panels (e--f) show parametric and style optimization with increasing weight $w_2$, resulting in progressive color and lighting adjustments.}
    \label{fig:coco_results_short}
\end{figure*}

\paragraph{Qualitative findings}
Figure~\ref{fig:coco_results_short} shows examples of how the different approaches adapt images toward the desired effect. \diff{} alters not only color but also content, producing plastic transformations, e.g., increasing apparent age (Fig.~\ref{fig:coco_results_short:diff}).
In contrast, style and \param{} primarily alter color values of images, introducing cooler tones (Figure \ref{fig:coco_results_short:style}) or a red tint (Figure \ref{fig:coco_results_short:param}).
Similar changes are affected by \man{} (Figure \ref{fig:coco_results_short:manual}),
Image property-level changes were partly directionally consistent with the psychological literature (App.~\ref{apx:image-property-analysis}); further qualitative examples are provided in the Supplementary Material.

\paragraph{Summary}
All three approaches modify image properties in the direction of the \emref{} while maintaining technical image quality, but through different mechanisms: style and \param{} through low-level photometric adjustment, \diff{} through semantic modification. 
Ablations show that neither classifier guidance nor text conditioning alone matches \diff{}: CG introduces artifacts and higher reconstruction error, while text-only variants fail to reach the \emref{} (App.~\ref{app:diffusion-ablation}). Diffusion approaches using CG occasionally introduced artifacts, due to CG gradients conflicting with CFG (Sec.~\ref{sec:limitations-method}).
Further technical analyses are reported in App.~\ref{apx:technical-evaluation}. To examine whether
these effects translate into perceptual outcomes, we evaluate the adapted images in a controlled image-rating experiment.

\section{Image Rating Study}
\label{sec:image-study}
This study evaluates whether our approaches shift perceived valence toward \emref{} and reduce perceived arousal while preserving perceived image quality (\rques{1}, \crit{1}--\crit{3}). 
We compare the adapted images against the original images and the grayscale baseline.

\subsection{Study design}
\label{sec:image-study-design}
% In the following, we detail conditions, stimuli, task and participants of the experiment.

\paragraph{Conditions}
Following the experimental setup of preceding studies \cite{mould2012emotional, besanccon2018reducing}, the study contains one within-subject factor, \ivfilter{}, with 5 factor levels: \cdefault{}, \cgray{}, \cdiff{}, \cstyle{}, \cparam{}.
With each adaptation approach, we adapted the same block of images (see \emph{Stimuli}).
To avoid order effects, we used latin-square counterbalancing to adjust the order of conditions between participants. 

Adaptation weights were selected by the authors from the trade-off curves of the technical evaluation, choosing for each approach the weight that moved images toward the \emref{} without artifacts. 
% We then verified these weights qualitatively on the stimuli used here and adjusted where needed, as no single metric captures both objectives.
The weights were \diff{}: $w = 2.0$, $s = 0.2$; \style{}: $w_1 = 1.0$, $w_2 = 0.2$; \param{}: $w_1 = 1.0$, $w_2 = 0.15$.

\paragraph{Stimuli}
The stimuli stem from the Nencki Affective Picture System (NAPS)~\cite{marchewka2014nencki} and its erotic subset NAPS ERO~\cite{wierzba2015erotic}, two databases of realistic, high-quality photographs with normative valence and arousal ratings. 
We selected images from the people, faces, animals, and landscapes categories, as these represent genres frequently seen on social media~\cite{leder2022swipes}. 

Following Mould et al.~\cite{mould2012emotional}, we sampled from four regions of the valence--arousal space. 
This yielded 12 stimuli, plus 5 calibration images sampled across the full space. The 12 images form a block presented in each condition, ordered to begin and end with lower-arousal, positive images. Cluster definitions and ordering are detailed in Appendix~\ref{app:rating-stimuli}.
NAPS usage terms prohibit reproducing images; stimulus IDs are given in Appendix~\ref{apx:im-naps-ids}.

\paragraph{Task and procedure}
Participants viewed each image for a minimum of five seconds, after which they could continue viewing or advance by pressing the space key. They then rated valence and arousal on a validated questionnaire~\cite{bradley1994measuring} and image quality on a Likert item adapted from~\cite{mould2012emotional}. A neutral stimulus followed each image--questionnaire pair to prevent emotional contamination. The task repeated for all 12 images in each condition.
Participant instructions are shown in Appendix~\ref{apx:im-task}.

% \paragraph{Procedure}
Participants provided consent, completed a demographic questionnaire, and familiarized themselves with the setup in a trial run. After a two-minute relaxation video, they proceeded through the calibration images and the five condition blocks. Sessions lasted approximately 30 minutes and were compensated with £4.50.

\paragraph{Participants}
Based on effect sizes in similar studies~\cite{de2010effects, bekhtereva2017bringing}, an a priori power analysis established a minimum sample of $N = 48$ (Apx.~\ref{apx:im-power}). We recruited 59 participants (33 female, 26 male), ages 18--74 (M=36, SD=11.5) via Prolific and excluded three whose responses indicated inattentive completion (e.g., identical ratings across all images), yielding 
$N = 56$. 
Importantly, results remain unchanged regardless of whether these outliers were included or not.

\paragraph{Statistical analysis}
We conducted repeated-measures ANOVAs on each dependent variable.
As Mauchly's test indicated sphericity violations for all measures (all \pvall{<}{.003}), we are using Greenhouse--Geisser corrected degrees of freedom throughout; uncorrected and corrected ANOVAs never differed in significance.

For the posthoc comparisons, we tested the normality of the paired differences (Shapiro--Wilk); as 18 of 26 distributions violated normality, we computed Wilcoxon signed-rank tests, applying Holm correction throughout to control the familywise error rate. Since the non-parametric tests never diverged in significance from the paired $t$-tests, and are more robust to the observed non-normality, we report Holm-corrected Wilcoxon $p$-values throughout.

\subsection{Results}
\begin{figure*}[thb]
    \centering
        \begin{subfigure}{.34\textwidth}
        \centering
        \includegraphics[width=.95\linewidth]{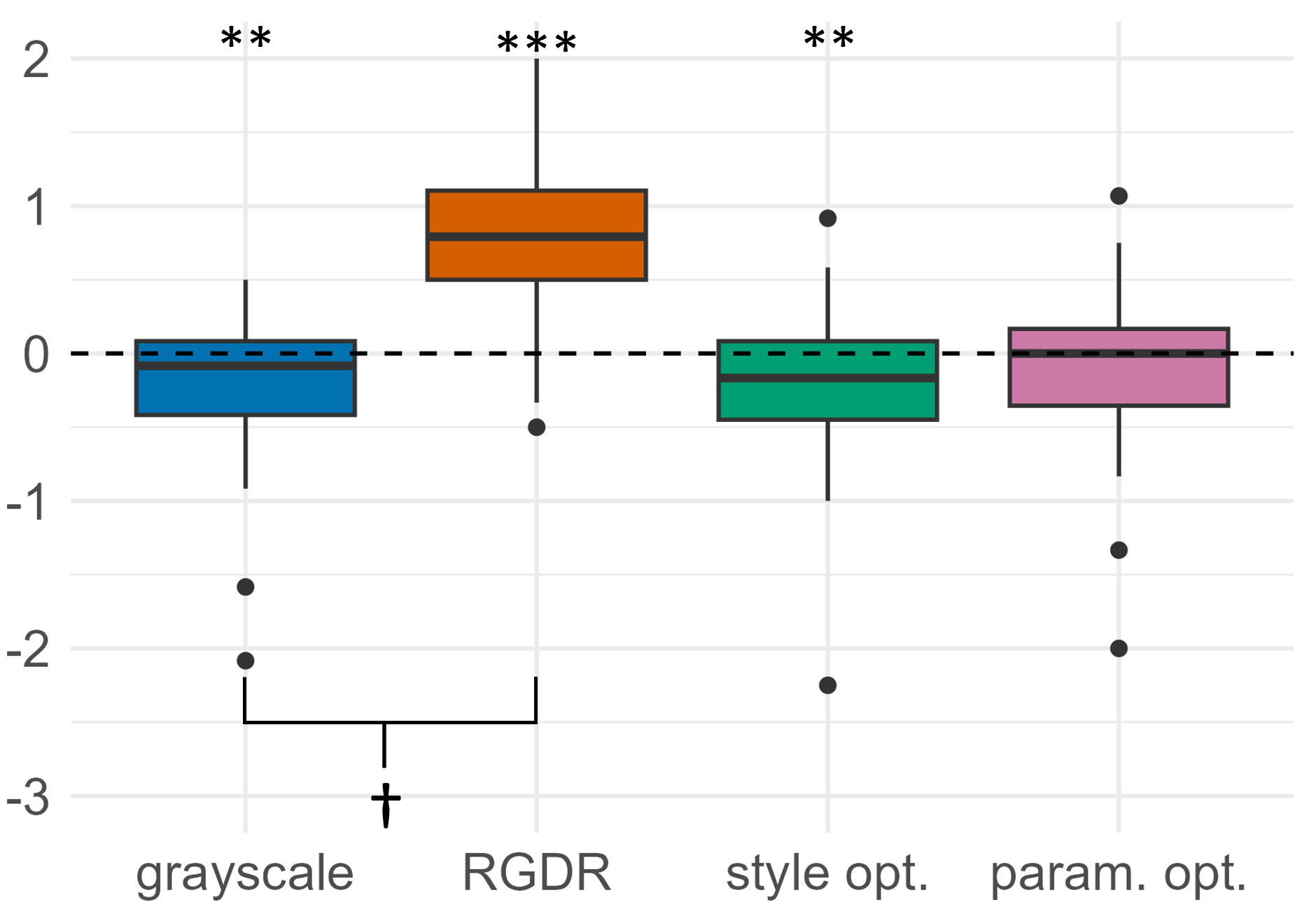}  
        \caption{valence}
    \end{subfigure}
    \hspace{-4mm}
    \begin{subfigure}{.34\textwidth}
        \centering
        \includegraphics[width=.95\linewidth]{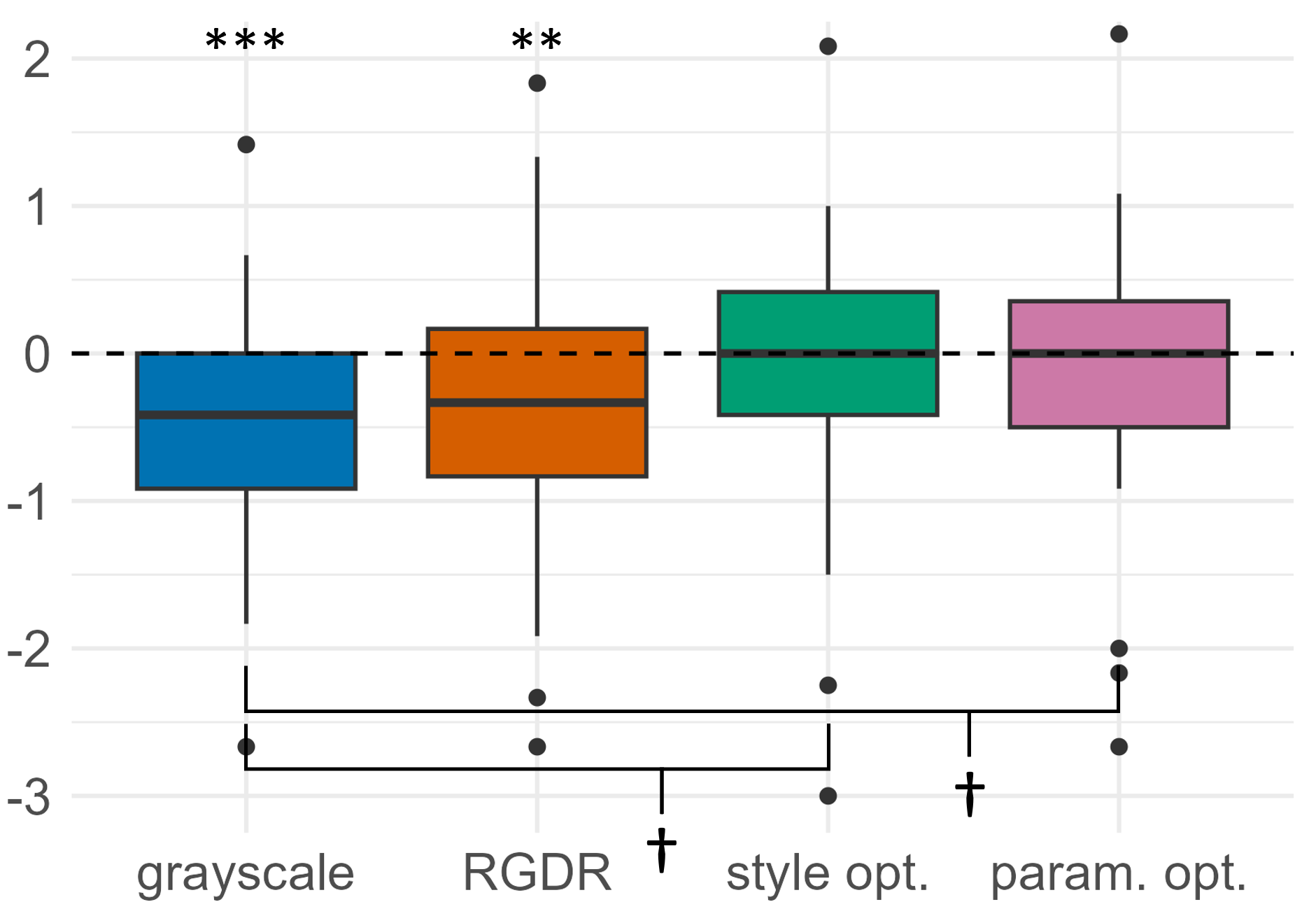}
        \caption{arousal}
    \end{subfigure}
    \hspace{-4mm}
    \begin{subfigure}{.34\textwidth}
        \centering
        \includegraphics[width=.95\linewidth]{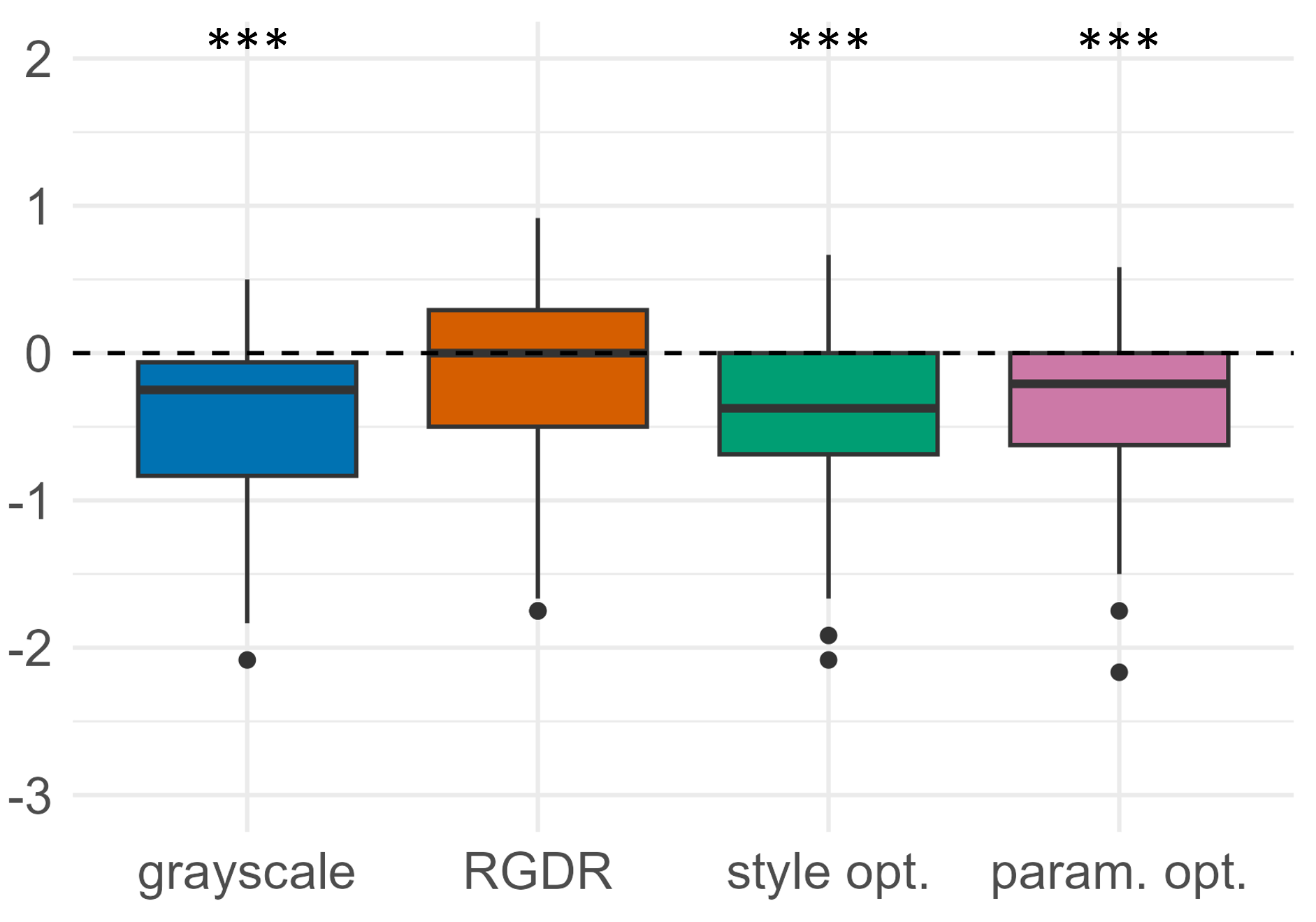}  
        \caption{image quality}
    \end{subfigure}
    \caption{\textbf{Results of image rating study.} Per-participant rating differences between each condition and \cdefault{} for valence (a), arousal (b), and image quality (c). Each box summarizes one difference score per participant; values below the dashed zero line indicate lower ratings than for the original images. 
    Asterisks mark paired $t$-tests against \cdefault{}; double daggers, arousal comparisons where pairwise differences violate normality significant only under Wilcoxon; daggers, exploratory differences from \cgray{} ($^{*}p<.05$, $^{**}p<.01$, $^{***}p<.001$, $^{\dagger}p<.001$). All Holm-corrected.
    The box represents the interquartile range (IQR; Q1–Q3), the central line marks the median, and the whiskers extend to the smallest and largest values within 1.5 × IQR, with points beyond plotted as outliers. The y-axes show Likert response levels of the respective item.}
    \Description{Boxplots comparing five methods (grayscale, RGDR, style optimization, parametric optimization) across three measures. Panel (a) shows valence ratings, with RGDR rated highest and grayscale lowest. Panel (b) shows arousal ratings, where grayscale and RGDR. Panel (c) shows image quality, where grayscale, style and parametric optimization are rated low, while RGDR is higher. Statistical significance is indicated with brackets, stars, and daggers.}
    \label{fig:boxplot_calibration_study} 
\end{figure*}

% In the following, we present the results of the study.
Figure \ref{fig:boxplot_calibration_study} shows an overview of the study results.
Descriptive statistics for all measures are reported in Appendix~\ref{apx:im-descriptive}.

We found an effect of the \ivfilter{} factor levels on valence ratings \anova{2.80}{153.84}{69.32}{<}{.001}{.56}.
Post-hoc pairwise comparisons showed that \cdiff{} was associated with an increase in valence ratings relative to \cdefault{} \pvald{<}{.001}{1.51}, whereas \cstyle{} \pvald{=}{.009}{-0.42} and \cgray{} \pvald{=}{.018}{-0.43} showed reductions.
No other conditions differed significantly from \cdefault{} in valence.

% Valence results (pes == effect size)
% Anova Table (Type 3 tests)
%           num Df den Df     MSE      F     pes    Pr(>F)    
% condition 2.7971 153.84 0.20484 69.317 0.55758 < 2.2e-16 ***

%   family          contrast                            t    df p         d  d_lo  d_hi shapiro_p wilcoxon_p p_holm wilcoxon_p_holm
% 1 primary_valence valence : grayscale vs original -3.18    55 0.002 -0.43 -0.7  -0.15 <.001     0.009      0.007  0.018          
% 2 primary_valence valence : RGDR vs original      11.3     55 <.001  1.51  1.12  1.89 0.511     <.001      <.001  <.001          
% 3 primary_valence valence : style_opt vs original -3.13    55 0.003 -0.42 -0.69 -0.14 0.002     0.003      0.007  0.009          
% 4 primary_valence valence : param_opt vs original -1.81    55 0.075 -0.24 -0.51  0.02 0.001     0.145      0.075  0.145

In an exploratory analysis, we additionally compared each condition's valence ratings against \cgray{}. 
\diff{} elicited higher valence ratings than \cgray{} \pvald{<}{.001}{1.44}. 
No other condition differed from \cgray{} in valence.

%   family                   contrast                             t    df p         d  d_lo  d_hi shapiro_p wilcoxon_p p_holm wilcoxon_p_holm       
% 1 exploratory_gray_valence valence : RGDR vs grayscale      10.8     55 <.001  1.44  1.06  1.81 <.001     <.001      <.001  <.001          
% 2 exploratory_gray_valence valence : style_opt vs grayscale  0.02    55 0.988  0    -0.26  0.26 <.001     0.444      0.988  0.740          
% 3 exploratory_gray_valence valence : param_opt vs grayscale  1.45    55 0.154  0.19 -0.07  0.46 <.001     0.370      0.307  0.740 

Because the reference point for valence optimization corresponds to the midpoint of the Likert scale (5), rather than zero, we conducted exploratory one-sample t-tests comparing mean valence ratings for each condition against this midpoint. 
Valence ratings for \cdefault{} \ttestd{55}{-5.55}{<}{.001}{-.74}, \cgray{} \ttestd{55}{-8.97}{<}{.001}{-1.2}, \cstyle{} \ttestd{55}{-7.86}{<}{.001}{-1.05}, and \cparam{} \ttestd{55}{-7.39}{<}{.001}{-0.99} differed from 5.
In contrast, only \diff{} did not differ from the neutral midpoint \ttestd{55}{1.91}{=}{.157}{0.25}.

%   family                       contrast                            t    df p         d  d_lo  d_hi shapiro_p wilcoxon_p p_holm wilcoxon_p_holm
% 1 exploratory_midpoint_valence valence : original vs midpoint  -5.55    55 <.001 -0.74 -1.04 -0.44 <.001     <.001      <.001  <.001          
% 2 exploratory_midpoint_valence valence : grayscale vs midpoint -8.97    55 <.001 -1.2  -1.54 -0.85 0.020     <.001      <.001  <.001          
% 3 exploratory_midpoint_valence valence : style_opt vs midpoint -7.86    55 <.001 -1.05 -1.37 -0.72 0.028     <.001      <.001  <.001          
% 4 exploratory_midpoint_valence valence : param_opt vs midpoint -7.39    55 <.001 -0.99 -1.3  -0.66 0.013     <.001      <.001  <.001          
% 5 exploratory_midpoint_valence valence : RGDR vs midpoint       1.91    55 0.062  0.25 -0.01  0.52 <.001     0.157      0.062  0.157

In addition, the ANOVA showed that there was an effect of the \ivfilter{} on arousal ratings \anova{3.24}{178.08}{7.79}{<}{.001}{.12}.
Post-hoc tests showed that \cdiff{} \pvald{=}{.003}{-.45} and \cgray{} \pvald{<}{.001}{-.68} both caused lower arousal ratings than default.
No other arousal differences were observed relative to \cdefault{}.

% Arousal results (pes == effect size)
% Anova Table (Type 3 tests)
%           num Df den Df    MSE      F     pes    Pr(>F)    
% condition 3.2379 178.08 0.4077 7.7935 0.12411 3.905e-05 ***

 %   family          contrast                            t    df p         d  d_lo  d_hi shapiro_p wilcoxon_p p_holm wilcoxon_p_holm             
 % 5 primary_arousal arousal : grayscale vs original -5.11    55 <.001 -0.68 -0.97 -0.39 0.245     <.001      <.001  <.001          
 % 6 primary_arousal arousal : RGDR vs original      -3.39    55 0.001 -0.45 -0.73 -0.18 0.502     0.001      0.004  0.003          
 % 7 primary_arousal arousal : style_opt vs original -0.61    55 0.545 -0.08 -0.34  0.18 0.003     0.788      1.000  1.000          
 % 8 primary_arousal arousal : param_opt vs original -0.52    55 0.606 -0.07 -0.33  0.19 0.004     0.965      1.000  1.000 

Exploratorily, we further conducted pairwise comparisons between conditions and \cgray{}. 
These revealed that \cgray{} elicited lower arousal ratings than both \cstyle{} \pvald{<}{.001}{.56} and \cparam{} \pvald{<}{.001}{.52}, while no differences between \cgray{} and \cdiff{} were found \pval{=}{.623}.

%   family                   contrast                             t    df p         d  d_lo  d_hi shapiro_p wilcoxon_p p_holm wilcoxon_p_holm      
% 4 exploratory_gray_arousal arousal : RGDR vs grayscale       0.64    55 0.526  0.09 -0.18  0.35 0.727     0.623      0.526  0.623          
% 5 exploratory_gray_arousal arousal : style_opt vs grayscale  4.19    55 <.001  0.56  0.28  0.84 <.001     <.001      <.001  <.001          
% 6 exploratory_gray_arousal arousal : param_opt vs grayscale  3.93    55 <.001  0.52  0.24  0.8  0.007     <.001      <.001  <.001

The ANOVA also revealed an effect of \ivfilter{} on perceived image quality \anova{2.73}{150.19}{10.04}{<}{.001}{.15}.
Pairwise comparisons showed that \cgray{} \pvald{<}{.001}{-.71}, \cparam{} \pvald{<}{.001}{-.64}, and \cstyle{} \pvald{<}{.001}{-.72} were associated with lower image quality ratings than \cdefault{}. 
In contrast, \cdiff{} did not differ from \cdefault{} \pval{=}{.298}.
%
% Image quality results (pes == effect size)
% Anova Table (Type 3 tests)
%           num Df den Df     MSE      F     pes    Pr(>F)    
% condition 2.7307 150.19 0.29714 10.039 0.15435 9.646e-06 ***
% family          contrast                            t    df p         d  d_lo  d_hi shapiro_p wilcoxon_p p_holm wilcoxon_p_holm         
%  9 primary_quality quality : grayscale vs original -5.35    55 <.001 -0.71 -1.01 -0.42 0.003     <.001      <.001  <.001          
% 10 primary_quality quality : RGDR vs original      -1.75    55 0.085 -0.23 -0.5   0.03 0.001     0.298      0.085  0.298          
% 11 primary_quality quality : style_opt vs original -5.43    55 <.001 -0.72 -1.02 -0.43 0.072     <.001      <.001  <.001          
% 12 primary_quality quality : param_opt vs original -4.8     55 <.001 -0.64 -0.93 -0.35 0.014     <.001      <.001  <.001
%
In the posthoc comparisons between conditions and \cgray{}, no differences were found.
%   family                   contrast                             t    df p         d  d_lo  d_hi shapiro_p wilcoxon_p p_holm wilcoxon_p_holm         
% 7 exploratory_gray_quality quality : RGDR vs grayscale       2.31    55 0.025  0.31  0.04  0.57 0.558     0.017      0.075  0.051          
% 8 exploratory_gray_quality quality : style_opt vs grayscale  0.01    55 0.990  0    -0.26  0.26 0.659     0.954      1.000  0.954          
% 9 exploratory_gray_quality quality : param_opt vs grayscale  0.49    55 0.629  0.06 -0.2   0.33 0.303     0.434      1.000  0.867

% To confirm our results are not driven by a subset of the 12 stimuli, we repeated the primary contrasts with image, rather than participant, as the unit of analysis in Appendix~\ref{apx:im-rating-by-image}.

\subsection{Discussion}
\label{sec:image-rating-discussion}
The ability to shift perceived emotion toward \emref{} without degrading perceived quality did not emerge gradually across the spectrum: it appeared only at the high-abstraction end.

At the low-abstraction end, \cgray{} reduced arousal substantially, consistent with prior work showing colored images are more arousing than grayscale ones \cite{bekhtereva2017bringing,cano2009affective}. Its effect on valence follows a distinct mechanism: since color amplifies valence in whichever direction it already points~\cite{bekhtereva2017bringing,cano2009affective}, and original images had slightly negative average valence, removing color should sharpen that negativity. Consistently, grayscale images were rated furthest of any condition from the neutral midpoint.

The low-level adaptations \cparam{} and \cstyle{} reduced perceived quality without reliably reducing arousal, and likewise did not approach the \emref{} in valence. 
This is notable, since both approaches optimize, one implicitly, the other explicitly, image properties that prior work associates with changing emotional responses (Sec.~\ref{sec:image-properties}).
Their per-image parameter optimization was technically effective (Sec.~\ref{sec:technical-evaluation}), yet the resulting changes were evidently too subtle to register perceptually.
While not effective at shifting perceived emotion, the photometric changes induced by these mechanisms remained visible enough to be judged as quality degradation.
% (as identified in the technical evaluation)

Only \cdiff{}, at the high-abstraction end, satisfied all three: it was the sole condition whose valence ratings did not differ from the neutral midpoint, it reduced arousal relative to \cdefault{} without differing from \cgray{}, and it was the only adapted condition whose perceived quality did not differ significantly from \cdefault{}. This pattern is consistent with the results of the technical evaluation: \cdiff{} appears to achieve emotional modulation through semantic change rather than the pixel-level photometric change that degraded perceived quality for \cparam{} and \cstyle{} 

We therefore judge \cdiff{} to be the only approach that addresses \rques{1}, and, hence, carry it forward to the social media study, excluding \cparam{} and \cstyle{} as they did not satisfy the required criteria. 
Note that \cdiff{} advances despite a known cost of its own: the technical evaluation showed that its emotional modulation is achieved through semantic change, at a measured cost to fidelity to the original image (\crit{4}). 
Because participants in the following study view only the adapted feed and never the originals, this cost cannot affect their in-feed experience directly.
However, it is central to the ethical standing of the intervention, and we discuss it in Section~\ref{sec:ethics}.

\section{Social Media Study}
\label{sec:social-study}
This study examines whether emotionally adapted images shift users' experienced emotion and, through that, reduce the time they spend browsing a social media feed (\rques{2}). 
% Collecting self-reported valence and arousal alongside browsing duration from the same participants lets us test both links of this chain: whether the adaptations shift perceived emotion, and whether browsing behavior changes accordingly.
% We compare the adapted images of \cdiff{} against the original images and the grayscale filter.

\subsection{Study design}

\paragraph{Conditions}
The study followed a between-subjects design with a single factor, \ivfilter{}, with 3 factor levels: \cdefault{}, \cgray{}, \cdiff{}.
All images in the first feed were adapted according to the assigned level (see \emph{Task and procedure}).
The weights for \diff{} were the same as in the image rating study: $w = 2.0$, $s = 0.2$.

\paragraph{Stimuli}
We used real social media posts, including both text and images, as stimuli, drawn from the Instagram Influencer Dataset~\cite{kim2020multimodal}. We sampled 812 images from the fitness and travel categories, chosen to match the preferences of a broader audience. 
Two screening passes reduced this set: the first discarded images for which none of eight \diff{} generations was artifact-free, the second retained only images in which the adaptation had introduced a visible change relative to the original.
This yielded 268 images, used in the first feed across all conditions.
The stimulus set therefore represents the subset of content on which the adaptation is effective, rather than a random sample. App.~\ref{app:sm-stimuli} details the pipeline and the feed-ordering procedure.

\paragraph{Task and procedure}
After providing informed consent and completing a demographic questionnaire, participants received a brief study introduction and completed a one-minute practice task featuring comic animal images instead of social media content. A two-minute relaxation video preceded the main task.

Participants then browsed two curated feeds in sequence: the first contained images adapted according to their assigned condition, the second unadapted images. They began in the first feed, and after 30 seconds a button to access the second feed became available. Clicking it triggered a questionnaire comprising the same Likert-scale items used in the image rating study (perceived arousal, valence, and image quality), together with comprehension and attention checks, after which participants proceeded to the second feed. Total browsing time was fixed at five minutes across both feeds, so participants who remained in the first feed for the full duration never saw the second. 
Therefore, time spent in the first feed, our primary dependent variable, could range from 30 to 300 seconds. Because the total duration was fixed, leaving the first feed was a choice between browsing options rather than a way to end the study sooner, and switching was most plausibly motivated by dissatisfaction with the current content.

The instructions shown to participants are reproduced in Appendix~\ref{apx:sm-task}. Appendix~\ref{apx:sm-feed} illustrates the feed interface. 
Upon completion, participants received a code to claim a £2.25 reward. The study took approximately 15 minutes.

\paragraph{Participants}
An a priori power analysis based on effect sizes of similar studies \cite{de2010effects, bekhtereva2017bringing} indicated a required sample size of N = 192 participants (App.~\ref{apx:sm-power}).
We recruited participants via Prolific, prescreening for active social media users. Of 216 people recruited, 192 passed the attention checks; four were excluded for not interacting with the feed or due to a technical issue, and six could not be linked to their questionnaire responses due to token-entry errors, yielding an analysis sample of $N = 182$ (\cdefault{}: 60, \cgray{}: 59, \cdiff{}: 63). 
Appendix~\ref{apx:sm-analysis-sample} details the full derivation and robustness checks across sample definitions, all of which left conclusions unchanged.

The final sample consisted of 182 individuals (88 female, 93 male, 1 non-binary), aged 19–75 years (M = 39, SD = 11.7). On average, participants reported spending 3.6 hours per day on social media (SD = 2.9; range = 0–20).
% We did not assess participants' interest in the two content categories used as stimuli (fitness and travel).
%
Data collection proceeded in two stages; a comparability analysis found no significant differences on measured pre-treatment characteristics across stages (App.~\ref{apx:sm-comparability}).

\paragraph{Statistical analysis}
For significance testing, we conducted one-way between-subjects ANOVAs, as specified in the preregistration [anonymized] (App.~\ref{apx:sm-deviations} documents deviations from the preregistration).
Levene's test indicated variance homogeneity for all dependent variables (all \pvall{>}{.15}); we nonetheless report the more robust Welch's $F$. Uncorrected and corrected ANOVAs never differed in significance.

The preregistered analysis plan specified two-tailed independent-samples $t$-tests for pairwise comparisons against \cdefault{}; however, residual normality was violated for all four dependent variables (Shapiro-Wilk; all \pvall{<}{.003}), so we instead computed the more robust Wilcoxon signed-rank tests, applying Holm correction throughout to control the familywise error rate.
We report Cohen's $d$ as the effect size for all posthoc comparisons.
Welch's $t$-test, Holm-corrected Student's $t$-test, and the Holm-corrected non-parametric test agreed on all comparisons but one, which we report below.

\subsection{Results}
% We first report the statistical analysis of the preregistered hypotheses, followed by an exploratory analysis of the temporal pattern of disengagement and qualitative observations on the types of edits introduced to social media images.

% Confirmatory results
\subsubsection{Preregistered analyses}
\label{sec:social-study:registered}
Descriptive statistics for all measures are reported in Appendix~\ref{apx:sm-descreptive-stats}.
We first tested whether the adaptations shifted perceived emotion, the precondition for the mediation \rques{2} posits.
Self-reported valence did not differ across conditions \anova{2.0}{117.4}{.65}{=}{.524}{.008}.
For arousal, a Welch ANOVA indicated an omnibus effect \anova{2.0}{118.1}{4.63}{=}{.012}{.046}, but no pairwise contrast against \cdefault{} survived post-hoc correction (\cgray{}, \pvall{=}{.106}; \cdiff{}, \pvall{=}{.423}).
An exploratory comparison between the two adapted conditions showed that \cdiff{} was rated more arousing than \cgray{} \pvald{=}{.034}{0.55}. 

We found no main effect of \ivfilter{} on time spent in the first feed \anova{2.0}{114.1}{.13}{=}{.874}{.001}, nor on perceived image quality \anova{2.0}{117.2}{2.71}{=}{.071}{.026}.
% Four preregistered exploratory behavioral measures, median scroll speed, median per-post viewing time, number of posts viewed, and likes, likewise showed no differences by condition (Apx.~\ref{apx:sm-behaviroal-features}).
As neither the mediator nor the outcome variable differed between the adapted conditions and \cdefault{}, we did not conduct a mediation analysis.

% DURATION: Welch ANOVA: F(2.0, 114.1) = 0.13, p = 0.8741, pes = 0.001
% VALENCE: Welch ANOVA: F(2.0, 117.4) = 0.65, p = 0.5237, pes = 0.008
% QUALITY: Welch ANOVA: F(2.0, 117.2) = 2.71, p = 0.0708, pes = 0.026
% AROUSAL: Welch ANOVA: F(2.0, 118.1) = 4.63, p = 0.0116, pes = 0.046

%   family          contrast                            t    df p     p_registered_student     d  d_lo  d_hi wilcoxon_p p_holm wilcoxon_p_holm        
% 1 planned_arousal arousal : grayscale vs original -1.95  116. 0.053               0.0533 -0.36 -0.72  0.01 0.081      0.106  0.162          
% 2 planned_arousal arousal : RGDR vs original       0.8   116. 0.423               0.421   0.15 -0.21  0.5  0.470      0.423  0.470  

%   family      contrast                        t    df p     p_registered_student     d  d_lo  d_hi shapiro_p_a shapiro_p_b wilcoxon_p p_holm wilcoxon_p_holm
% 3 exploratory arousal: RGDR vs grayscale   3     118. 0.003              0.00318  0.55  0.18  0.91 0.019       0.003       0.008      0.013  0.034

\begin{figure*}[thb]
    \centering
        \begin{subfigure}{.54\textwidth}
        \centering
        % \includegraphics[width=.95\linewidth]{figures/plots_social_media_study/duration_histogram.png}  
        % \caption{distribution of duration in first feed (in seconds)}
        \includegraphics[width=.95\linewidth]{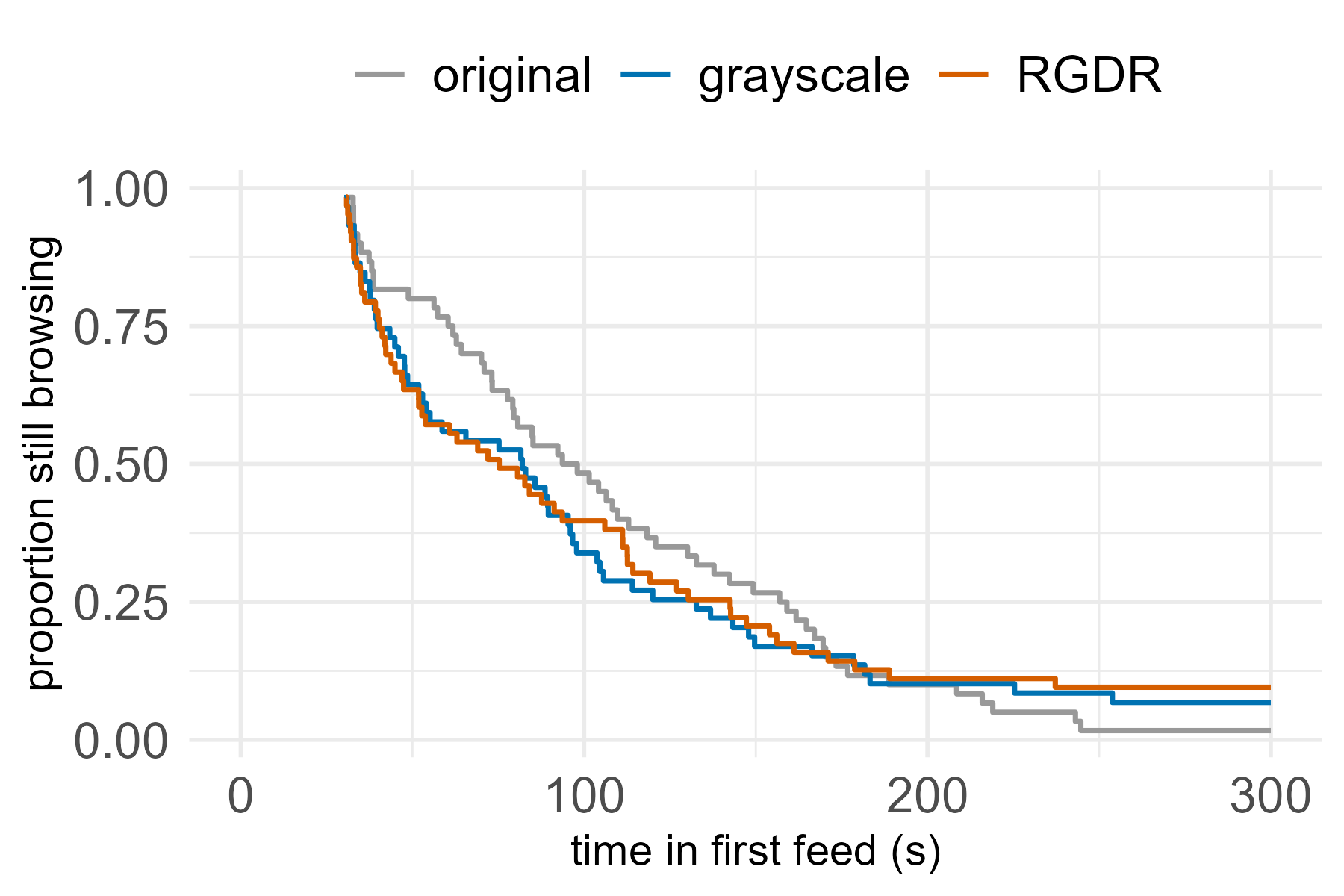}  
        \caption{Proportion of participants remaining in first feed over time.}
        \label{fig:plots_social_media_study:a}
    \end{subfigure}
    \hspace{-4mm}
    \begin{subfigure}{.46\textwidth}
        \centering
        \includegraphics[width=.95\linewidth]{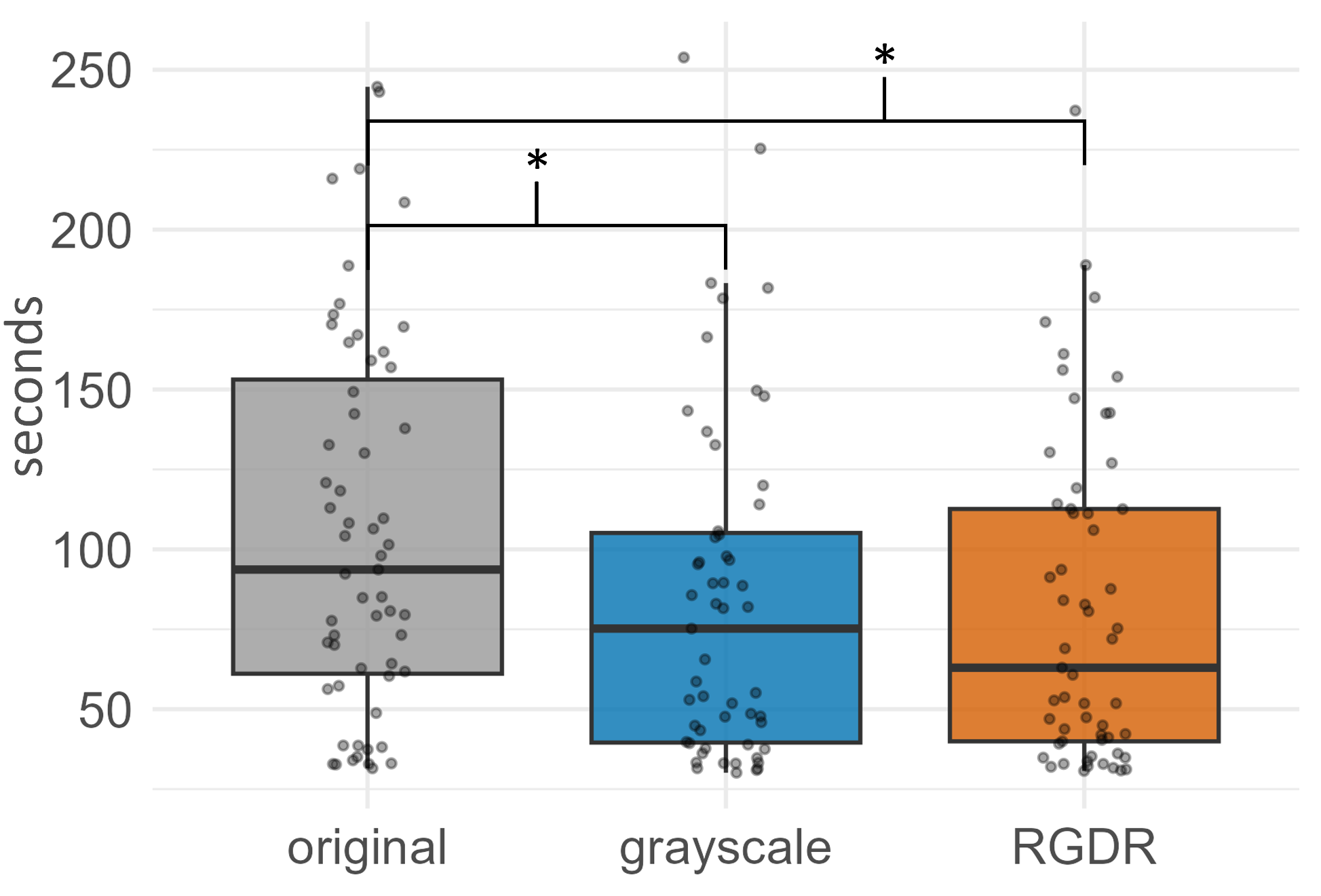}
        \caption{Duration in first feed conditioned on leaving.}
        \label{fig:plots_social_media_study:b}
    \end{subfigure}
    \caption{\textbf{Results of social media study.} 
    % (a) Histogram depicting the distribution of durations participants spent in the first feed. 
    (a) Kaplan-Meier plot estimating the proportion of participants remaining in the first feed over time, by condition. Curves step down at each observed departure.
    (b) Boxplots illustrating durations spent in the first feed conditioned on the sample of participants that were switching feeds, with significant differences relative to the original images indicated (* p < 0.05).
    % , ** p < 0.01, and *** p < 0.001).  
    In a boxplot, the box represents the interquartile range (IQR; Q1–Q3), the central line marks the median, and the whiskers extend to the smallest and largest values within 1.5 × IQR, with points beyond plotted as outliers.}
    \Description{Results of the social media study across three conditions: original, grayscale, and RGDR. Panel (a) shows Kaplan-Meier curves of the proportion of participants still browsing over time. The original condition remains highest for most of the 300-second window, while grayscale and RGDR decline faster and more steeply, converging with each other and with original by around 200 seconds. Panel (b) shows boxplots of time spent in the first feed, restricted to participants who left before the time limit. Original has the highest median duration, with grayscale and RGDR both lower and similar to each other. Significance markers indicate that both grayscale and RGDR differ significantly from original, with no significant difference between grayscale and RGDR.}
    \label{fig:plots_social_media_study}
\end{figure*}

\subsubsection{Temporal pattern of disengagement}
\label{sec:social-study:exploratory}
% While significance testing over the whole population did not yield differences, we exploratorily analyzed if a temporal pattern of disengagement can be identified among participants.
%
The Kaplan--Meier plot (Fig.~\ref{fig:plots_social_media_study:a}) shows a bimodal pattern: participants in adapted conditions tended to leave earlier, yet were also more likely to remain for the full 300-s window (participants reaching 300\,s: \cdiff{}~6, \cgray{}~4, \cdefault{}~1; Fisher's exact \pvall{=}{.190}).
Motivated by this pattern, we examined disengagement timing among the subset of participants who left the feed within the window ($n = 171$, Figure~\ref{fig:plots_social_media_study:b}). As leaving the feed is itself a post-treatment behavior, this analysis characterizes disengagement timing conditional on leaving. 
Among these participants, we found a main effect of \ivfilter{} on time spent in the first feed \anova{2.0}{111.7}{3.22}{=}{.044}{.039}. Post-hoc pairwise comparisons showed that both \cdiff{} \pvald{=}{.037}{-.44} and \cgray{} \pvald{=}{.044}{-.38} were associated with earlier disengagement relative to \cdefault{}.
A comparison between \cdiff{} and \cgray{} did not reveal a significant difference \pval{=}{.772}.

Conclusions for all other dependent variables remained unchanged between this subset and the full sample, with one exception: for the exploratory \cdiff{}-\cgray{} arousal comparison, the Holm-corrected parametric test reached significance \textbf{\pvald{=}{.028}{0.52}} while the Holm-corrected Wilcoxon test did not (\pvall{=}{.058}).

% ===== DURATION (LEAVERS ONLY) =====
% Welch ANOVA: F(2.0, 111.7) = 3.22, p = 0.0436, pes = 0.039494
%   family           contrast                             t    df p     p_registered_student     d  d_lo  d_hi wilcoxon_p p_holm wilcoxon_p_holm
% 1 planned_duration duration : grayscale vs original -2.04  112. 0.044               0.0444 -0.38 -0.75 -0.01 0.044      0.044  0.044          
% 2 planned_duration duration : RGDR vs original      -2.39  113. 0.018               0.0188 -0.44 -0.81 -0.07 0.021      0.037  0.042   
%             exploratory duration: RGDR vs grayscale -0.29  109. 0.772                0.772 -0.05 -0.43  0.32 0.762      0.772  0.762

%   family      contrast                        t    df p     p_registered_student     d  d_lo  d_hi shapiro_p_a shapiro_p_b wilcoxon_p p_holm wilcoxon_p_holm        
% 3 exploratory arousal: RGDR vs grayscale   2.75  110. 0.007              0.00701  0.52  0.14  0.89 0.024       0.009       0.015      0.028  0.058

\begin{figure*}[h!]
    %\centering
    \begin{subfigure}{0.45\textwidth}
        %\centering
        \includegraphics[width=0.99\linewidth]
        {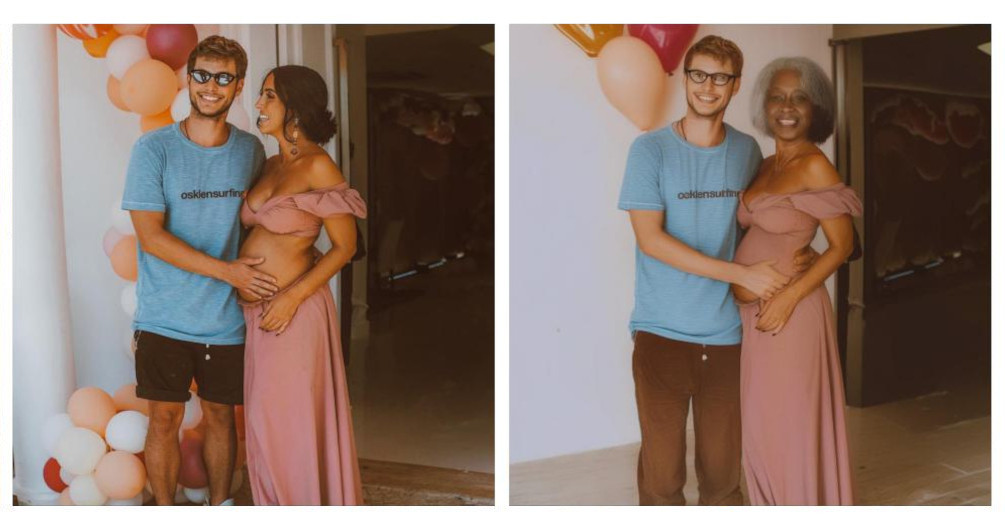}
        % {figures/social_media_results/coachkrystal_-1948820367988575749.jpg}  
        \captionsetup{width=0.95\linewidth}
        \caption{Increased apparent age. "The woman in a pink dress holding her belly with the man in a blue t-shirt."}
        \label{fig:social_media_results_1:age}
    \end{subfigure}
    \begin{subfigure}{0.45\textwidth}
        %\centering
        \includegraphics[width=0.99\linewidth]
        {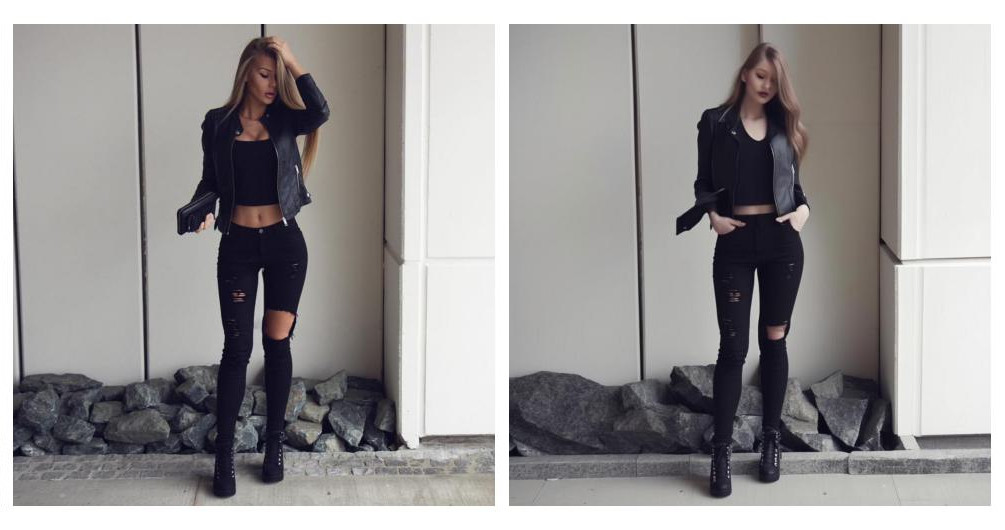}
        \captionsetup{width=0.95\linewidth}
        \caption{Reduced attractiveness cues. "A blonde woman poses in black ripped jeans and black top."}
        \label{fig:social_media_results_1:attractiveness}
    \end{subfigure}
    \begin{subfigure}{0.45\textwidth}
        %\centering
        \includegraphics[width=0.99\linewidth]
        {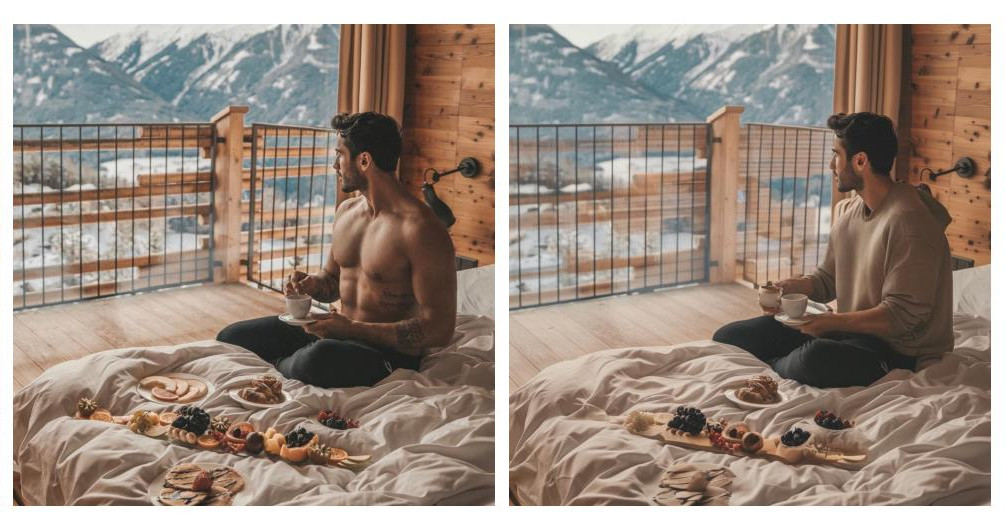}
        \captionsetup{width=0.95\linewidth}
        \caption{Increases clothing coverage. "A shirtless man holding a cup, sitting on a bed with a scenic mountain view."}
        %"Man holding a cup, sitting shirtless on a bed with a scenic mountain view outside."
        \label{fig:social_media_results_1:coverage}
    \end{subfigure}
    \begin{subfigure}{0.45\textwidth}
        % \centering
        \includegraphics[width=0.99\linewidth]
        {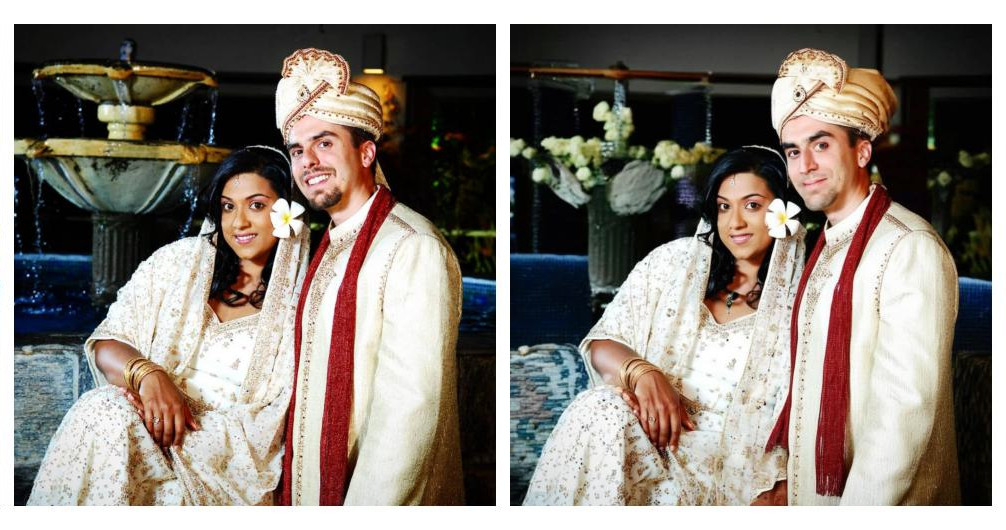}  
        \captionsetup{width=0.95\linewidth}
        \caption{Alters facial expression. "Bride and groom in traditional attire in front of a fountain"."}
        \label{fig:social_media_results_1:expression}
    \end{subfigure}
    \begin{subfigure}{0.45\textwidth}
        %\centering
        \includegraphics[width=0.99\linewidth]
        {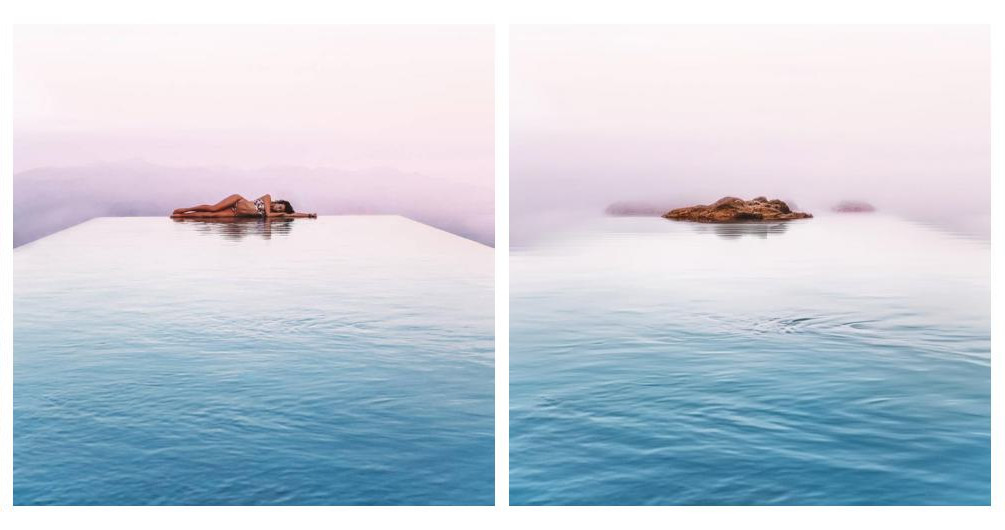}  
        \captionsetup{width=0.95\linewidth}
        \caption{Removes subjects. "Woman lying on a wooden deck above a calm water surface during a sunset."}
        \label{fig:social_media_results_1:remove}
    \end{subfigure}
    \begin{subfigure}{0.45\textwidth}
        %\centering
        \includegraphics[width=0.99\linewidth]
        {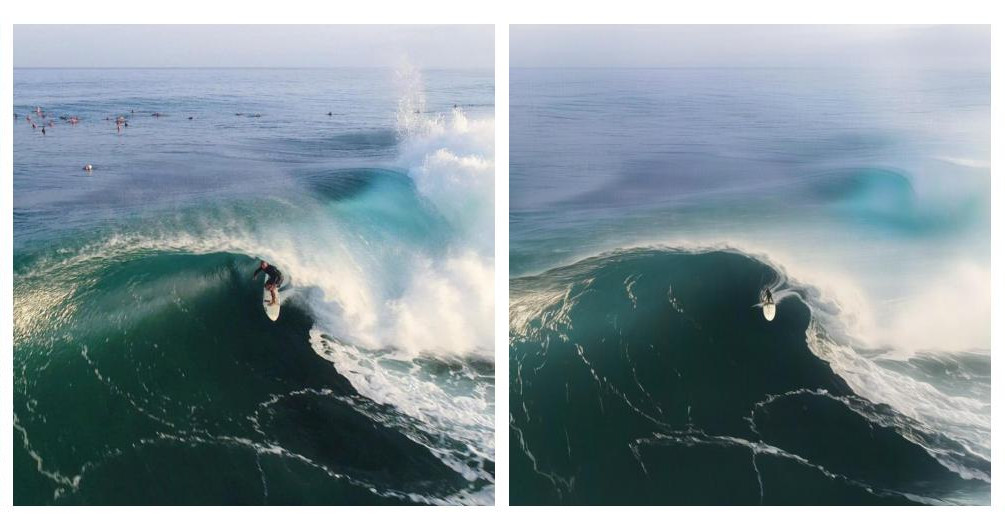}  
        \captionsetup{width=0.95\linewidth}
        \caption{Reduces subject size. "Surfer riding a large ocean wave with people in the background."}
        \label{fig:social_media_results_1:size}
    \end{subfigure}
    \begin{subfigure}{0.45\textwidth}
        %\centering
        \includegraphics[width=0.99\linewidth]
        {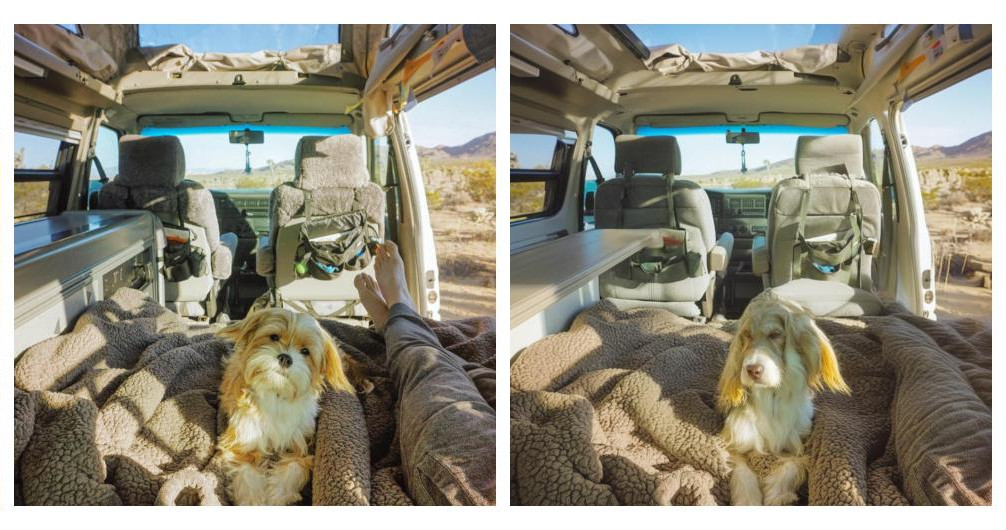}  
        \captionsetup{width=0.95\linewidth}
        \caption{Reduced baby-schema features. "A small dog sitting on a bed inside a camper van with mountains visible through the front window."}
        \label{fig:social_media_results_1:animal}
    \end{subfigure}
    \begin{subfigure}{0.45\textwidth}
        %\centering
        \includegraphics[width=0.99\linewidth]
        {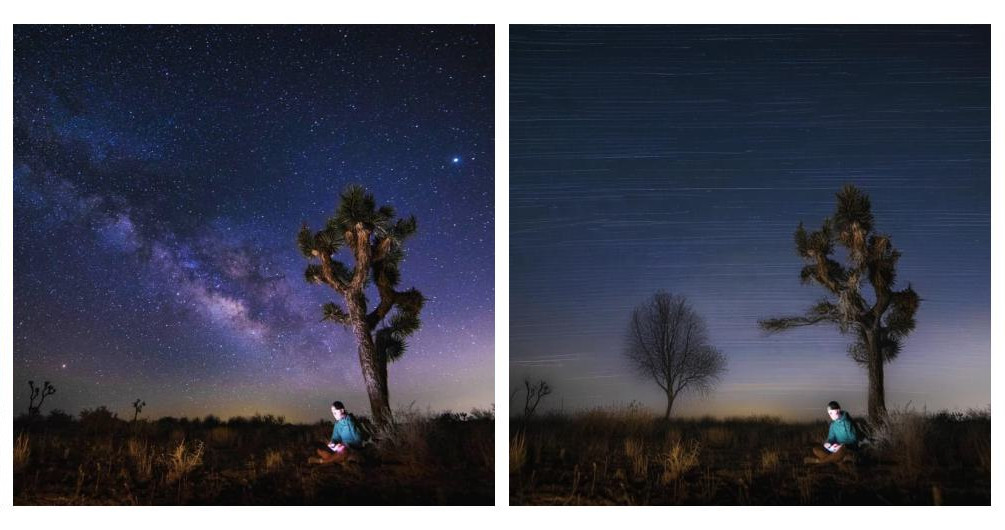}  
        \captionsetup{width=0.95\linewidth}
        \caption{Simplifies background. "Person sitting under a tree gazing at the starry night sky with the Milky Way visible in the background."}
        \label{fig:social_media_results_1:simple}
    \end{subfigure}
    \caption{\textbf{\diff{}-adapted social media images}; each illustrating a typical change it induces (and input caption).}
    \Description{Social media images adapted with RGDR, each illustrating a typical induced change. (a) Subjects appear older in a portrait of a couple. (b) A woman’s attractiveness is reduced in a fashion pose. (c) A shirtless man in a mountain-view bedroom is shown clothed. (d) A wedding couple have their facial expression changed. (e) A woman lying on a wooden deck at sunset is removed. (f) A surfer riding a large ocean wave is reduced in size. (g) A small dog in a camper van appears less cute. (h) A person under a tree gazing at the starry night sky is shown with a simpler background.}

    \label{fig:social_media_results_1}
\end{figure*}

\subsubsection{Qualitative observations of adapted stimuli}
\label{sec:sm-qualitative}
The authors inspected all 268 stimuli and compiled recurring, illustrative types of change.
Figure~\ref{fig:social_media_results_1} shows characteristic examples.

\paragraph{Intended shifts.}
\cdiff{} introduces semantic modifications consistent with its emotional
conditioning.
Commonly observed changes include increasing apparent age (Fig. \ref{fig:social_media_results_1:age}), shifting subjects away from conventional attractiveness cues (Fig. \ref{fig:social_media_results_1:attractiveness}), increasing clothing coverage (Fig. \ref{fig:social_media_results_1:coverage}), or altering facial expressions (Fig. \ref{fig:social_media_results_1:expression}).
Additionally, \diff{} may remove subjects from images (Fig. \ref{fig:social_media_results_1:remove}), reduce their size (Fig. \ref{fig:social_media_results_1:size}), attenuate baby-schema features in animals (Fig. \ref{fig:social_media_results_1:animal}), or simplify the background (Fig. \ref{fig:social_media_results_1:simple}).
Collectively, these changes are consistent with theoretical accounts of how high-level content features influence emotional responses (e.g., nudity increasing arousal or attractiveness eliciting positive emotions; see Sec.~\ref{sec:image-properties}), and provide qualitative evidence that \cdiff{} shifted image content toward the defined \emref{}.

\paragraph{Unintended biases.}
The same inspection revealed ethically problematic changes. 
\cdiff{} sometimes altered the apparent skin tone of depicted people (Fig. \ref{fig:social_media_results_1:age}), and reached lower arousal partly by increasing apparent age (Fig. \ref{fig:teaser}) or decreasing conventional attractiveness by Western beauty standards (Fig.\ref{fig:social_media_results_1:attractiveness}). 
We discuss the implications in Section~\ref{sec:ethics}.

\subsection{Discussion}
\label{sec:social-study:discussion}
Because \rques{2} posits a mediated effect, our design tested two links in sequence: whether the adaptations shift perceived emotion, and whether browsing behavior changes accordingly. Neither held. Self-reported valence and arousal did not differ from \cdefault{} in either adapted condition, and time spent in the first feed was unaffected as well as image quality. 
% \rques{2} is therefore not supported.
%
We are cautious about reading this null result as evidence of no true effect, given the measurement limitations of a single, retrospective, between-subjects rating (Sec.~\ref{sec:limitations-internal}). The emotional pathway \rques{2} posits therefore remains untested rather than disconfirmed.

Exploratory analyses further indicated a disengagement-timing effect: among participants who left the feed, both \cdiff{} and \cgray{} were associated with leaving sooner than \cdefault{}, with no difference between the two. Because only those who chose to leave are included, this speaks to how quickly people left once they did, not to whether the adaptations caused them to leave in the first place.

\cdiff{} and \cgray{} were associated with comparable disengagement timing, yet \cdiff{} was rated significantly \emph{more} arousing than \cgray{}. 
If lowered arousal were driving disengagement, the two conditions producing equivalent timing should not differ on the proposed mediator.
This difference did not survive restriction to the leavers-only sample. 
However, because leaving the feed may itself depend on how aroused participants felt, this restriction may have removed the participants driving the arousal difference. We therefore cannot tell whether the effect is fragile or merely obscured.

\section{Discussion}
\label{sec:discussion}
We now discuss findings that cut across our three evaluations.

\paragraph{Converging evidence}
\cdiff{} is the only approach for which technical, perceptual, and behavioral evidence align: it moved images toward \emref{} on technical metrics (Sec.~\ref{sec:technical-evaluation}), was the only approach to shift perceived emotion without a perceived-quality cost (Sec.~\ref{sec:image-study}), and, among participants who left the feed, was associated with earlier disengagement (Sec.~\ref{sec:social-study}).

Notably, in the social media study, \cdiff{} and \cgray{} did not differ behaviorally, but the two differ in cost: grayscale, an established digital well-being tool, reduced perceived image quality relative to original in the image rating study, whereas \diff{} did not. This positions regressor-guided image editing as a candidate mechanism for digital well-being interventions that reduce time spent online without the visible cost to image quality incurred by existing approaches.

\paragraph{Level of visual abstraction}
The evaluated approaches span a continuum of visual abstraction, from pixel-level filters (grayscale, manual) through per-image photometric optimization (style, parametric) to semantic content editing (\diff{}). Positioning them along this continuum reveals a consistent trade structure across four dimensions: emotional efficacy, image quality, semantic fidelity, and computational cost.

In terms of emotional efficacy, the technical evaluation showed that style and parametric optimization shifted images toward the \emref{} more strongly than \diff{}, as predicted by the regressor.
However, this came at the cost of a higher reconstruction error and higher KID/FID, indicating that their modulation is achieved through photometric change. In the image rating study, this stronger technical shift did not translate into a stronger perceptual one: neither approach significantly changed perceived emotion relative to \cdefault{}, while both were rated lower in perceived quality.
\diff{} shows the inverse pattern: its perceptual advantage comes at the cost of semantic fidelity, evident in both the DINO score and the qualitative examples.

Grayscale occupies a distinct position in this space. Like \style{} and \param{}, it operates at the pixel level and leaves scene semantics unchanged, yet it did not differ behaviorally from \diff{}. Unlike \diff{}, however, it incurred a loss in perceived image quality, showing that technical quality (KID/FID), can be dissociated from perceived quality. 

The four approaches also differ substantially in computational cost: grayscale requires negligible compute, \param{} requires little (17.8\,s/image), and the model-based methods \style{} (71.7\,s/image) and \diff{} (68.7\,s/image) require roughly 4x more (App.~\ref{apx:runtime}).

The resulting trade structure, not any single approach, is the more general contribution of the comparison: emotional efficacy, perceived quality, and semantic fidelity cannot currently be maximized simultaneously, and a practitioner must choose which to prioritize, considering a compute budget that rises toward the high-abstraction end of the spectrum. 
% Further tuning of the objective functions may move individual approaches along this frontier, but we see no evidence in our results that it would remove the trade-off itself.

\paragraph{Emotional mediation}
Prior work linking emotion to online engagement is largely observational and concerns up-regulation: platforms and content that elicit stronger emotional responses are associated with greater engagement (Sec.~\ref{sec:emotions_internet_use}). 
Whether the reverse holds, namely that deliberately down-regulating emotional response reduces engagement, does not follow automatically from this correlational literature and had not previously been tested.
Unlike prior work, we manipulate the emotional properties of viewed content directly and observe the behavioral consequence, which lets us speak to the direction of the effect rather than only its association. Despite an emerging consensus that emotion mediates engagement, no study has yet demonstrated this mediation statistically \cite{schreiner2021impact}.
Our social media study is no exception: its results can neither confirm the hypothesized mediation as the underlying mechanism nor rule it out.
We attribute this to aspects of our study design that limit sensitivity to emotional change (Sec.~\ref{sec:limitations-internal}).

\paragraph{Implications for digital well-being design}
Our results position content adaptation as a distinct paradigm for digital well-being, alongside access restriction (blockers, timers, lockouts), reflective nudges (usage statistics, break reminders), and gamification (streaks, goals, rewards; see Sec.~\ref{sec:related-work:dscts}). Unlike access restriction, it does not depend on sustained self-control and avoids the circumvention dynamic common to lockout tools \cite{LyngsHackMyself2020}. Unlike gamification, it does not rely on extrinsic reward loops that echo the same engagement-optimization logic driving excessive use in the first place. 
And unlike an abstinence framing shared by restriction and reflection-based approaches, it changes what is experienced during use rather than asking users to reduce or resist it, allowing disengagement to follow as a natural consequence. 
Grayscale's existing deployment as a mobile well-being feature shows that this class of intervention already exists.

\section{Limitations \& Future Work}
\label{sec:limitations}
While our findings provide valuable insights, they remain preliminary and several aspects of our method, study design, and usage context constrain what they establish.

\subsection{Method and implementation}
\label{sec:limitations-method}
Both parametric and style optimization vary in whether they achieve directionally consistent adaptation across images: while their average effect is strong (Sec.~\ref{sec:technical-evaluation}), per-image optimization can converge to a local minimum that moves an image away from the \emref{}. 
Learning-based approaches that learn a shared manifold across images, rather than optimizing each independently, could yield more consistent results~\cite{goetschalckx2019ganalyze, mejjati2020look}. Given that technical and perceived shifts do not align (Sec.~\ref{sec:image-rating-discussion}), this could reveal perceptual effects currently masked by inconsistency across images.

\diff{} introduces emotional changes by backpropagating classifier gradients to the predicted noise, which can destabilize synthesis when these gradients conflict with textual conditioning.
% Existing diffusion-based methods address this using attention-based regressor guidance, but this can introduce oversaturation that conflict with our goal of emotionally neutral images (see Sec.~\ref{sec:rw-diffusion}).
Rectified flow models \cite{esser2024scaling,lipman2022flow}, which follow straighter sampling trajectories, may be easier to guide precisely.
Future work should investigate whether these models can better align CG and CFG gradients to ensure qualitatively stable synthesis.

Runtime remains unsuitable for real-time use across all methods (App.~\ref{apx:runtime}), a particular constraint given that video is the dominant modality on social media.
A parametric approach is promising here, since parameters could be predicted once and applied across frames~\cite{mejjati2020look}. For \diff{}, a teacher-student distillation approach, commonly used in industry for real-time video filters \cite{vakunov2025massive}, could train a smaller model on the image pairs our method generates to enable real-time, per-frame adaptation.
Finally, our ablations showed that regressor guidance contributes beyond text conditioning alone (App.~\ref{app:diffusion-ablation}); whether the stronger text-driven editing of recent multimodal models \cite{hurst2024gpt,google2025nanobananapro} closes this gap remains open.

\subsection{Study design}

\subsubsection{Internal validity}
\label{sec:limitations-internal}
Our image rating study used 12 images from the NAPS database, spanning the CMA dimensions. 
This number was chosen to keep the within-subject session length feasible, given that each image was rated under five conditions. 
With only 12 stimuli, effects could in principle be driven by a small subset of particularly emotionally charged images. 
However, repeating the primary contrasts with image rather than participant as the unit of analysis found effects directionally consistent across images (App.~\ref{apx:im-rating-by-image}).
Yet, future work should replicate these findings with a larger and more diverse stimulus set.

Our assessment of fidelity (\crit{4}) relies on computational proxies. While DINO similarity is an established, human-aligned measure of content preservation \cite{basu2023editval}, we did not collect side-by-side human judgments comparing adapted images against their originals. Future work should collect such comparisons.

The hypothesized mediation of \rques{2} remains unconfirmed in part because several aspects of our design limit sensitivity to emotional change. We measured perceived emotion only once, retrospectively, for the whole feed, in a between-subjects design; measures of this kind are difficult to detect against inter-individual variability~\cite{bekhtereva2017bringing,de2010effects}. A within-subjects design would be more robust to this variability, but pre-studies showed that repeated exposure to the same feed produced familiarity effects, while varying feeds across conditions introduced content confounds we could not control. 
Future work should collect in-feed affect ratings per post rather than once per session, using designs that reduce familiarity effects without introducing content confounds, e.g., by using longer breaks between conditions.

\subsubsection{External validity}
Because our methods cannot yet adapt content in real time, our study used a curated feed rather than participants' own social media content. This improved internal validity by ensuring all participants viewed the same content, but did not reflect the personal relevance or social ties that characterize everyday social media use, and effects that depend on these factors would not have surfaced.
Observing participants during a fixed-duration session is itself unlike everyday social media use and likely influenced behavior further.

We approximated competing real-world activities with a single alternative feed, chosen over an unrelated activity to minimize interpersonal variation in perceived task value. This does not reflect the diversity of everyday alternatives, which prior work shows influences social media exit decisions~\cite{rixen2023loop}, and unrepresented activities may have shaped participants' disengagement.

Both studies operated under time constraints suitable for a single session. 
To ensure measurable effects within this window, we deliberately selected images likely to elicit strong emotional responses: in the image-rating study based on pre-labeled valence and arousal values, and in the social media study by manual selection of images showing the clearest shift relative to their originals.
Real-world content is typically more emotionally neutral and consumed over longer periods. Engagement dynamics may differ accordingly.

Future work should test whether these effects generalize to personalized, unobserved, real-world browsing; to the fuller range of everyday competing activities; and to more emotionally neutral content consumed over longer timescales.

\subsubsection{Sample}
Both studies recruited UK participants via Prolific, yielding a culturally homogeneous sample. Culture shapes both emotional experience~\cite{redies2015combining} and social media use~\cite{lopez2023problematic}, so our findings may not generalize to more diverse populations. 
Participants ranged from 18--75 years (M=36 and 39), whereas PIU is concentrated among those aged 15--24~\cite{lavoie2023relationship,Stortz2024}. Future work should test whether these findings hold in this younger, more vulnerable population.

\subsection{Outlook}
Our results represent a snapshot of one parameter configuration per approach. Because the weighting of content preservation against emotional guidance is freely parameterizable, better operating points on each approach's trade-off curve likely exist. Identifying them is an empirical question best addressed through user-centered methods, such as JND-style staircase procedures that establish the minimal adaptation strength producing a perceptible emotional shift. In deployment, the appropriate setting is unlikely to be static: personal relevance strongly shapes emotional response \cite{pilarczyk2014emotional, humphrey2012salience}, suggesting closed-loop systems that assess user state and adapt intervention intensity over time.

A more fundamental question concerns the outcome measure itself. We treated reduced browsing time as the target, but reduced usage is not unconditionally beneficial: whether usage is experienced as positive or negative depends substantially on user intention \cite{hsu2026does}. Users also deliberately seek out social media for its emotional effects \cite{davis2025you}, which our approach attenuates by design. A deployable system would need to distinguish intentional from unintentional use, and it remains unclear how users would evaluate an intervention that dampens an effect they actively sought. Understanding how users and other stakeholders perceive these adaptations is a necessary next step.
% , which we discuss further in Sec.~\ref{sec:ethics}.

Beyond images, online content also comprises text and audio, and future work should examine whether comparable emotional adaptation is feasible in these modalities.

Finally, emotional adaptation may also benefit users beyond reduced time online. Social media's emphasis on aspirational content is linked to negative social comparison and its downstream harms, particularly among younger users~\cite{de_vries_social_2018}. Because our approach reduces the emotional intensity such content relies on, we hypothesize that it may also attenuate social comparison and associated feelings of inadequacy. 

\section{Ethical Implications}
\label{sec:ethics}
Emotion-modulating image adaptation raises ethical concerns. 
However, the relevant comparison is not an unmodified baseline. Social media feeds are already algorithmically optimized for engagement \cite{wu2017attention}. Creators also routinely use tools such as beautification filters to amplify emotional appeal. Our approach does not introduce emotional manipulation into a neutral system. Instead, it relocates control over emotional intensity from platforms and creators to the consumer. The ethical question is therefore not whether emotional manipulation exists, but whether user-initiated adaptation is more or less legitimate than the producer-side manipulation users already face.

However, this legitimacy is not automatic and it depends on how the system is deployed.
Adaptation should be explicitly initiated by users, align with their own goals, be transparently communicated, and remain reversible on demand. Absent these conditions, the same mechanism becomes covert reality-shaping or censorship rather than user-chosen self-regulation. This distinction needs to be safeguarded through system design.

A related concern is that adaptation alters depictions of people who did not consent to it, whether public figures or a user's own friends and family. 
Rendering an image privately for a single viewer makes redistribution less likely but does not prevent it.
% : the rendered image can still be captured and shared onward like any other image.
A partial technical mitigation is to mask human subjects and exclude them from adaptation, though this forgoes adaptation of exactly the content that drives the strongest emotional responses~\cite{aharon2001beautiful,gerger2011faces}. 
How to protect depicted people under these conditions remains an open problem that needs to be addressed before deployment.

Our qualitative analysis also surfaced a bias-related finding: \diff{} systematically altered demographic attributes such as age or skin color (Sec.~\ref{sec:sm-qualitative}).
We attribute them to the guidance signal rather than the editing mechanism: because the regressor is trained on human valence and
arousal annotations drawn largely from affective databases collected in Western populations, any correlation between these ratings and demographic attributes becomes a direction the gradient actively optimizes toward.
The risk is not negligible: systematic attenuation could disproportionately fade depictions of particular groups. Before any consumer-facing deployment, an audit protocol is needed that measures drift in demographic attributes before and after editing across a balanced stimulus set, and reports this drift as a fairness metric alongside efficacy.

The same machinery could also be optimized toward manipulative targets, e.g., increasing arousal to drive online engagement. 
However, this capability is not unique to our work, since it is latent in the underlying generative models rather than something we introduced. 
Our contribution instead makes explicit how readily emotional image adaptation can be realized with existing techniques. 
We see this transparency as ethically valuable, since it can support informed debate and the development of norms and safeguards. 
In this sense, \diff{}'s edited images function as an Artefact in Mauri et al.'s taxonomy of design futures \cite{mauri2026taxonomy}: a tangible manifestation of a possible future that can prompt early ethical reflection.  
Future work should use these artefacts to study how stakeholders perceive these adaptations, and what ethical concerns that perception surfaces.

% Both studies received IRB approval. They were conducted with informed consent and compensated participation, and collected only anonymous, non-health-related ratings.

\section{Conclusion}
This paper examined whether modifying the emotional properties of images can support non-coercive reductions in time spent online.
We introduced and systematically analyzed three regressor-guided image adaptation approaches spanning levels of visual abstraction: \param{}, \style{}, and \diff{}.

Across a technical evaluation, a controlled image-rating study, and a social media experiment, we found that while all three approaches can technically steer images toward a target of neutral valence and low arousal, only \diff{} altered perceived emotions without degrading perceived visual quality. Total browsing duration did not differ across conditions. Exploratory analyses, however, showed that participants who disengaged before the time limit did so earlier under both \diff{} and grayscale feeds than under unedited ones. Notably, grayscale, an already deployed digital well-being feature, achieved this at a measurable cost to perceived image quality, whereas \diff{} did not.

Positioning these approaches along a continuum of visual abstraction revealed a consistent trade structure: emotional efficacy, perceived image quality, and fidelity to the original could not be maximized simultaneously. \diff{}'s preservation of perceived quality comes at the cost of altering the semantic content of an image. 
Realizing this potential responsibly requires addressing the demographic drift introduced by diffusion-guided editing. Future work should explore these approaches in more ecologically valid settings and with designs better suited to isolating emotional mechanisms.

\section*{Note on Paper Length}
This paper's length reflects the scope of its contribution: three image-editing methods, evaluated across a technical analysis, a controlled rating study ($N=56$), and a preregistered behavioral study ($N=188$), together with the fairness findings this evidence requires. Derivations, ablations, and study materials are placed in appendices; the main text contains only material needed to assess the claims.

\bibliographystyle{ACM-Reference-Format}
\bibliography{citations}

\appendix

\section{Method Details}

\subsection{Regressor}
\label{app:regressor}
We train the emotion regressor by fine-tuning a pre-trained ResNet50 model (ImageNetV2) on a custom dataset composed of multiple emotional databases. 
Specifically, we integrate NAPS \cite{marchewka2014nencki}, DIRTI \cite{haberkamp2017disgust}, CGnA10766 \cite{kim2018building}, EmoMadrid \cite{carretie2019emomadrid}, Emotion6 \cite{peng2015mixed}, GAPED \cite{dan2011geneva}, OASIS \cite{kurdi2017introducing}, OLAF \cite{miccoli2016affective}, MAPS \cite{goodman2016military}, and IAPS \cite{lang1997international}, which are open-science emotional image databases providing human valence and arousal ratings consistent with the CMA, resulting in a combined dataset of 20,460 images.

\begin{figure}[tbh]
	\centering
    \includegraphics[width=0.4\linewidth]{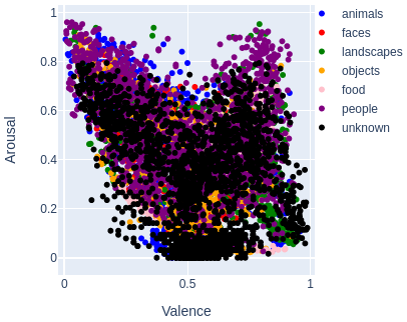} 
    \caption{
    \textbf{Location of images in CMA}: Location of stimuli of our integrated dataset within the valence arousal space of the Circumplex Model of Affect \cite{russell_circumplex_1980}. 
    }
    \Description{Scatter plot of arousal versus valence for images across categories. Data points cluster densely in the mid to high arousal and low to mid valence range. Colors indicate categories: animals, faces, landscapes, objects, food, people, and unknown. People and unknown dominate the distribution, while other categories appear sparser.}
    \label{fig:va_datasets}
\end{figure}

We apply min–max normalization to valence and arousal ratings across databases, mapping them onto a unified scale from 0 to 1.
Figure~\ref{fig:va_datasets} shows the distribution of the integrated dataset within the resulting valence--arousal space. 
It exhibits the characteristic V-shape of affective rating data, with high arousal concentrated at both valence extremes

\subsection{Parameterization of the transform}
\label{apx:transform}

We implemented the transform $T$ with standard differentiable Kornia~\cite{eriba2019kornia} operations for exposure, saturation, contrast, sharpening, blur, and scaling. The tone curve follows Mejjati et al.~\cite{mejjati2020look} and the color curve follows Hu et al.~\cite{hu2018exposure}; both are applied as $K$-step piecewise functions. Table~\ref{tab:transform-params} lists the full parameterization.

\begin{table}[h]
\centering
\small
\caption{\textbf{Sub-transforms composing $T$.} Parameters are optimized by Eq.~\ref{eq:param-opt} before being applied in the fixed order listed above. The \textit{Manual} column reports the single fixed configuration used for the manual baseline of Section \ref{sec:technical-evaluation}.}
\label{tab:transform-params}
\begin{tabular}{@{}lllll@{}}
\toprule
Operation & Dim. & Range & Identity & Manual \\
\midrule
Exposure   & 1          & $(-\infty,\infty)$ & 0.0 & $-0.1$ \\
Saturation & 1          & $[0,\infty)$        & 1.0 & 0.85 \\
Tone curve & $K$  & $(-\infty,\infty)$  & 1.0 & 1.0 (unchanged) \\
Color curve& $3K$       & $(-\infty,\infty)$  & 1.0 & 0.7 (R, B channels) \\
Contrast   & 1          & $[0,\infty)$        & 1.0 & 0.85 \\
Sharpness  & 1          & $[0,\infty)$        & 1.0 & 0.0 (unchanged) \\
Blur ($\sigma$) & 1     & $(0,\infty)$        & $\approx 0$ & 1.0 \\
Scale      & 4          & $[1,\infty)^2\!\times\![0,\text{input\_size}]^2$ & 1.0 & 1.0 (unchanged) \\
\bottomrule
\end{tabular}
\end{table}

$T$ applies these seven operations sequentially in the order listed in Table~\ref{tab:transform-params}: exposure (exponential brightness scaling), saturation (scales color intensity in HSV space), an 8-step piecewise tone curve, an 8-step per-channel color curve (hue rotation in HSV space), contrast (scales deviation from mean intensity), sharpening (high-frequency edge amplification), Gaussian blur, and an affine scale that zooms the image about an optimizable center point. 
For the affine scale transformation, each image dimension has a scale factor, constrained to $[1,\infty)$, and a center coordinate, constrained to the image extent $[0,\text{input\_size}]$.
Tone and color curves are multi-dimensional sub-transforms ($K{=}8$); scale is a 4-dimensional sub-transform; all remaining operations are scalar.

\subsection{Training MUNIT}
\label{apx:munit}
MUNIT uses an autoencoder-based generator within a GAN framework to factorize an image into a content vector and a style vector for unsupervised style transfer across domains. We adapt this approach by training MUNIT’s GAN on ImageNetV2 to obtain a general-purpose content–style disentanglement, following a methodology similar to \cite{zhu2022emotional}.

The model was trained for approximately 200,000 iterations, after which qualitative inspection and an FID of 3.2 on swapped-style images indicated effective disentanglement. Consistent with prior work \cite{zhu2022emotional}, the generator learned to separate high-level content (e.g., objects, people) from low-level stylistic characteristics (e.g., color, texture, brightness).
% Epoch 00000, Iteration 000200000, FID a 3.112177223180538, FID b 3.256906222781936
We used a high-resolution implementation of MUNIT provided by NVIDIA\footnote{\url{https://github.com/NVlabs/imaginaire}}.

\subsection{Training the guidance regressor}
\label{app:guidance-regressor}
To train $R_u$, we first encode the images from our custom dataset (Sec. \ref{sec:regressor}) into latent representations using the encoder of SDXL, $z_0 = E(I)$. We then add noise to these latents according to the forward diffusion process by randomly sampling a timestep $t$, resulting in $z_t = \text{ForwardDiffusion}(z_0, t, \beta)$ (see Supplementary Material for a definition of $\beta$).
These noisy latents, along with their corresponding timestep and the null-text embedding $\emptyset$, are passed through the diffusion model to obtain its middle-layer output $u_t = DM_{\text{half}}(z_t, t, \emptyset)$.
The output $u_t$ is then used as input to $R_u$, with valence and arousal ratings from our dataset serving as prediction targets.
This process is repeated until performance stabilizes.

\section{Implementation}
\label{app:implementation}
All approaches were implemented using PyTorch and operated on images at a resolution of 1024×1024.
Both, parametric and style optimization, employed Adam with parameters \(\beta_1 = 0.9\) and \(\beta_2 = 0.999\) and a learning rate of 0.05.

For \diff{}, we used the Diffusers implementation of SDXL from Stability AI\footnote{\url{https://huggingface.co/stabilityai/stable-diffusion-xl-base-1.0}}. Since SDXL relies on a float16 tensor representation, using default null-text optimization led to vanishing gradients. To address this, we re-implemented NTO with mixed-precision capabilities.
The network of RGDR's guidance regressor consisted of four convolutional layers, each followed by ReLU activation and max-pooling, before passing through two fully connected layers to produce the final output predictions. 
Image captions were generated using GPT-4 Vision (Azure OpenAI, gpt-4-vision-2024-05-01-preview) with the following prompt, applied when no captions were provided by the dataset:

\begin{quote}
\itshape
Provide a caption that describes the content of the image in a clear and concise manner. The caption should be suitable for use as ALT text in an HTML website and as descriptive label for a machine learning dataset. Ensure that the caption accurately reflects the key elements and context of the image.
% to improve accessibility
\end{quote}
\section{Technical Evaluation Details}

\subsection{Manual baseline configuration}
\label{apx:manual-baseline}
The \man{} baseline applies a single fixed parameter configuration for $T$ (Sec.~\ref{apx:transform}) uniformly to all images. One of the authors tuned this configuration on a subset of the COCO validation images, adjusting parameters iteratively until regressor-predicted valence approached $0.5$ and arousal approached $0.0$, subject to keeping the L1 reconstruction error within a visually acceptable range. The resulting values are listed in the \man{} column of Table~\ref{tab:transform-params}.

\subsection{OLS model statistics}
\label{apx:ols}
Table~\ref{tab:ols-full} reports the full ordinary least squares model terms underlying the omnibus statement in Sec.~\ref{sec:technical-evaluation:results}: the constant term ($\beta_0$), predictor term ($\beta_1$), their $p$-values, and the model $R^2$, for each approach and outcome (valence, arousal).

\begin{table}[h]
\centering
\caption{\textbf{OLS regression results.} Results predicting the change in valence and arousal from adaptation weight, for each approach.}
\label{tab:ols-full}
\begin{tabular}{llrrrrr}
\toprule
Outcome & Approach & $\beta_0$ & $p(\beta_0)$ & $\beta_1$ & $p(\beta_1)$ & $R^2$ \\
\midrule
\multirow{4}{*}{Valence}
& \style{} & $-0.0320$ & $<.001$ & $-0.0121$ & $.022$ & $.866$ \\
& \param{} & $-0.0233$ & $.004$  & $-0.0220$ & $.018$ & $.883$ \\
& CG       & $\phantom{-}0.0188$ & $<.001$ & $-0.0960$ & $<.001$ & $.994$ \\
& \diff{}  & $-0.0159$ & $<.001$ & $-0.0401$ & $.001$  & $.983$ \\
\midrule
\multirow{4}{*}{Arousal}
& \style{} & $-0.0784$ & $<.001$ & $-0.0274$ & $.027$ & $.848$ \\
& \param{} & $-0.0691$ & $.003$  & $-0.0640$ & $.017$ & $.884$ \\
& CG       & $-0.0643$ & $<.001$ & $-0.0282$ & $.003$ & $.967$ \\
& \diff{}  & $-0.0669$ & $<.001$ & $-0.0513$ & $.005$ & $.948$ \\
\bottomrule
\end{tabular}
\end{table}

\subsection{Fidelity to the original image}
\label{apx:fidelity-metrics}
\begin{figure*}[t]
    \centering
    \begin{subfigure}{.33\textwidth}
        \centering
        \includegraphics[width=.95\linewidth]{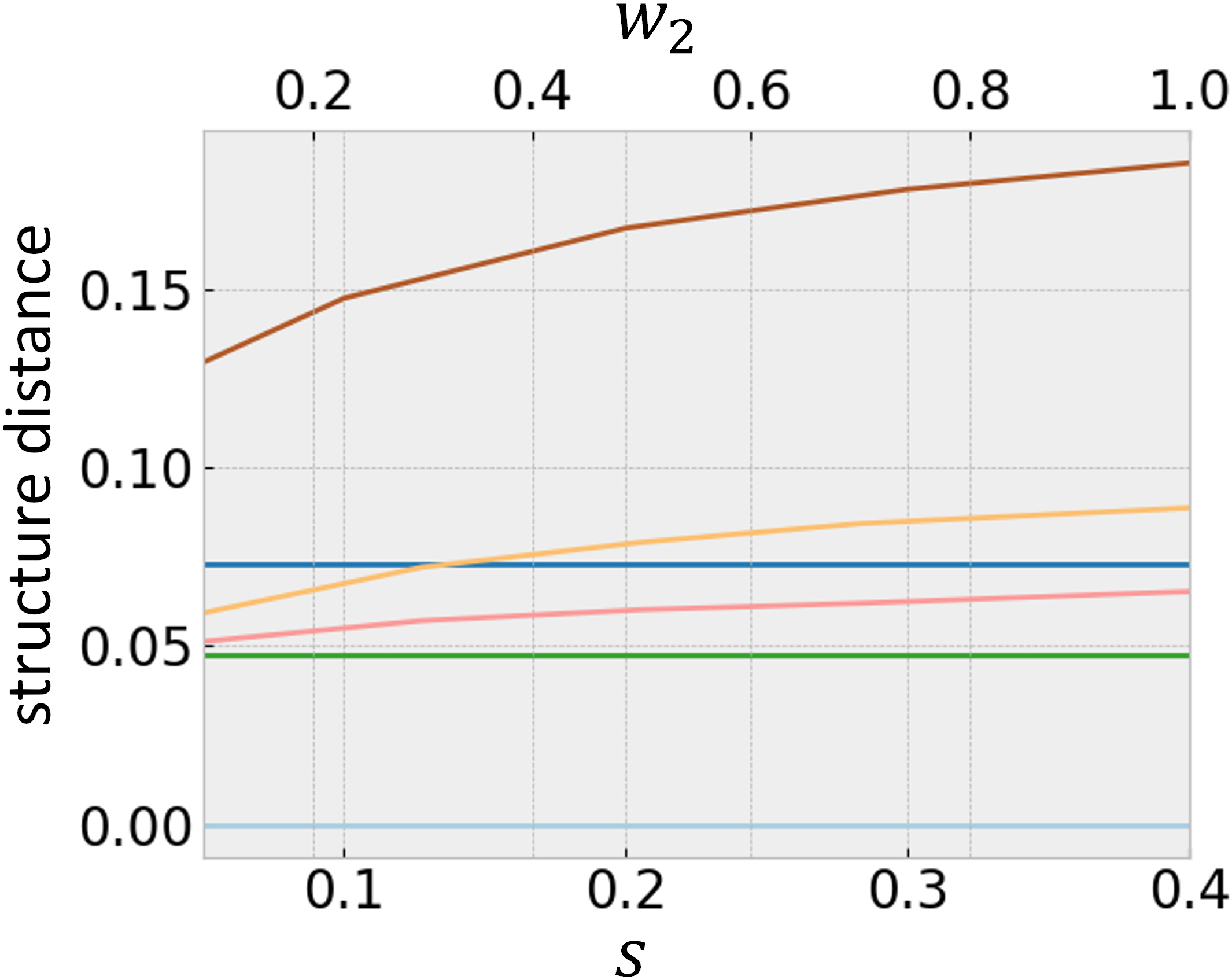}  
        \caption{structure distance}
        \label{fig:fidelity_original:structure_distance}
    \end{subfigure}
    \begin{subfigure}{.33\textwidth}
        \centering
        \includegraphics[width=.95\linewidth]{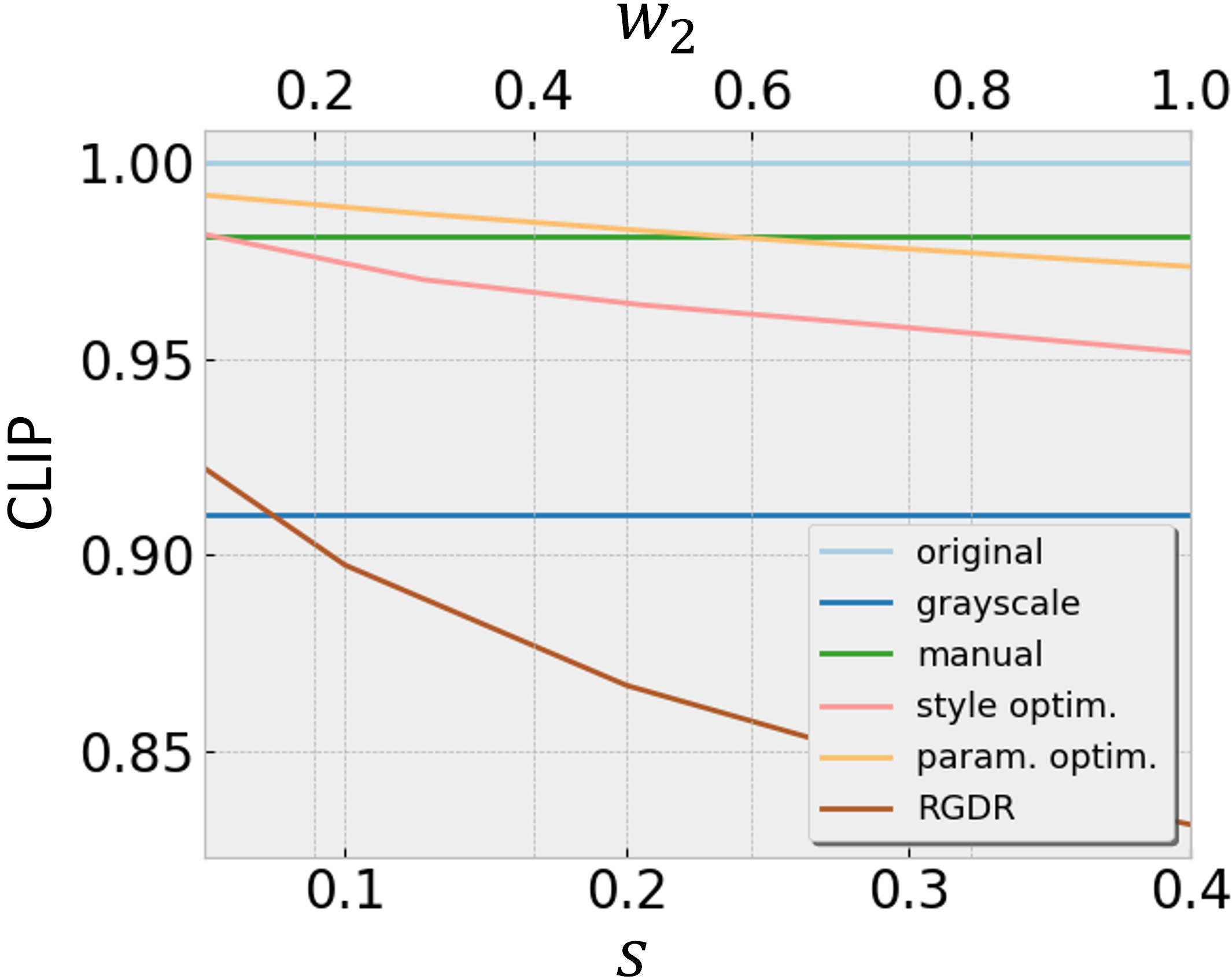}  
        \caption{CLIP score}
        \label{fig:fidelity_original:clip}
    \end{subfigure}
    \caption{\textbf{Fidelity to the original image analysis}. The plots show how CLIP and Structure Distance evolve as the weighting of \emph{emotional adaptation objective} increases. In each plot, the upper x-axis represents the values of $w_2$ for \param{} and \style{}, while the lower x-axis indicates the values of $s$ for \diff{}. The y-axes show the metric values.}
    \label{fig:fidelity_original}
\end{figure*}

Figure~\ref{fig:fidelity_original} illustrates two further fidelity metrics, CLIP similarity \cite{radford2021learning}, which computes the cosine similarity of image vectors in CLIP embedding space, and Structure Distance~\cite{tumanyan2022splicing}, the Frobenius distance between the self-similarity matrices of DINO-ViT keys, reproducing the ordering and crossover behavior of the DINO score reported in Section~\ref{sec:technical-evaluation}.

\section{Additional Technical Evaluation}
\label{apx:technical-evaluation}

\subsection{Runtime benchmark}
\label{apx:runtime}
We benchmarked the wall-clock runtime of our three image-editing approaches (\param{}, \style{}, \diff{}) on 50 images (1024$\times$1024) from the Instagram Influencer Dataset~\cite{kim2020multimodal}, using one exclusively reserved NVIDIA H200 GPU per approach, on a server with an Intel Xeon Platinum 8480C CPU (224 threads) and 2\,TB RAM, running Ubuntu 24.04.4 LTS. 
%
% Due to differing software dependencies, \param{} and \diff{} ran under PyTorch 2.5.1/CUDA 12.1 (Python 3.10.12), while \style{} ran under PyTorch 1.9.0/CUDA 11.3 (Python 3.8.10); version differences may affect kernel-level performance, but runtimes remain informative as an estimate of each method's practical, deployed cost.
%
For each method, we ran 5 untimed warm-up calls followed by 5 timed repetitions on each of the 50 images. 
% Timings were measured using \texttt{torch.cuda.synchronize()} around \texttt{time.perf\_counter()} calls.
Table \ref{tab:runtime-benchmark} summarizes the results.

\begin{table}[h]
\centering
\small
\caption{\textbf{Per-image runtime (seconds) across 50 images, 5 repetitions per image.} Median is computed over per-image medians; IQR and p95 characterize the spread of these per-image medians across the 50 images.}
\label{tab:runtime-benchmark}
\begin{tabular}{@{}lrrr@{}}
\toprule
Method & Median (s) & IQR (s) & p95 (s) \\
\midrule
\param{}  & 17.81 & 0.04  & 17.88  \\
\style{}  & 71.68 & 1.19  & 72.41  \\
\diff{}   & 68.66 & 39.07 & 120.07 \\
\bottomrule
\end{tabular}
\end{table}

Parametric and \style{} show low spread across images (IQR $\leq$ 1.19\,s), consistent with a fixed, content-independent number of optimization steps. \diff{} shows two orders of magnitude greater spread (IQR = 39.07\,s, p95 = 120.07\,s), which we attribute to image-dependent differences in the number of null-text optimization iterations required to converge at each denoising timestep~\cite{mokady2023null}.
% Within-image standard deviation (averaged across images) was $\leq$0.04\,s for \param{} and \style{}, but 3.16\,s for \diff{}. Since DDIM sampling/inversion and null-text optimization are deterministic given a fixed image, we attribute this to system-level timing noise (e.g., GPU clock state, allocator behavior) rather than algorithmic stochasticity. 
% This is amplified in \diff{} by its longer runtime and far larger number of sequential GPU operations per call.

\subsection{\diff{} ablation}
\label{app:diffusion-ablation}
To analyze the contributions of different components in our diffusion-based approach, we introduced three ablated versions of the method. The variants were proposed to identify the impact of individual components of our methodology:

\begin{itemize}
    \item \textbf{CG (classifier guidance only)}: This version employs the deterministic DPM multistep solver to invert an image \cite{lu2022dpm}. Resampling is performed using only classifier guidance, without text conditioning. 
    For this baseline, we use DPM instead of DDIM as our initial experiments have shown that it produces inverted images of higher quality without NTO (as CG is not text conditioned NTO cannot be applied).
    
    \item \textbf{\cfg{} (instruction-guided resampling)}: In this version, images are inverted using NTO and resampled with CFG only, without the use of CG. To guide resampling, the caption is extended with the sentence: \emph{``The image should elicit low arousal and neutral valence in its viewers.''}
    
    \item \textbf{\cfgedit{} (resampling with refined captions)}: This version applies NTO for image inversion and resampling with a modified caption. Original captions were refined using GPT-4 by removing emotionally charged words or phrases, yielding text that remains close to the original while ensuring a neutral tone.
\end{itemize}

All diffusion-model-based approaches guided by text, used the a CFG scale of $2.0$ and the image captions provided by the COCO dataset as textual input.
In addition, since the resolution of images in the COCO dataset is approximately 512×512, we utilized the Diffusers implementation of Stable Diffusion from Stability AI as the backbone for this experiment.
The classifier guidance scale for \diff{} and the CG baseline takes values in $s \in \{0.05, 0.1, 0.2, 0.3, 0.4\}$.

\begin{figure*}[t]
    \centering
    \begin{subfigure}{.33\textwidth}
        \centering
        \includegraphics[width=.95\linewidth]
        {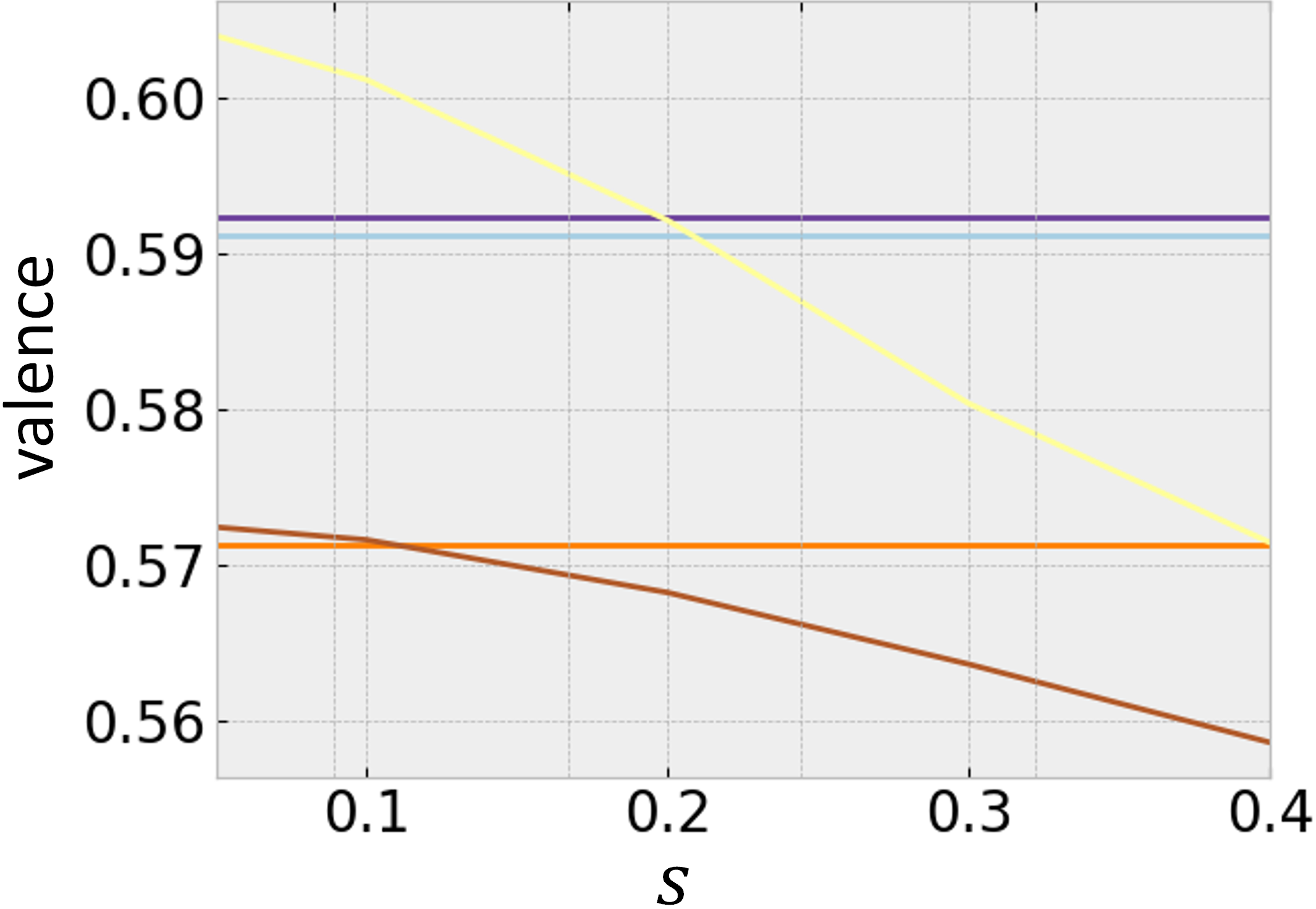}  
        \caption{valence}
        \label{fig:criteria_satisfaction_appendix:valence}
    \end{subfigure}
    %\hspace{-4mm}
    \begin{subfigure}{.33\textwidth}
        \centering
        \includegraphics[width=.95\linewidth]
        {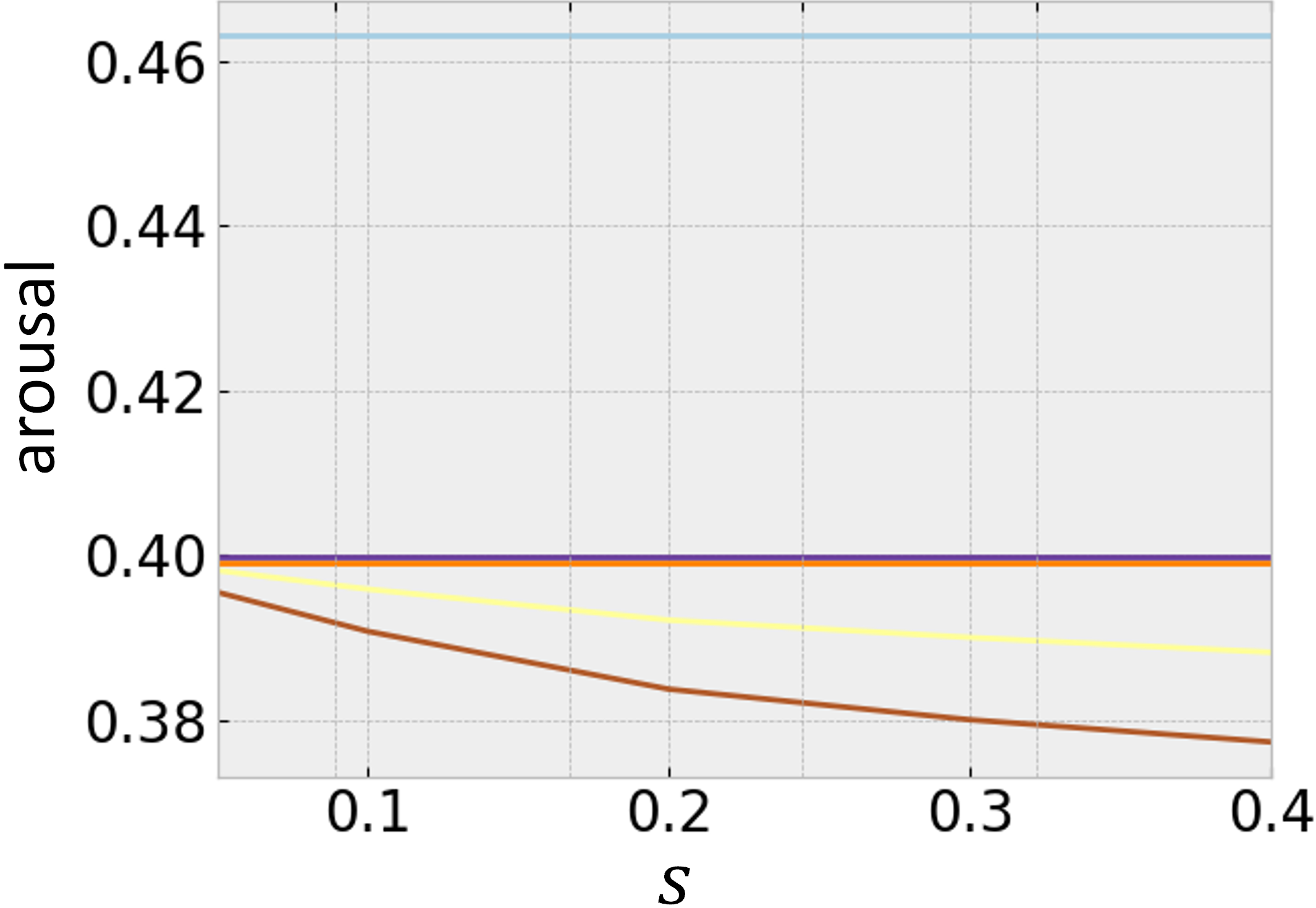}  
        \caption{arousal}
        \label{fig:criteria_satisfaction_appendix:arousal}
    \end{subfigure}
    \begin{subfigure}{.33\textwidth}
        \centering
        \includegraphics[width=.95\linewidth]{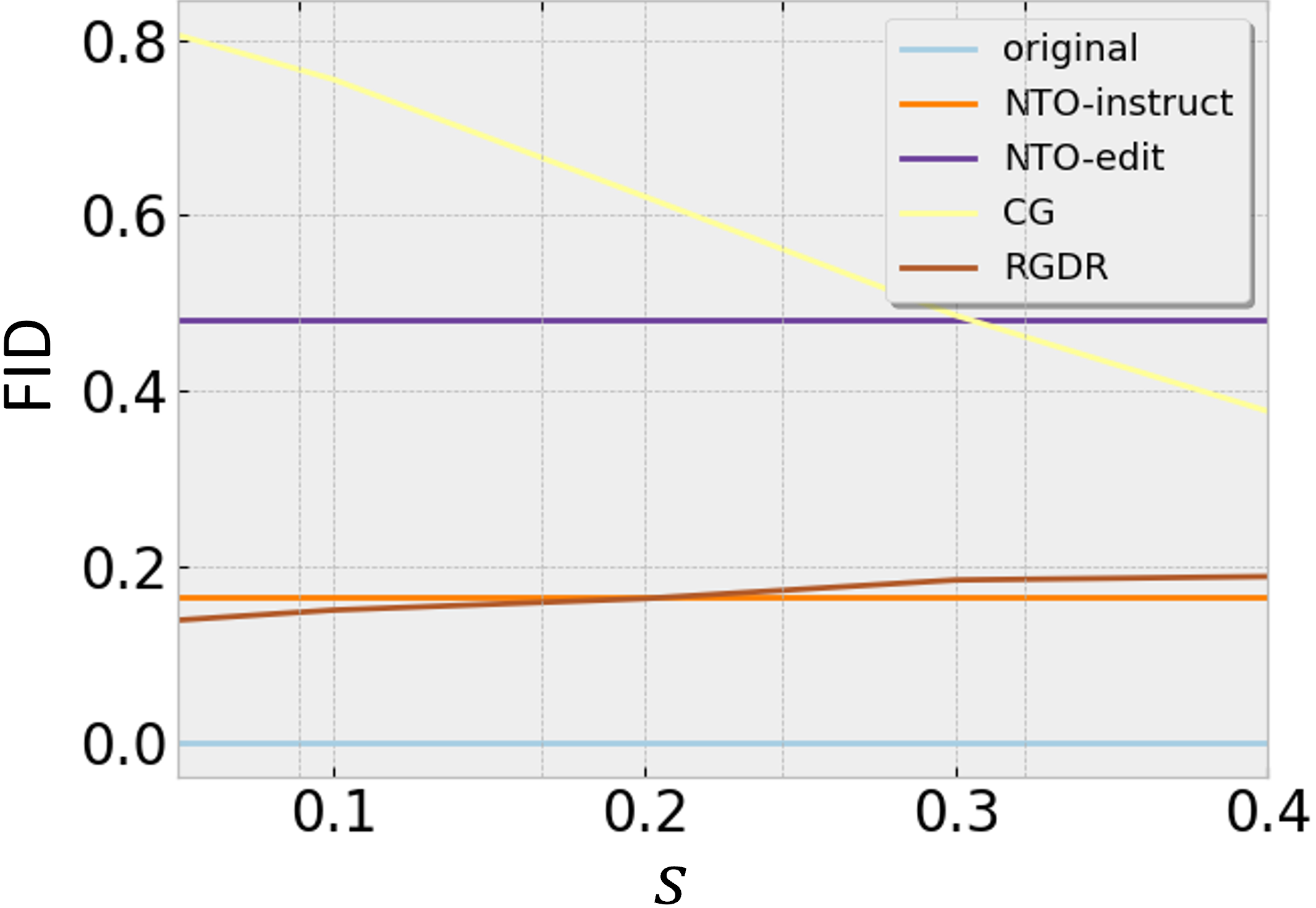}
        \caption{FID}
        \label{fig:criteria_satisfaction_appendix:fid}
    \end{subfigure}
    \begin{subfigure}{.33\textwidth}
        \centering
        \includegraphics[width=.95\linewidth]{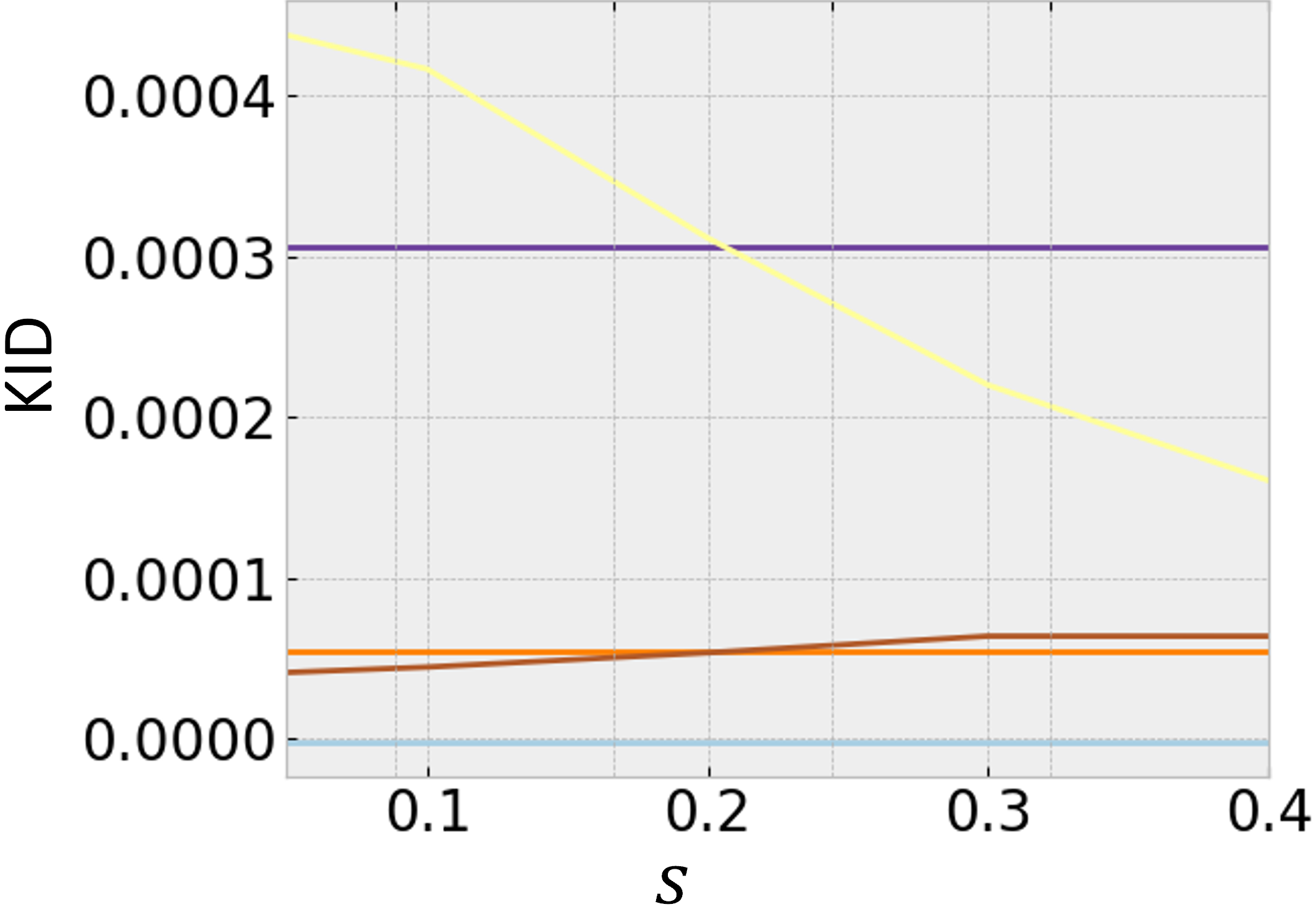}  
        \caption{KID}
        \label{fig:criteria_satisfaction_appendix:kid}
    \end{subfigure}
    \begin{subfigure}{.33\textwidth}
        \centering
        \includegraphics[width=.95\linewidth]
        {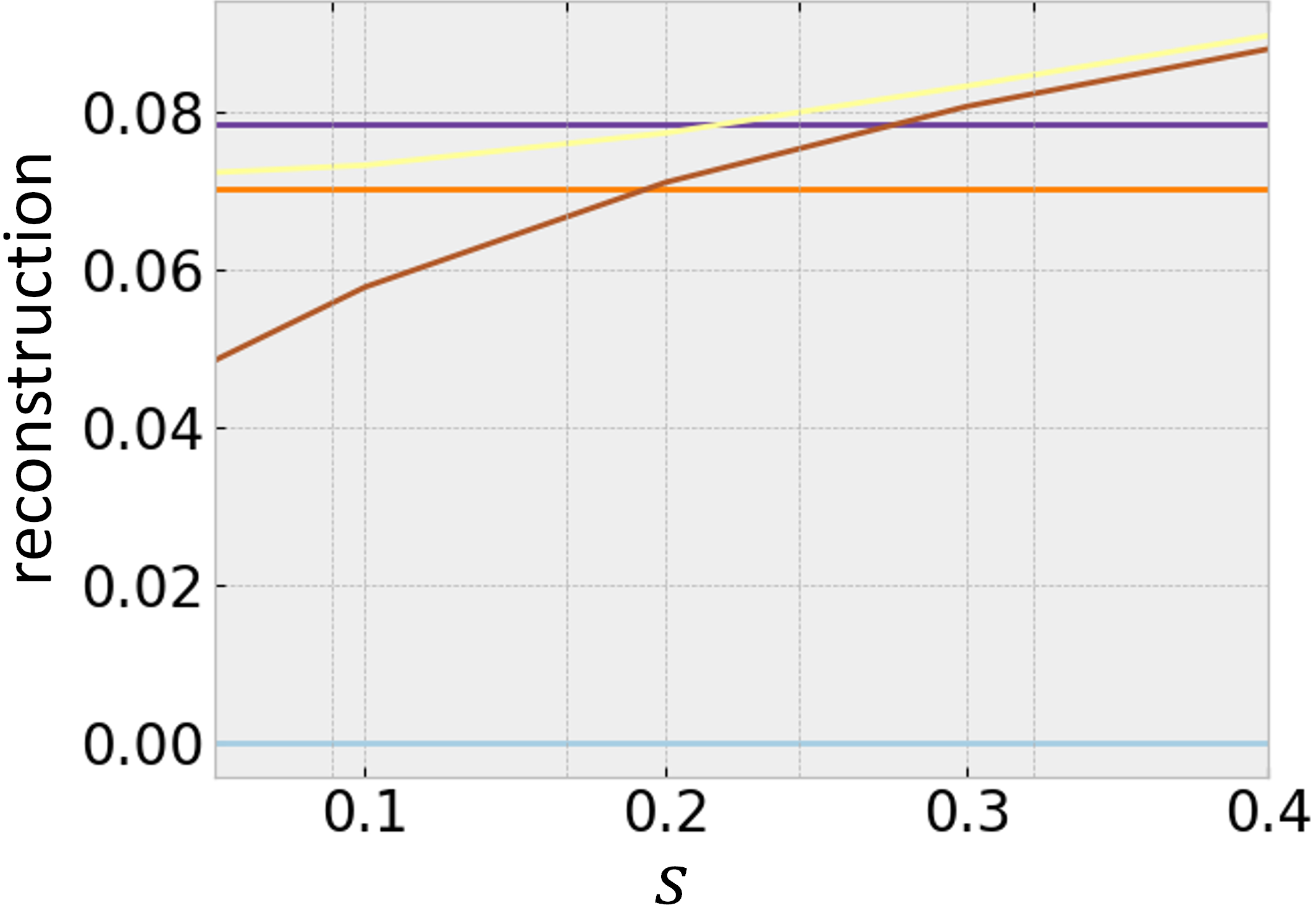}  
        \caption{reconstruction (L1)}
        \label{fig:criteria_satisfaction_appendix:l1}
    \end{subfigure}
    \begin{subfigure}{.33\textwidth}
        \centering
        \includegraphics[width=.95\linewidth]{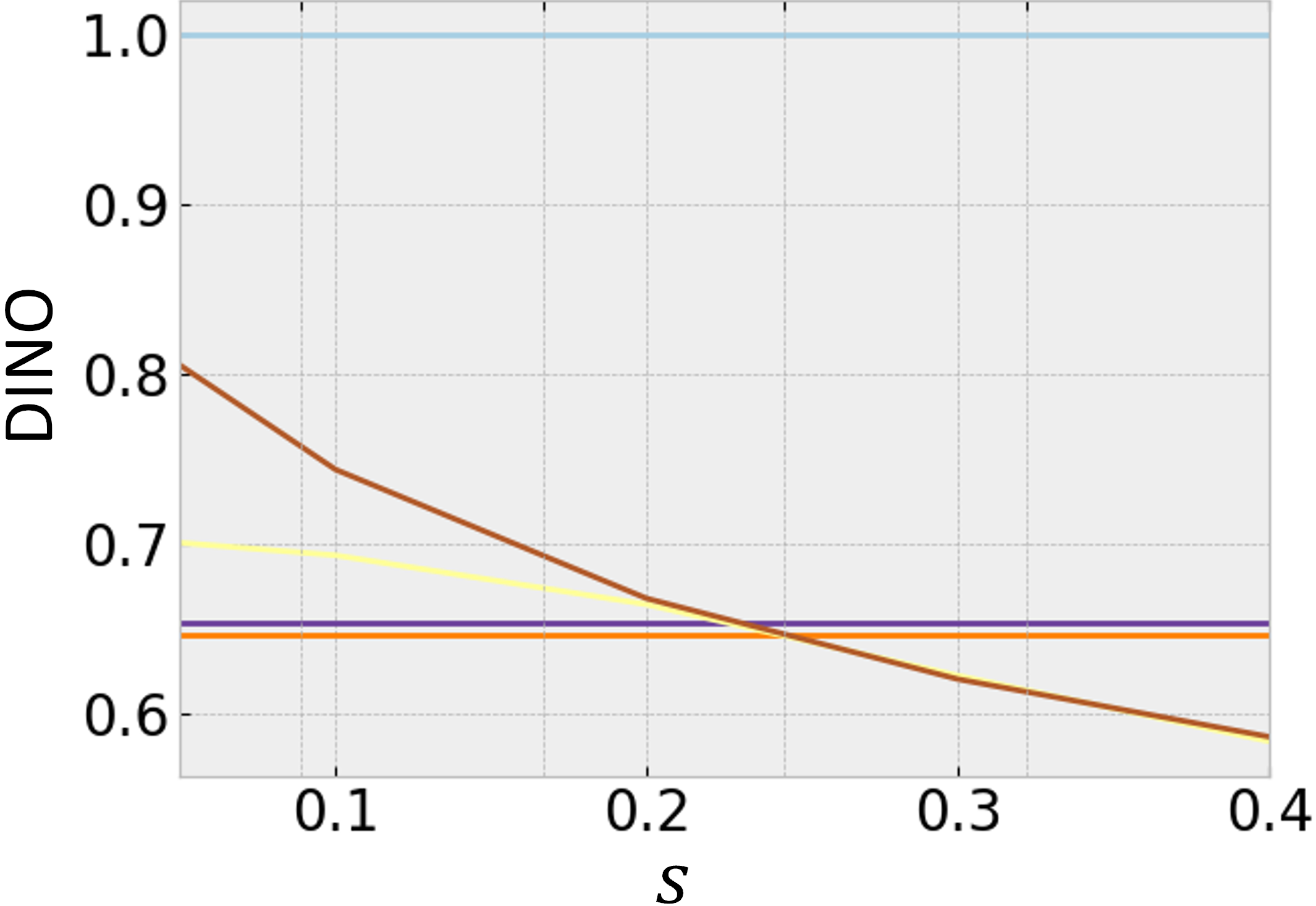}  
        \caption{content preservation (DINO)}
        \label{fig:criteria_satisfaction_appendix:dino}
    \end{subfigure}

    \caption{\textbf{Criteria alignment analysis.} The plots show how the metrics evolve as the weighting of the objective terms that introduce emotional changes in the images increases. The x-axis indicates the values of $s$ for \diff{} and CG. The y-axes show the metric values.}
    \Description{Criteria alignment analysis for four diffusion-based ablation variants (RGDR, NTO-instruct, NTO-edit, CG) across increasing values of the adaptation-strength parameter $s$ from 0.05 to 0.4, with original as a flat reference line in each panel. Panel (a) valence: RGDR declines steadily toward the emotional reference of 0.5, NTO-edit rises above original, and NTO-instruct and CG stay close to original with little change. Panel (b) arousal: RGDR declines steadily while the other three variants remain roughly flat near or above original. Panel (c) FID and panel (d) KID: all variants stay low and close to original across the full range of $s$, indicating maintained visual quality. Panel (e) reconstruction error (L1): RGDR increases steadily with $s$, exceeding the other variants at higher values, while NTO-instruct, NTO-edit, and CG remain lower and closer to original. Panel (f) content preservation (DINO score): RGDR starts high and declines sharply as $s$ increases, dropping below the other three variants, which remain relatively stable near or above 0.65.}
    \label{fig:criteria_satisfaction_appendix}
\end{figure*}

\subsubsection{Results}
We evaluate the ablated variants against the same metrics used in Sec.~\ref{sec:technical-evaluation}, reading each variant as a test of what the removed component contributes.  

Removing text conditioning retains emotional steering: OLS models confirm that higher guidance weights for CG are associated with reductions in valence ($\beta_1 = -0.0960$, $p < 0.001$) and arousal ($\beta_1 = -0.0282$, $p = 0.003$), directionally consistent with the \emref{}.
The cost appears in \crit{3} and \crit{4}: CG shows higher KID and FID (Fig.~\ref{fig:criteria_satisfaction_appendix:kid},
\ref{fig:criteria_satisfaction_appendix:fid}), increased reconstruction error (Fig.~\ref{fig:criteria_satisfaction_appendix:l1}), and visible qualitative degradation (see Supplementary Material). Caption conditioning via CFG is therefore what anchors the edit to the original image.

Removing regressor guidance yields two distinct failure patterns. Re-rendering with a neutralized caption (\cfgedit{}) reduces arousal but moves valence \emph{away} from the \emref{}, rising above that of the original images (Fig.~\ref{fig:criteria_satisfaction_appendix:valence}), and yields lower technical image quality than \diff{}. The instruction-guided variant (\cfg{}) preserves KID and FID comparable to \diff{}, but fails to steer valence and arousal toward the \emref{} to the same extent. Neither text-only variant reliably reaches the emotional target, which the
regressor gradient supplies.

\diff{}, combining both, is the only variant satisfying \crit{1}--\crit{4} simultaneously: it reaches the \emref{} on both dimensions ($\beta_1 = -0.0401$, $p = 0.001$ for valence; $\beta_1 = -0.0513$, $p = 0.005$ for arousal; see Tab.~\ref{tab:ols-full} for the corresponding intercepts and $R^2$), maintains KID and FID comparable to the text-conditioned variants, and is the only diffusion-based approach exceeding a DINO score of 0.7 at low $s$ (Fig.~\ref{fig:criteria_satisfaction_appendix:dino}). Its higher reconstruction error relative to the other diffusion variants occurs only outside our operating range ($s > 0.2$). Note, however, that diffusion approaches using CG occasionally introduced artifacts where CG gradients conflict with CFG (Sec.~\ref{sec:limitations-method}).

\subsection{Analyzing changes in emotional image properties}
\label{apx:image-property-analysis}
Table~\ref{tab:attribute-metrics} lists the low-level image attributes used to characterize edits, and summarizes their established association with valence and/or arousal in the affective-imagery literature, which motivates using them as a diagnostic of whether our approaches act along their intended axis. 
The metrics were computed over the COCO 2017 validation set.

\begin{table}[tbh]
\setlength{\tabcolsep}{4pt}
\centering
\small
\caption{\textbf{Image attribute metrics and their established attribute--affect associations from prior literature.} Citations in the \emph{Description} column refer to the source of the computational metric; citations in \emph{References} refer to studies establishing the attribute's association with affect. $(+)$/$(-)$ denote a positive/negative association.}
\label{tab:attribute-metrics}
\begin{tabular}{@{}p{0.18\linewidth}p{0.36\linewidth}ccp{0.22\linewidth}@{}}
\toprule
Attribute & Description & Valence & Arousal & References \\
\midrule
Colorfulness & Std. dev.\ plus weighted mean of distances from the mean in the LAB $a$/$b$ channels \cite{hasler2003measuring}. & $(+)/(-)$\textsuperscript{a} & $(+)$ & \cite{bekhtereva2017bringing,cano2009affective} \\
Saturation & Mean of the HSV saturation channel \cite{hasler2003measuring}. & $(+)$ & $(+)$ & \cite{simola2015affective, redies2020global} \\
Brightness & Mean grayscale pixel intensity \cite{goetschalckx2019ganalyze}. & $(+)$ & - & \cite{specker2018universal, gao2007analysis} \\
Contrast & Band-limited contrast measure \cite{peli1990contrast}. & $(+)$ & - & \cite{yang2020emotion} \\
Blur & No-reference perceptual blur metric \cite{crete2007blur}. & $(-)$ & $(-)$ & \cite{de2010effects} \\
Edge orient.\ entropy & First-order entropy of edge orientations \cite{redies2017high}. & - & $(+)$ & \cite{redies2017high} \\
Low-frequency\ color/lum.\ change & Z-score transformed wavelet coefficients of RGB and grayscale images \cite{delplanque2007spatial}. & - & $(+)$ & \cite{delplanque2007spatial} \\
Symmetry & Left-right mirror symmetry via VGG16 filter responses (color, edges, texture, higher-order) \cite{brachmann2016using}. & $(+)$ & $(+)$ & \cite{brachmann2017computational,bertamini2013implicit,bertamini2018neural,bertamini2019symmetry,makin2012symmetry} \\
\bottomrule
\multicolumn{5}{@{}p{0.95\linewidth}@{}}{\footnotesize\textsuperscript{a}Colorfulness amplifies valence in both directions rather than shifting it monotonically.}
\end{tabular}
\end{table}

\begin{figure*}[t]
    \centering
    \begin{subfigure}{.24\textwidth}
        \centering
        \includegraphics[width=.95\linewidth]
        {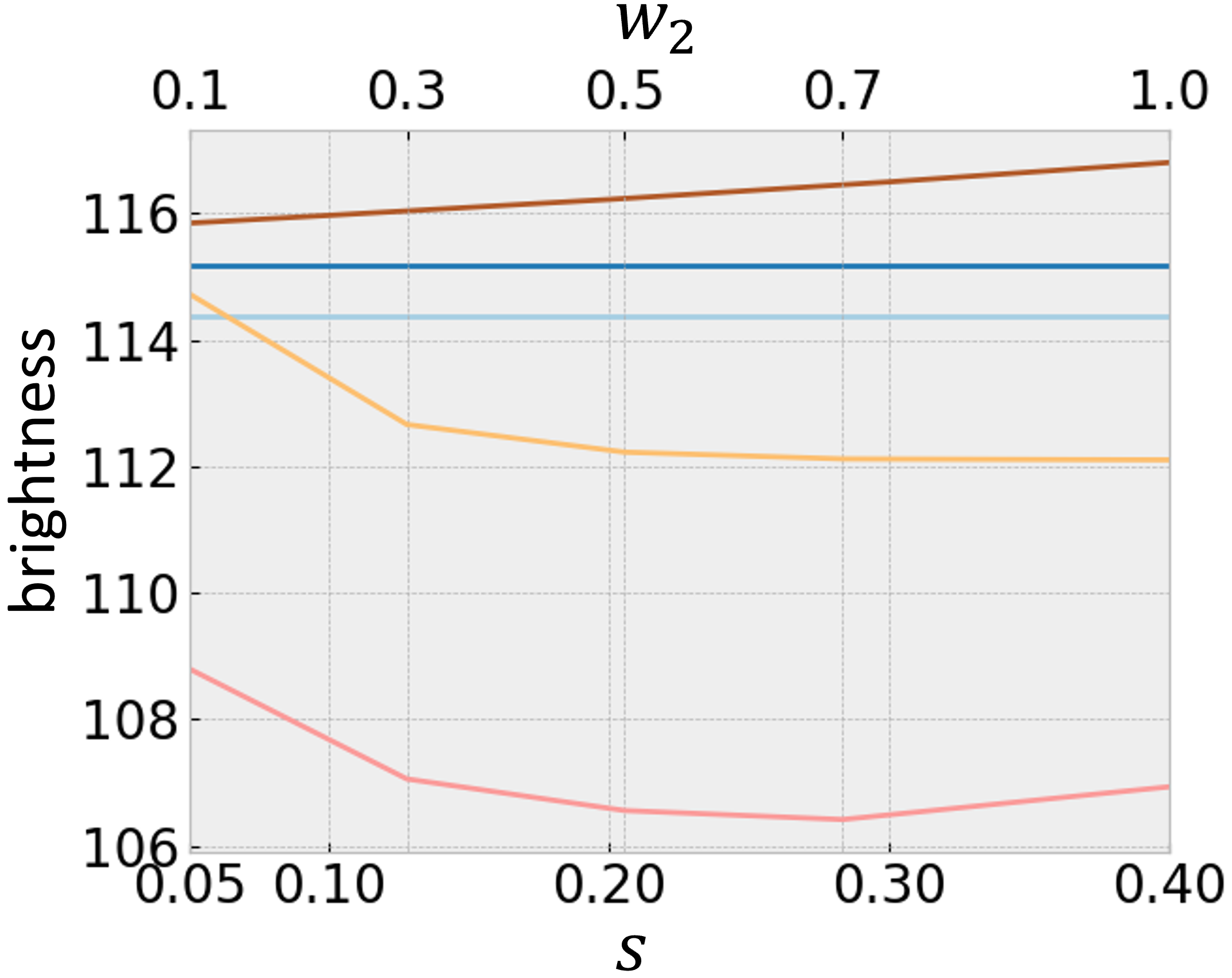}  
        \caption{brightness}
    \end{subfigure}
    %\hspace{20mm}
    \begin{subfigure}{.24\textwidth}
        \centering
        \includegraphics[width=.95\linewidth]
        {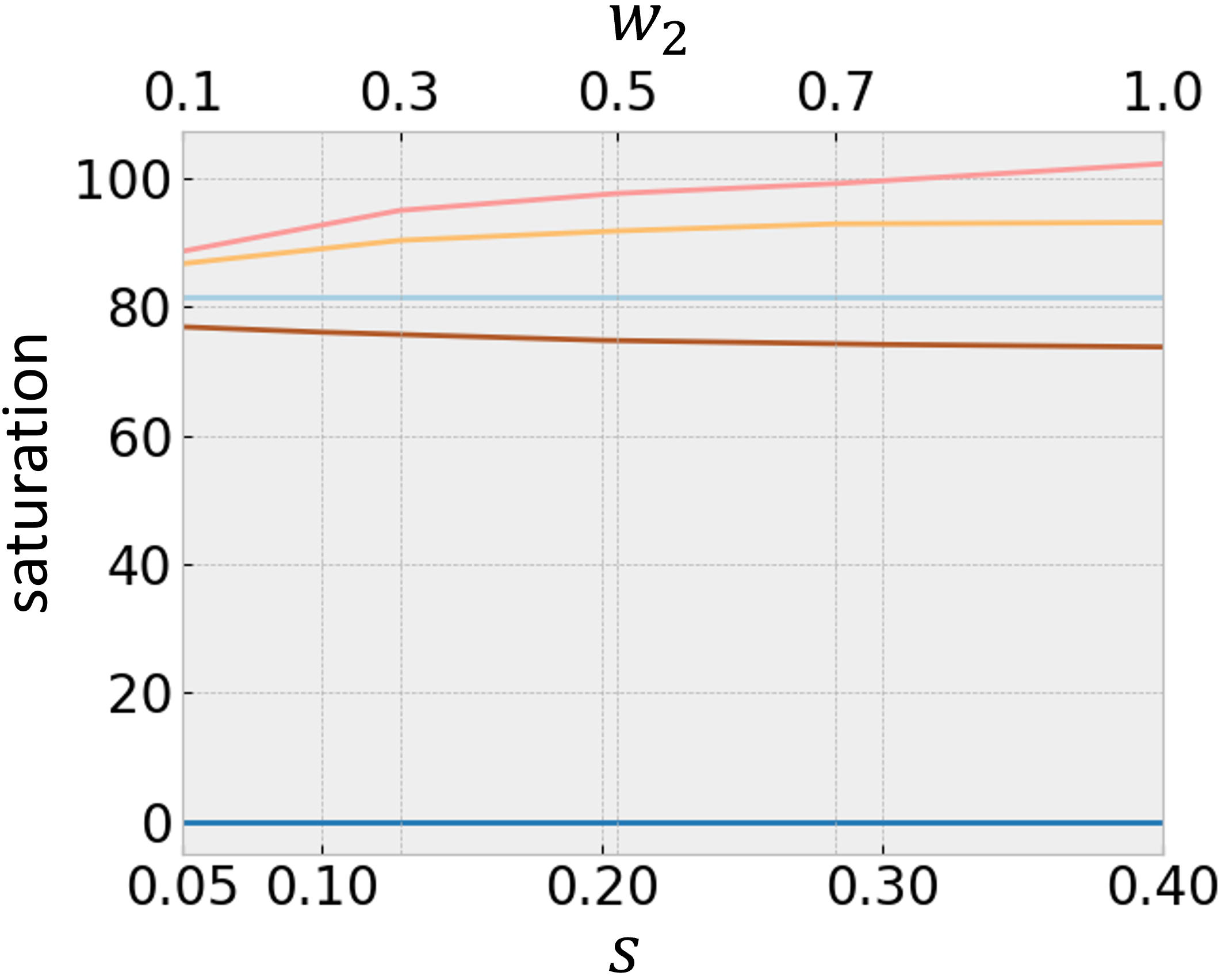}
        \caption{saturation}
    \end{subfigure}
    %\hspace{20mm}
    \begin{subfigure}{.24\textwidth}
        \centering
        \includegraphics[width=.95\linewidth]
        {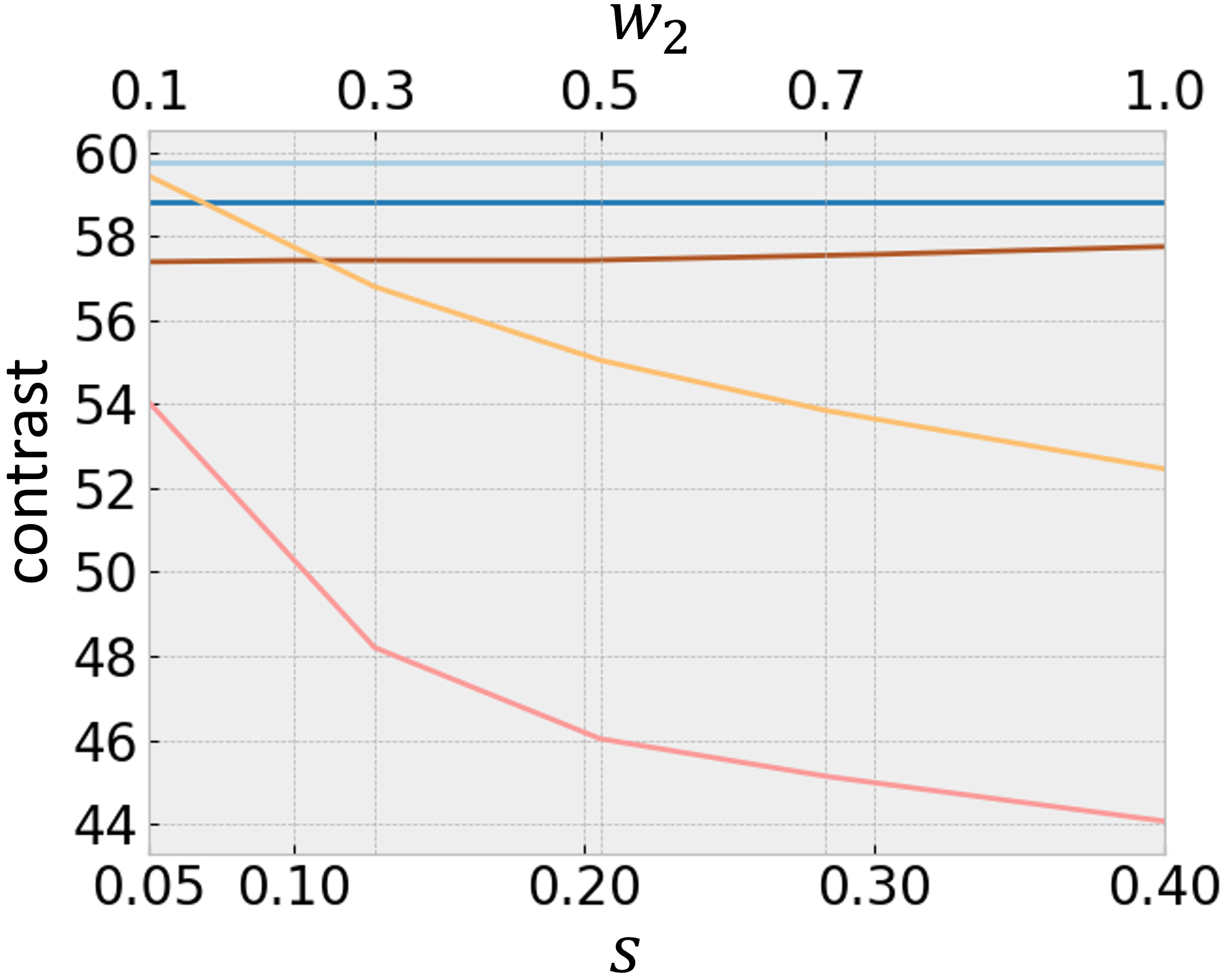}  
        \caption{contrast}
    \end{subfigure}
    %\hspace{20mm}
    \begin{subfigure}{.24\textwidth}
        \centering
        \includegraphics[width=.95\linewidth]
        {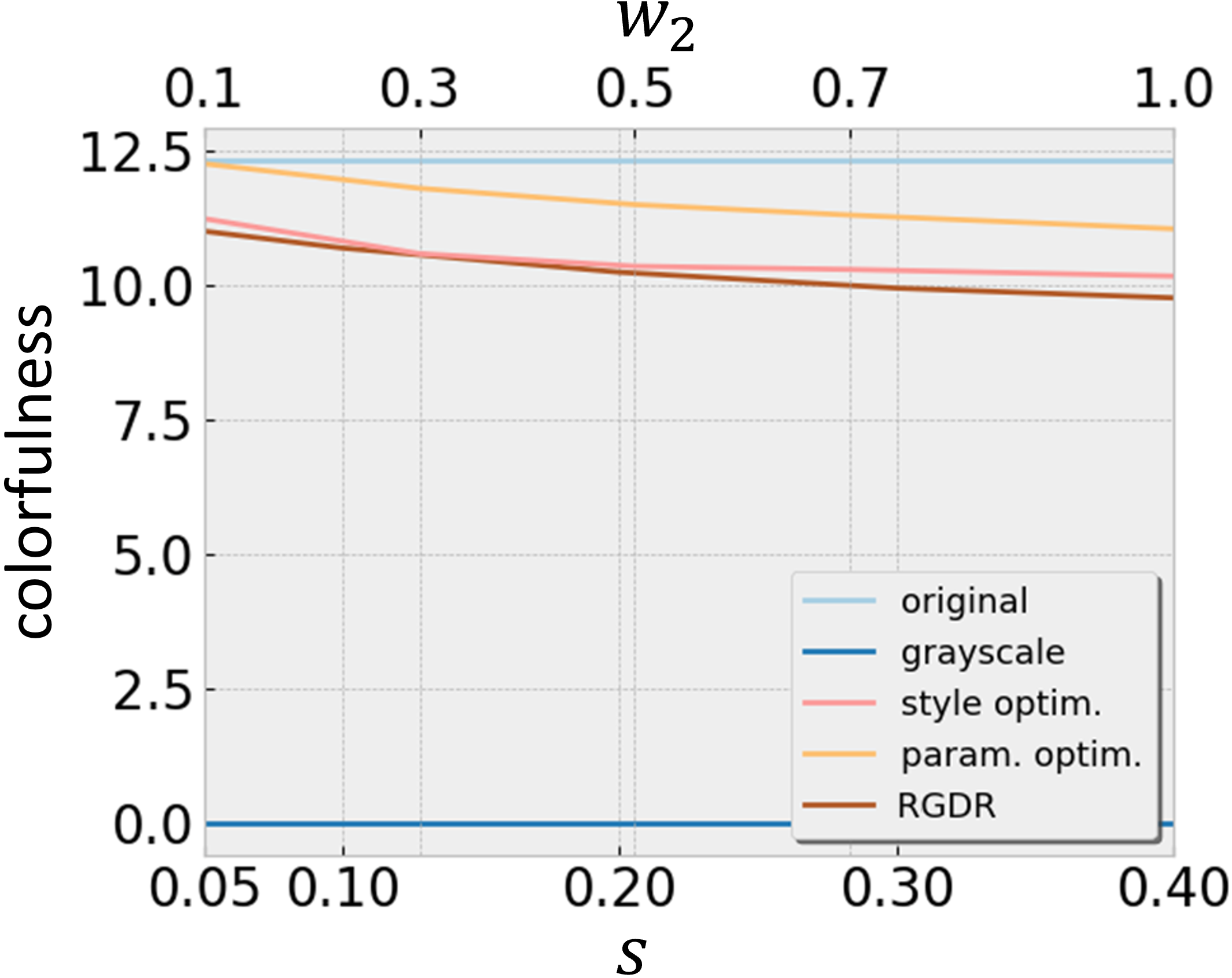}  
        \caption{colorfulness}
    \end{subfigure}
    \begin{subfigure}{.24\textwidth}
        \centering
        \includegraphics[width=.95\linewidth]
        {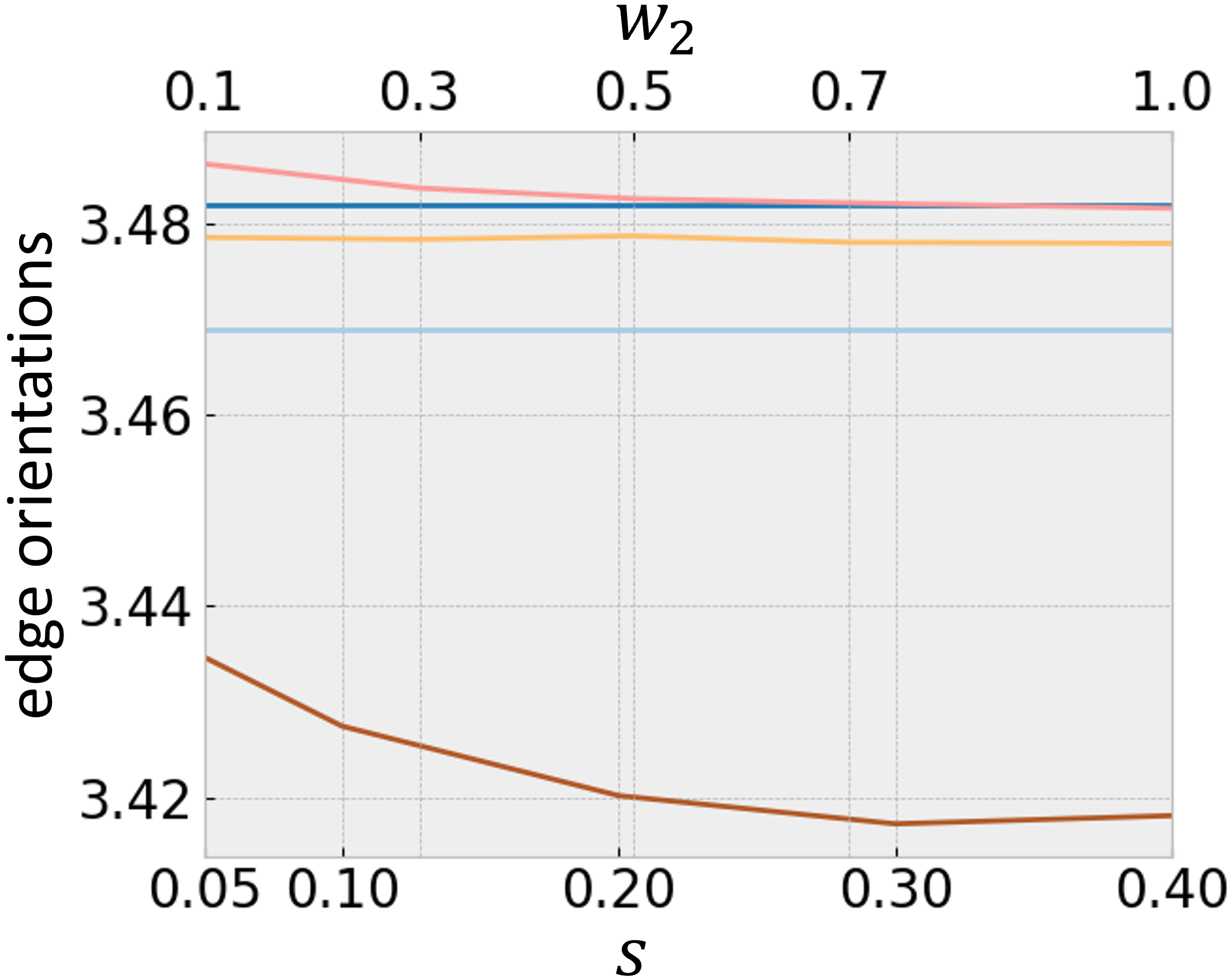}  
        \caption{FOE of edge orientation}
    \end{subfigure}
    \begin{subfigure}{.24\textwidth}
        \centering
        \includegraphics[width=.95\linewidth]
        {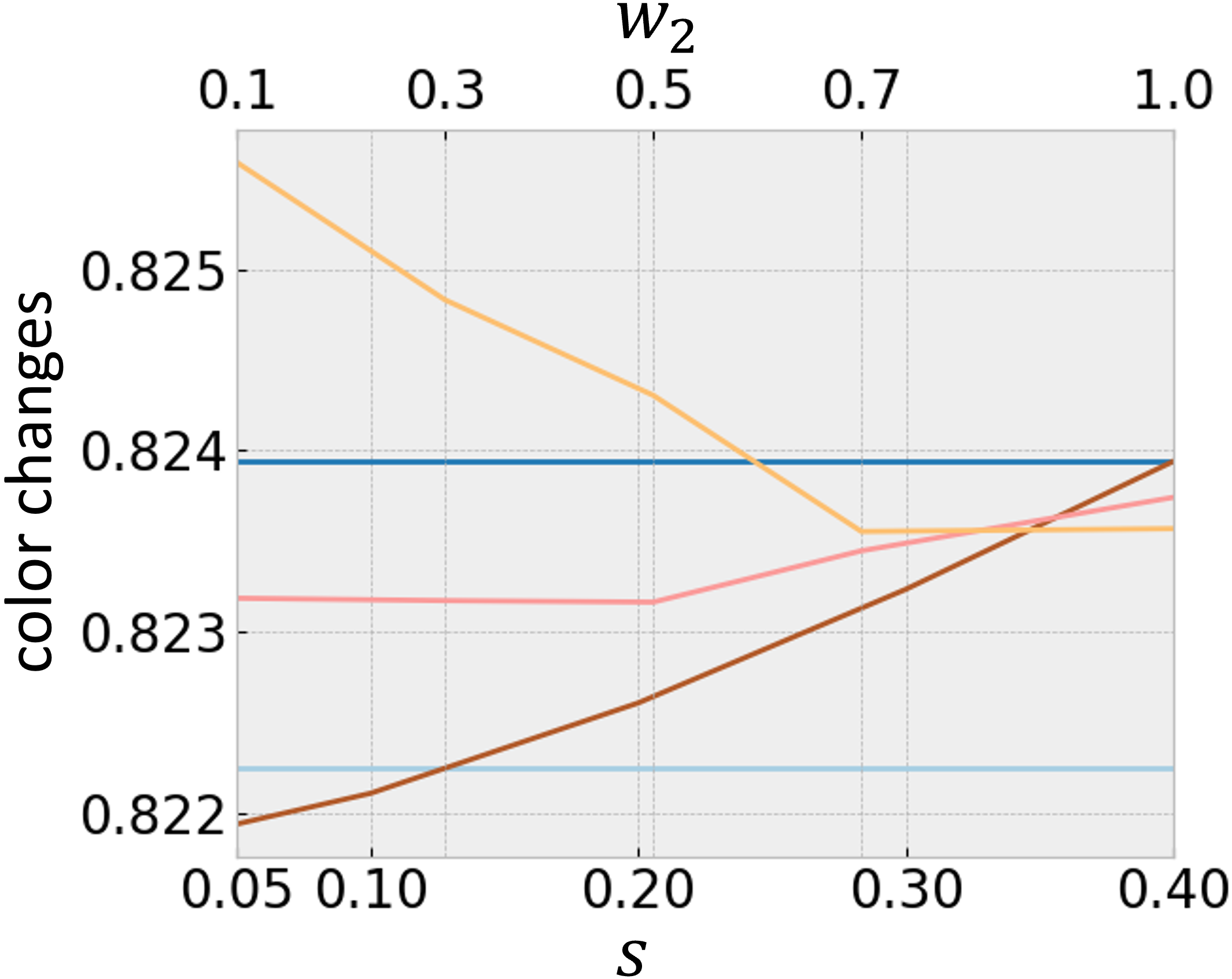}  
        \caption{low-freq. color changes}
    \end{subfigure}
    % %\hspace{20mm}
    \begin{subfigure}{.24\textwidth}
        \centering
        \includegraphics[width=.95\linewidth]
        {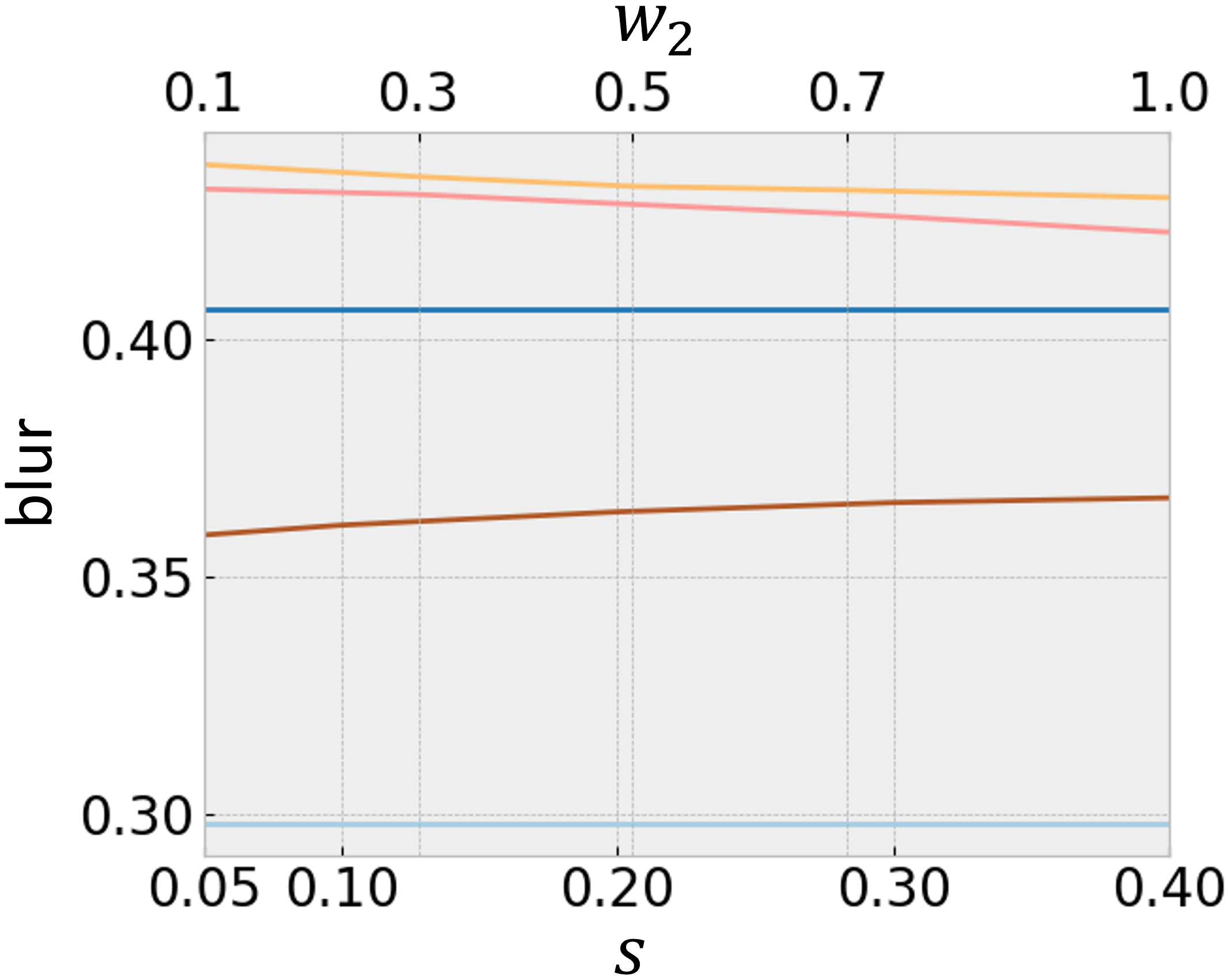}  
        \caption{blur}
    \end{subfigure}
    %\hspace{20mm}
    \begin{subfigure}{.24\textwidth}
        \centering
        \includegraphics[width=.95\linewidth]
        {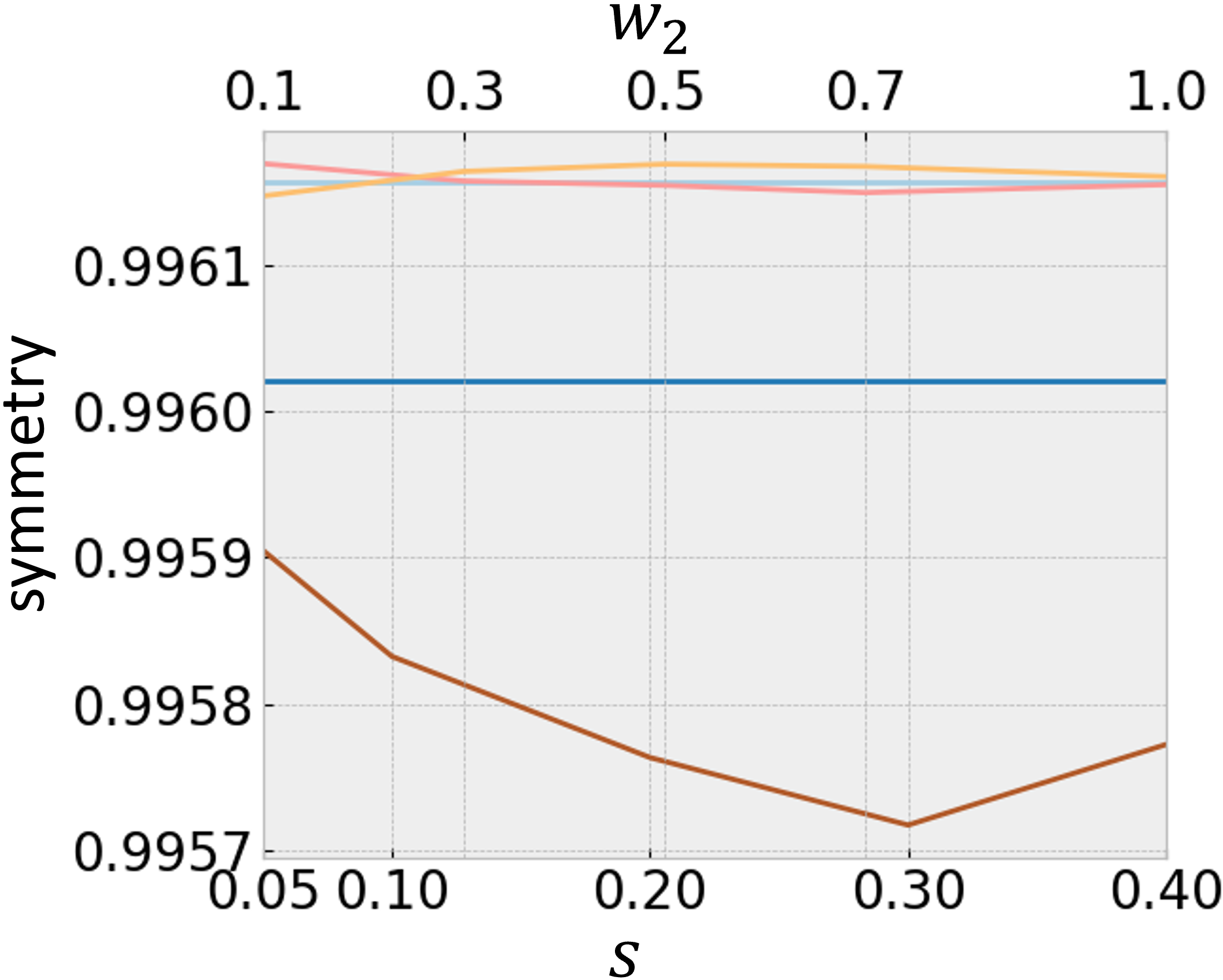}
        \caption{symmetry}
    \end{subfigure}
    \caption{Analyzing changes in image properties: The plots show how the metrics evolve as the weighting of the objective terms that introduce emotional changes in the images increases. In each plot, the upper x-axis represents the values of $w_2$ for \param{} and \style{}, while the lower x-axis indicates the values of $s$ for \diff{}. The y-axes show the metric values.}
    \Description{Eight line plots showing image property changes under different optimization methods. Metrics include (a) brightness, (b) saturation, (c) contrast, (d) colorfulness, (e) frequency of edge orientation, (f) low-frequency color changes, (g) blur, and (h) symmetry. Original and grayscale remain nearly constant, while style optimization, parametric optimization, and SGDR vary with parameters. Contrast and brightness show the largest shifts, whereas blur and symmetry remain stable across methods.}
    \label{fig:changes_image_properties}
\end{figure*}

\subsubsection{Results}
Figure~\ref{fig:changes_image_properties} shows how each attribute changes relative to the originals as the emotional adaptation weight increases; expected directions are given in Table~\ref{tab:attribute-metrics}.
Three attributes matched their expected direction across all four approaches: \emph{contrast} decreased, consistent with the originals' slightly positive average valence (Fig.~\ref{fig:changes_image_properties}c); \emph{colorfulness} decreased (Fig.~\ref{fig:changes_image_properties}d); and \emph{blur} increased (Fig.~\ref{fig:changes_image_properties}g).

For the remaining attributes, approaches diverged. \emph{Brightness} decreased for \style{} and \param{} as expected, but not for \gray{} or \diff{} (Fig.~\ref{fig:changes_image_properties}a). \emph{Saturation} decreased for \gray{} and, to a lesser extent, \diff{}, while \style{} and \param{} increased it (Fig.~\ref{fig:changes_image_properties}b). \emph{Edge-orientation entropy} declined only for \diff{} (Fig.~\ref{fig:changes_image_properties}e). \emph{Symmetry} decreased for \diff{} and \gray{}, consistent with the \emref{}, but not for \style{} or \param{}, though all differences were minor (Fig.~\ref{fig:changes_image_properties}h). 

Contrary to expectations, \emph{low-frequency color/luminance change} increased slightly across all approaches (Fig.~\ref{fig:changes_image_properties}f).
\section{Image Rating Study Details}
\label{apx:image-rating}
This section reports additional design details and results for the image rating study, supplementing Section~\ref{sec:image-study}.

\subsection{Study design}
\label{apx:im-design}

\subsubsection{Stimulus selection}
\label{app:rating-stimuli}
\begin{figure}[tbh]
	\centering
    \includegraphics[width=0.95\linewidth]{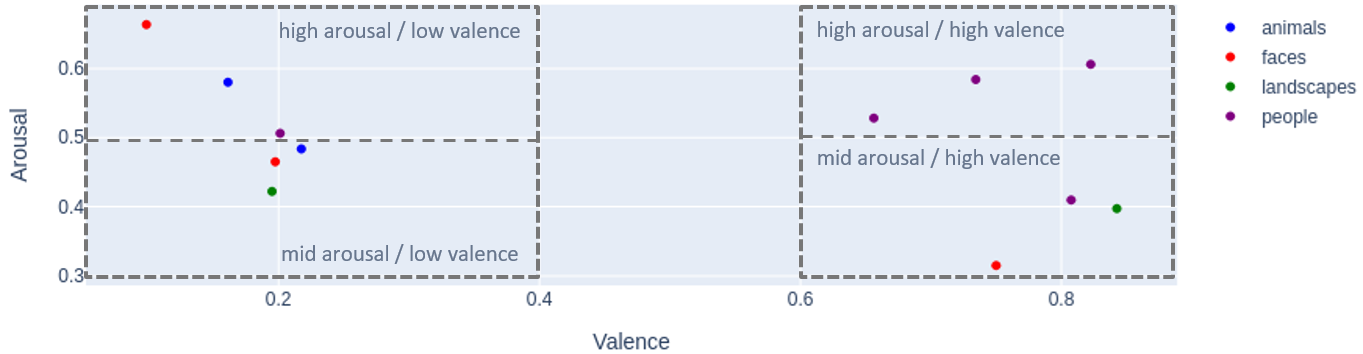} 
    \caption{
    \textbf{Location of stimuli within the CMA space}. To select images from a spectrum of valence and arousal values, they were drawn from four clusters: mid arousal/low valence, high arousal/low valence, mid arousal/high valence, and high arousal/high valence.
    }
    \Description{Scatter plot of arousal versus valence with points grouped into four quadrants: high arousal/low valence, mid arousal/low valence, high arousal/high valence, and mid arousal/high valence. Data points represent categories including animals, faces, landscapes, and people. Most points fall into either high arousal/high valence or high arousal/low valence regions.}
    \label{fig:selected_stimuli}
\end{figure}

Following Mould et al.~\cite{mould2012emotional}, we divided the valence--arousal space of the CMA into distinct regions and excluded regions positioned near the optimization target (low arousal / neutral valence), as adaptations are minimal by design for images already located there. The final selection was drawn from four clusters: mid arousal / low valence, high arousal / low valence, mid arousal / high valence, and high arousal / high valence (Figure~\ref{fig:selected_stimuli}).
We selected three images selected per cluster, yielding 12 stimuli. We additionally sampled 5 images from across the full valence--arousal spectrum for the calibration phase.

Within each block, we started and ended with lower-arousal, positive images to ease participants into a new condition and leave them with a positive sentiment. Each block began and concluded with images from the mid arousal / high valence cluster, then proceeded counterclockwise through the remaining clusters in the valence--arousal space.

\subsubsection{Stimulus IDs}
\label{apx:im-naps-ids}
Because the NAPS usage terms prohibit publishing its images, the stimuli cannot be reproduced here. Table~\ref{tab:naps-ids} lists the image IDs used in each phase of the study.

\begin{table}[tbh]
\centering
\caption{\textbf{Stimuli IDs.} NAPS and NAPS ERO image IDs used in the calibration phase and
the experimental block.}
\label{tab:naps-ids}
\begin{tabular}{llp{9cm}}
\toprule
\textbf{Phase} & \textbf{Dataset} & \textbf{Image IDs} \\
\midrule
Calibration & NAPS & Animals\_104\_h, Faces\_001\_h, Faces\_309\_h, Objects\_125\_h \\
 & NAPS ERO & Opposite-sex\_couple\_019\_v \\
\midrule
Experimental block & NAPS & People\_190\_h, People\_196\_h, Animals\_077\_h, Faces\_365\_v, \mbox{People\_222\_h}, Faces\_283\_h, Animals\_025\_h, Landscapes\_026\_h, Faces\_109\_v, Landscapes\_178\_h \\
 & NAPS ERO & Female\_017\_v, Opposite-sex\_couple\_017\_v \\
\bottomrule
\end{tabular}
\end{table}

\subsubsection{Task description}
\label{apx:im-task}
The following task description was shown prior to the study:

\begin{quote}
\itshape
In this experiment, you will report the emotions that are evoked by images displayed on the screen. You will see the same picture multiple times during the experiment, but please try to answer each question as if you are seeing the picture for the first time.
\end{quote}

\subsubsection{Power analysis}
\label{apx:im-power}
We conducted an a priori power analysis to establish the minimum sample size needed to test our hypotheses. 
To achieve 95\% statistical power at $\alpha = 0.05$ for a repeated-measures ANOVA with five conditions and a Cohen’s f of 0.2, we estimated a required sample size of $N = 48$. The chosen effect size was based on reported values in similar studies (0.25–1.5; \cite{de2010effects, bekhtereva2017bringing}).
We deliberately assumed a smaller effect ($f = 0.2$) than prior work to ensure the study remained adequately powered even if our adaptations produced weaker effects than previously observed.

\subsection{Additional results}
This subsection reports supplementary analyses for the image rating study: a robustness check for image-level variability, followed by additional descriptive statistics for the study's main results.

\subsubsection{Image-Level Variability}
\label{apx:im-rating-by-image}
The analyses reported in the main paper average across our 12 stimuli, treating them as fixed. Because the stimuli were deliberately sampled from four regions of the valence-arousal space (Sec.~\ref{sec:image-study-design}), image-specific characteristics could in principle drive the observed condition effects rather than the adaptations generalizing across stimuli. To check this, we repeated the primary contrasts (\cgray{}, \cdiff{}, \cstyle{}, \cparam{} vs.\ \cdefault{}) with image, rather than participant, as the unit of analysis: for each of the 12 images, we computed its mean rating per condition across all 56 participants, and tested these image-level means using Holm-corrected Wilcoxon signed-rank tests. We report Wilcoxon rather than paired $t$-tests here, as Shapiro-Wilk tests of normality are themselves underpowered at $n=12$, making the assumption underlying the parametric test difficult to verify at this sample size.

Given the reduced sample ($n=12$), this analysis is underpowered relative to the participant-level tests and is intended as a robustness check rather than a substitute for them. Effects were directionally consistent with the participant-level analysis across all 12 contrasts. Large effects remained significant at the image level: \cgray{} \pval{=}{.01} and \cdiff{} \pval{=}{.032} both reduced arousal relative to \cdefault{}. 
\emph{Grayscale} \pval{=}{.01}, \cstyle{} \pval{=}{.01}, and \cparam{} \pval{=}{.01} all reduced quality relative to \cdefault{}, while \cdiff{} still did not \pval{=}{.224}.

At the image level, valence effects were not consistent across conditions: only \cstyle{} showed a significant shift \pval{=}{.04}, while \cgray{} \pval{=}{.293} and \cdiff{} \pval{=}{.293} were not significant.
These results suggest that while \cdiff{}'s valence effect is robust and highly significant at the participant level ($N=56$), its magnitude varies across images and does not generalize with equal strength to every stimulus in our set.

\subsubsection{Observation time}
Participants viewed each emotional image for a minimum of five seconds, after which they could continue viewing or press space to proceed.
We recorded \emph{observation time} as the mean additional viewing duration per image beyond this five-second minimum (Tab.~\ref{tab:msd-full-image-rating}).
No significant difference was found in the case of observation time \anova{2.12}{116.51}{1.46}{=}{.23}{.03}.
% Observation time per image
% Anova Table (Type 3 tests)
%           num Df den Df    MSE      F      pes Pr(>F)
% condition 2.1184 116.51 43.487 1.4665 0.025971 0.2344
%
This task, however, is poorly suited to detecting a disengagement effect: participants were paid a flat fee rather than by time spent, giving them a financial incentive to move through images quickly. 
We therefore do not treat this null as evidence against a behavioral effect, and turn to the social media study for a setting better suited to detecting one.

\subsubsection{Descriptive statistics}
\label{apx:im-descriptive}
In Table~\ref{tab:msd-full-image-rating}, we show mean and standard deviation values of all dependent variables.

\begin{table}[tbh]
  \centering
   \small
    \caption{\textbf{Rating statistics.} Mean ratings (SD in parentheses) for valence, arousal, perceived image quality, and observation time per condition. Asterisks mark paired $t$-tests against \cdefault{}; daggers mark exploratory differences from \cgray{} ($^{*}p<.05$, $^{**}p<.01$, $^{***}p<.001$, $^{\dagger}p<.001$). All Holm-corrected within each family.}
  \label{tab:msd-full-image-rating}
  \setlength{\tabcolsep}{4pt}
  \begin{tabular}{@{}lcccc@{}}
    \toprule
    \textbf{Condition}   & \textbf{Valence} & \textbf{Arousal} & \textbf{Quality} & \textbf{Observation Time} \\
    \midrule
    \cdefault{} & 4.43 (0.76) & 4.39 (1.36) & 3.43 (0.86) & 4.14 (4.23) \\
    \cgray{}    & 4.23 (0.64)$^{**}$ & 3.93 (1.31)$^{***}$ & 3.00 (0.88)$^{***}$ & 4.33 (5.56) \\
    \cdiff{}    & 5.23 (0.89)$^{***\dagger}$ & 4.00 (1.45)$^{**}$ & 3.27 (0.64) & 6.12 (10.04) \\
    \cstyle{}   & 4.23 (0.73)$^{**}$ & 4.33 (1.47)$^{\dagger}$ & 3.00 (0.82)$^{***}$ & 4.75 (6.03) \\
    \cparam{}   & 4.32 (0.69) & 4.34 (1.51)$^{\dagger}$ & 3.05 (0.70)$^{***}$ & 4.80 (5.97) \\
    \bottomrule
  \end{tabular}
\end{table}

\section{Social Media Study Details}
\label{apx:social-media}
This section reports the preregistration and its deviations, followed by additional design details and results for the social media study, supplementing the summary in Section~\ref{sec:social-study}.

\subsection{Preregistration and deviations}
\label{apx:sm-deviations}
The social media study followed a revised registration [anonymized] and adhered to it in design, task, stimuli, per-condition sample size, and measures, with the following deviations. 
(i)~The parametric- and style-optimization conditions were omitted after the image rating study indicated weaker emotional modulation for these methods, reducing the design from five to three arms at the preregistered $n{=}64$ per condition. 
(ii)~For cost reasons, data collection proceeded in two stages: participants in the \cdefault{} and \cgray{} conditions were predominantly collected 13--20~June~2025 under concurrent randomization; the \cdiff{} condition was collected 4--7~July~2025, together with a small number of concurrently randomized control participants ($n{=}7$) on the final collection day.
% \cdiff{} contrasts therefore rely predominantly on non-concurrent controls.
Section~\ref{apx:sm-comparability} reports a comparability analysis, including the concurrently collected control sample. 
(iii)~Participants aged 18--74 were recruited rather than the preregistered 18--35 range. 
(iv)~Consistent with the registration's missing-data criterion (inclusion requires completing the task and the required questionnaires), the primary analysis sample comprises the $N{=}182$ participants whose feed data and both questionnaires could be linked; six participants with unlinkable questionnaire responses (token-entry errors, unrelated to condition; Fisher's exact \pvall{=}{.127}) were excluded, and three initially unlinkable responses were recovered through a mechanical timestamp-matching procedure. 
The preregistered exclusion criterion (``fail to correctly answer the comprehension and attention check questions'') is ambiguous as to whether failing either or both checks warrants exclusion.
All participants passed the attention check. Conclusions were unchanged with and without the 35 participants who failed the comprehension check, with the one exception detailed in Section~\ref{apx:sm-analysis-sample}. 
(v)~The preregistered distinct-emotions measure (Mikels' wheel) was not administered in either study.
(vi)~We report an ANOVA on the subsample of participants who left the feed before the limit, alongside the preregistered analysis.

\subsection{Study Design}

\subsubsection{Stimulus selection}
\label{app:sm-stimuli}
Stimuli were drawn from the Instagram Influencer Dataset~\cite{kim2020multimodal}, which contains 10,180,500 Instagram posts from 33,935 influencers, each comprising metadata and associated image files. Posts are classified into eight categories: beauty, family, fashion, fitness, food, interior, pet, and travel. We sampled 812 images from the fitness and travel categories.

Each image was processed eight times with \diff{} to account for quality variations in diffusion-generated images (discussed in Sec.~\ref{sec:limitations}).
One of the authors reviewed all versions of each image and selected the one that best preserved the original content without visible artifacts; images that showed artifacts in all eight versions (47 in total) were discarded, leaving 765 images.
Because many predicted valence and arousal values were already close to the \emref{}, not all images showed changes relative to their originals, and regressor-based predictions did not yield a clear decision boundary for identifying those that did.
The same author therefore conducted a second review, comparing each adapted image against its original and retaining those in which a visible change had been introduced.
This identified the 268 images used in the first feed.

We computed the order of posts using the quadrant-based approach of the image rating study. As valence and arousal values of images are distributed differently, we computed the 2D center of mass for valence and arousal across all post images, and defined four quadrants around it. We then created the feed by sampling from the quadrants such that the first element is drawn randomly from Q1, the second from Q2, then Q3 and Q4, until all posts are part of the feed. This ensures that the emotions elicited by the images are distributed similarly across the feed.

\subsubsection{Task description}
\label{apx:sm-task}
Participants were given the following task description prior to the study:

\begin{quote}
\itshape
In this experiment, you will browse two social media feeds for a total of five minutes. 
You will start in Feed 1. After a lock period, a button will appear in the top left, allowing you to switch to Feed 2. 
You can switch feeds only once. 
Hence, if you decide to switch from Feed 1 to Feed 2 after the lock period is over, you will not be able to go back to Feed 1. 
Your choice will not affect the total duration of the experiment: independently of whether or when you decide to switch from Feed 1 to Feed 2, the overall presentation period of the two feeds will be 5 minutes.
\end{quote}

\subsubsection{Feed}
\label{apx:sm-feed}
\begin{figure}[tbh]
	\centering
    \includegraphics[width=0.35\linewidth]{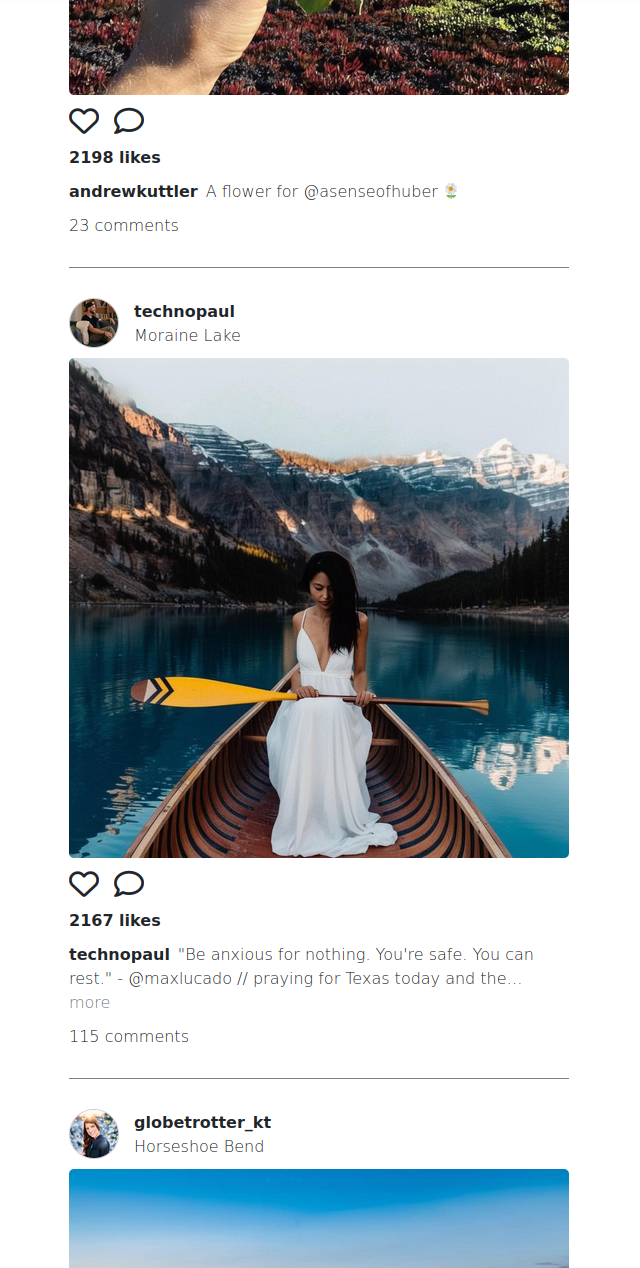} 
    \caption{
    \textbf{Social media feed.} Illustration of social media feed on our platform. 
    }
    \Description{Mockup of Instagram posts showing scenic outdoor photography. The central post depicts a woman in a white dress sitting in a canoe on a turquoise lake with mountains in the background. Other posts above and below feature landscape photos with short captions and engagement metrics.}
    \label{fig:feed}
\end{figure}

Figure \ref{fig:feed} depicts the social media feed of our experimental platform, as used in the social media study.

\subsubsection{Power analysis}
\label{apx:sm-power}
A priori power analysis indicated that a sample size of N = 159 would be required to achieve 80\% power to detect an effect of Cohen’s f = 0.25 at a significance level of $\alpha$ = .05 in a fixed-effect, omnibus one-way ANOVA with three factor levels (original, grayscale, and our image-editing approach). The chosen effect size was based on values reported in similar studies (f = 0.25-1.5; \cite{de2010effects, bekhtereva2017bringing}).
Despite having directional hypotheses, our preregistration specified the more conservative two-tailed independent-samples $t$-tests for post-hoc comparisons between image adaptation techniques [anonymized].
% \cite{gebhardt2025imageadaptations}.
Assuming an effect size of Cohen’s d = 0.5 (f = 0.25) and $\alpha$ = .05, a sample of N = 128 (n = 64 per group) would provide 80\% power. 
Adopting a more conservative approach, we based our target sample size on the pairwise comparison analysis, resulting in a required total of N = 192 participants (three conditions × 64 participants).

\subsection{Additional results}
This subsection reports additional results for the social media study, organized in four parts: (1) derivation of the final analysis sample, (2) a validity check of the two-stage sampling approach, (3) exploratory behavioral features, and (4) additional descriptive statistics for the study's main results.

\subsubsection{Analysis sample}
\label{apx:sm-analysis-sample}
We recruited participants via Prolific, prescreening for active social media users.
In total, 216 people were recruited to reach the target of 192 who passed the attention check; participants who failed were excluded automatically and replaced, without manual review of their responses.
Of these 192, four were excluded for not interacting with the feed ($n=3$) or due to a technical issue ($n=1$), leaving 188.
Nine of the 188 could not initially be linked to their questionnaire responses due to token-entry errors; three were recovered through a mechanical timestamp-matching procedure, and the remaining six were excluded per the registration's missing-data criterion.
The analysis sample thus comprises $N=182$ participants who completed the task and both questionnaires (\cdefault{}: 60, \cgray{}: 59, \cdiff{}: 63).
Linkage completeness was unrelated to condition (Fisher's exact \pvall{=}{.127}). 
Eleven participants (\textsc{original}: 1, \textsc{grayscale}: 4, \textsc{rgdr}: 6) never left the first feed within the 300-s window.
% and enter all duration analyses as right-censored observations. 
Thirty-five participants failed the comprehension check ($\chi^2(2) = 0.46$, \pvall{=}{.80} against condition), and, per the preregistered dual-reporting commitment, all analyses were computed with and without them. 
Conclusions for valence and quality were unchanged across all three sample definitions (full sample, $N=182$; comprehension-check-excluded, $n=147$; leavers-only, $n=171$). 
For arousal, conclusions were unchanged except for the exploratory \cgray{}--\cdiff{} contrast: Holm-corrected Wilcoxon tests reached significance in the full sample \pval{=}{.034} but not in the comprehension-check-excluded \pval{=}{.051} or leavers-only \pval{=}{.058} samples.
For duration, only the leavers-only sample yields a significant omnibus effect and post-hoc contrasts (Sec.~\ref{sec:social-study:exploratory}).

\subsubsection{Stage comparability}
\label{apx:sm-comparability}
Because the \cdiff{} condition was predominantly collected in a second stage (Sec.~\ref{apx:sm-deviations}), we examined the comparability of the two collection stages. Collection windows had the following structure: \cdefault{} and \cgray{} were interleaved throughout 13--20 June 2025, and stage two (4--7 July 2025) comprised the \cdiff{} condition plus seven concurrently randomized control participants on 7 July. 
On pre-treatment variables, stages showed small, non-significant differences: baseline valence (standardized mean difference, $\mathrm{SMD} = -0.24$, \pvall{=}{.122}), baseline arousal ($\mathrm{SMD} = 0.18$, \pvall{=}{.239}), age ($\mathrm{SMD} = 0.14$, \pvall{=}{.378}), and daily social-media hours ($\mathrm{SMD} = 0.17$, \pvall{=}{.296}); the stage-two sample contained a higher proportion of women (59\% vs.\ 42\%), however, differences remained non-significant (Fisher's exact \pvall{=}{.054}).
The seven bridge controls did not differ significantly from stage-one controls on any measure (all Wilcoxon \pvall{\geq}{.089}).
These analyses render a batch explanation for the observed results unsupported on all measured characteristics; differences on unobserved characteristics cannot, however, be excluded.

\subsubsection{Registered exploratory behavioral features}
\label{apx:sm-behaviroal-features}
None of the registered behavioral features differed by condition: median scroll speed \anovans{2.0}{102.0}{1.73}{=}{.182}, median per-post viewing time \anovans{2.0}{110.7}{.04}{=}{.963}, number of posts viewed \anovans{2.0}{107.7}{.15}{=}{.863}, and likes \anovans{2.0}{100.2}{.15}{=}{.858}; Kruskal--Wallis tests agreed (all \pvall{\geq}{.36}).

\subsubsection{Descriptive statistics}
\label{apx:sm-descreptive-stats}
In Table~\ref{tab:msd-sm} we show mean and standard deviation values of valence, arousal, quality, time spent in first feed of the full analyzed sample.

\begin{table}[tbh]
  \centering
      \small
    \caption{\textbf{Rating and behavioral statistics.} Mean values (SD in parentheses) for valence, arousal, perceived image quality, and time spent in first feed (s) per condition.}
  \label{tab:msd-sm}
  \setlength{\tabcolsep}{4pt}
  \begin{tabular}{@{}lcccc@{}}
    \toprule
    \textbf{Condition} & \textbf{Valence} & \textbf{Arousal} & \textbf{Quality} & \textbf{Time in 1st feed (s)} \\
    \midrule
    Original  & 5.57 (2.12) & 4.23 (2.28) & 3.73 (1.15) & 109.15 (63.36) \\
    Grayscale & 5.12 (2.19) & 3.46 (2.05) & 3.39 (1.40) & 107.39 (107.82) \\
    RGDR      & 5.40 (1.82) & 4.54 (1.92) & 3.27 (1.10) & 102.40 (80.60) \\
    \bottomrule
  \end{tabular}
\end{table}

\end{document}